\documentclass[journal]{IEEEtran}
\usepackage{amsmath,amsfonts}
\usepackage{algorithmic}
\usepackage{algorithm}
\usepackage{array}
\usepackage[caption=false,font=normalsize,labelfont=sf,textfont=sf]{subfig}
\usepackage{textcomp}
\usepackage{stfloats}
\usepackage{url}
\usepackage{verbatim}
\usepackage{subfig}
\usepackage{graphicx}
\usepackage{pifont}      
\usepackage{fontawesome}
\usepackage{enumitem}
\usepackage{marvosym}
\usepackage{mathrsfs}

\usepackage{booktabs}
\usepackage{multirow}

\usepackage{xcolor}
\usepackage{cite}
\definecolor{brickred}{rgb}{0.8, 0.25, 0.33}
\definecolor{blueish}{rgb}{0.0, 0.3, 0.6}

\usepackage[pagebackref=true,breaklinks=true,colorlinks,citecolor=blueish,linkcolor=brickred,bookmarks=false]{hyperref}
\usepackage{cleveref}
\usepackage{amssymb}
\usepackage{cuted}
\usepackage{capt-of}

\def\ie{\emph{i.e.,\ }}

\newif\ifshowcomments
\showcommentstrue      
\showcommentsfalse    

\newif\ifshowrevisionone
\showrevisiononetrue      
\showrevisiononefalse

\Crefname{equation}{Eq.}{Eqs.}
\Crefname{figure}{Fig.}{Figs.}
\Crefname{table}{Tab.}{Tabs.}
\Crefname{section}{Sec.}{Secs.}

\newcommand{\parag}[1]{\paragraph{#1}}
\newcommand{\Teta}[0]{\mathbf{\Theta}}

\ifshowrevisionone
    \newcommand{\re}[1]{{\color{red}#1}}
\else
    \newcommand{\re}[1]{#1}
\fi

\ifshowcomments
  \newcommand{\yy}{\textcolor{blue}}
  \newcommand{\YY}[1]{{\color{blue}{\bf YY: #1}}}

  \newcommand{\RL}[1]{{\color{orange}{\bf RL: #1}}}

  \newcommand{\CD}[1]{{\color{purple}{\bf CD: #1}}}

  \newcommand{\PF}[1]{{\color{green}{\bf PF: #1}}}
\else
  \newcommand{\yy}[1]{#1}
  \newcommand{\YY}[1]{}

  \newcommand{\RL}[1]{}

  \newcommand{\CD}[1]{}

  \newcommand{\PF}[1]{}
\fi

\newcommand{\bz}{\mathbf{z}}
\newcommand{\bu}{\mathbf{u}}
\newcommand{\bX}{\mathbf{X}}

\newcommand{\bc}{\mathbf{c}}

\newcommand{\bV}{\mathbf{V}}
\newcommand{\bS}{\mathbf{S}}
\newcommand{\bD}{\mathbf{D}}
\newcommand{\bn}{\mathbf{n}}
\newcommand{\bN}{\mathbf{N}}
\newcommand{\bC}{\mathbf{C}}
\newcommand{\bm}{\mathbf{m}}
\newcommand{\bg}{\mathbf{g}}

\newcommand{\bx}[0]{\mathbf{x}}

\newcommand{\bd}[0]{\mathbf{d}}
\newcommand{\bM}[0]{\mathbf{M}}

\newcommand{\bU}[0]{\mathbf{U}}

\newcommand{\bs}[0]{\mathbf{s}}

\newcommand{\ours}[0]{{\it DMap}}
\newcommand{\dmapS}[0]{{\it DMap-Static}}
\newcommand{\dmapD}[0]{{\it DMap-Dynamic}}

\begin{document}

\title{Spatio-Temporal Garment Reconstruction Using Diffusion Mapping via Pattern Coordinates}

\author{Yingxuan You$^*$, Ren Li$^*$\textsuperscript{\Letter}, Corentin Dumery, Cong Cao, Hao Li, \IEEEmembership{Member, IEEE}, Pascal Fua, \IEEEmembership{Fellow, IEEE}
\thanks{
Yingxuan You, Corentin Dumery, and Pascal Fua 
are with the CVLab, École Polytechnique Fédérale de Lausanne (EPFL), Switzerland. 
(E-mail: \{yingxuan.you, corentin.dumery, pascal.fua\}@epfl.ch)
}
\thanks{
Ren Li and Cong Cao
are with the Mohamed bin Zayed University of Artificial Intelligence, United Arab Emirates. Ren Li is also with the Department of Computer Science and Engineering, Southern University of Science and Technology (E-mail: \{ren.li, cong.cao\}@mbzuai.ac.ae)
}
\thanks{
Hao Li 
is with the Pinscreen and Mohamed bin Zayed University of Artificial Intelligence, United Arab Emirates. 
(E-mail: hao@hao-li.com)
}
\thanks{$^*$Equal Contribution. \textsuperscript{\Letter}Corresponding Author: Ren Li.}
}
\markboth{IEEE TRANSACTIONS ON PATTERN ANALYSIS AND MACHINE INTELLIGENCE}%
{You \MakeLowercase{\textit{et al.}}: Spatio-Temporal Garment Reconstruction Using Diffusion Mapping via Pattern Coordinates}

\maketitle

\begin{strip}
\vspace{-35mm}
\centering
\includegraphics[width=\textwidth]{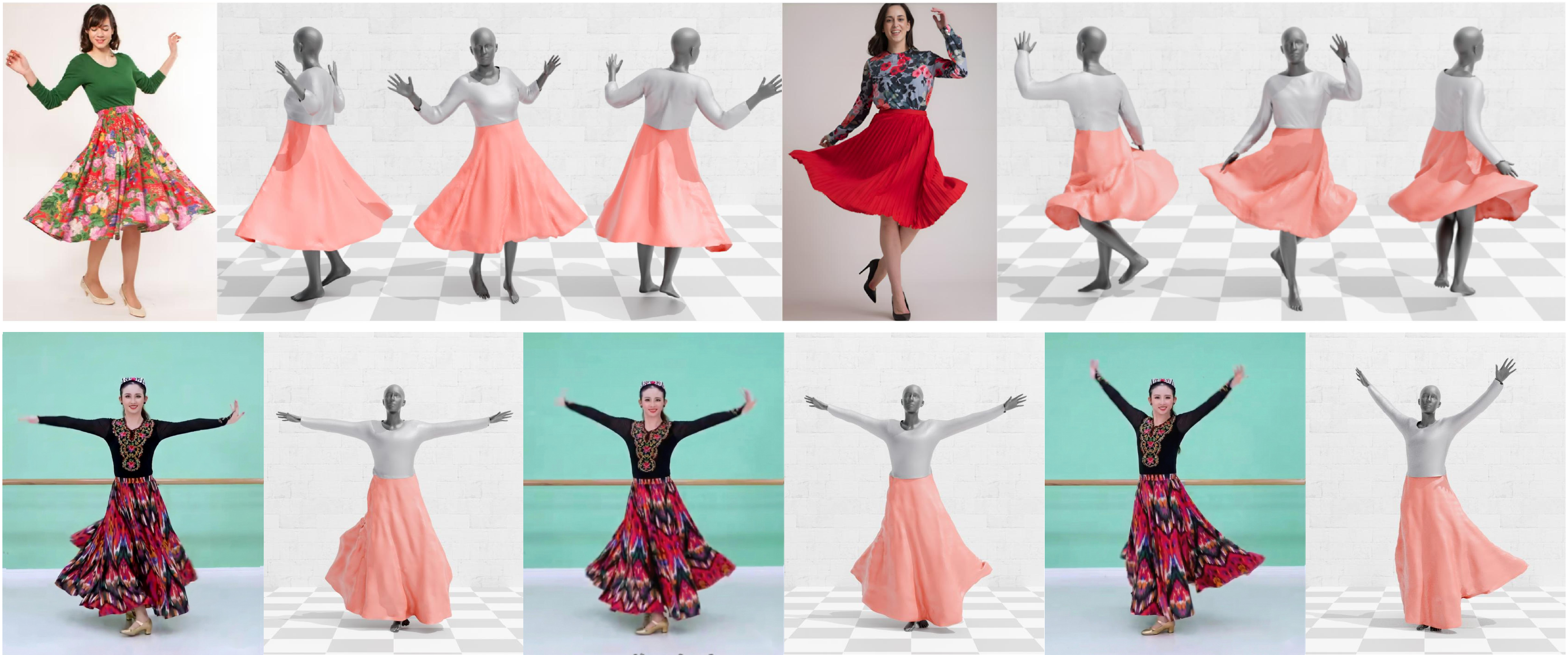}
\captionof{figure}{Given a single image~(top) or a monocular video~(bottom) of a clothed person, our proposed method can reconstruct high-fidelity 3D garment models with realistic details and temporal consistency.}
\label{fig:teaser}
\end{strip}

% !TEX root = ../main.tex
% !TEX spellcheck = en-US

\begin{abstract}
Reconstructing 3D clothed humans from monocular images and videos is a fundamental problem with applications in virtual try-on, avatar creation, and mixed reality. Despite significant progress in human body recovery, accurately reconstructing garment geometry, particularly for loose-fitting clothing, remains an open challenge. We propose a unified framework for high-fidelity 3D garment reconstruction from both single images and video sequences. Our approach combines \emph{Implicit Sewing Patterns} (ISP) with a generative diffusion model to learn expressive garment shape priors in 2D UV space. Leveraging these priors, we introduce a mapping model that establishes correspondences between image pixels, UV pattern coordinates, and 3D geometry, enabling accurate and detailed garment reconstruction from single images. We further extend this formulation to dynamic reconstruction by introducing a spatio-temporal diffusion scheme with test-time guidance to enforce long-range temporal consistency. We also develop analytic projection-based constraints that preserve image-aligned geometry in visible regions while enforcing coherent completion in occluded areas over time. Although trained exclusively on synthetically simulated cloth data, our method generalizes well to real-world imagery and consistently outperforms existing approaches on both tight- and loose-fitting garments. The reconstructed garments preserve fine geometric detail while exhibiting realistic dynamic motion, supporting downstream applications such as texture editing, garment retargeting, and animation.
\end{abstract}

% !TEX root = ../main.tex
% !TEX spellcheck = en-US

\section{Introduction} 
\label{sec:intro}

\IEEEPARstart{R}{ecovering} human body pose and shape, together with garment geometry, from visual input alone is a long-standing problem in computer vision with numerous applications, including fashion design, virtual try-on, 3D avatar creation, telepresence, and immersive VR/AR. In recent years, substantial progress has been made in modeling people wearing tight-fitting clothing, both in terms of body pose estimation~\cite{Bogo16,Lassner17a,Kanazawa18a,xu2024,li2024hyre} and garment shape reconstruction~\cite{Danerek17,Bhatnagar19,Jiang20d,Corona21,Moon22}. However, accurately modeling clothing, particularly loose-fitting garments, remains a significant challenge. Most existing approaches rely on a single unified 3D representation that jointly models the body and clothing. While such fused models can yield visually compelling results, they preclude realistic cloth simulation and limit applications such as virtual try-on.

To enable these applications, independent models for the body and garments are required. Unfortunately, garment modeling is inherently difficult due to the complex structure of clothing. Garments are thin surfaces with extremely high degrees of freedom and exhibit complex deformations driven by body motion and cloth dynamics. In addition, the diversity of garment designs and shape variations further complicates modeling and makes the acquisition of real 3D training data challenging, hindering the deployment of learning-based reconstruction methods. To mitigate these issues, several works~\cite{Danerek17,Bhatnagar19,Jiang20d,Casado22,Liu23b} adopt pre-designed mesh templates to define garment geometry and employ linear blend skinning (LBS)~\cite{Loper15} driven by an underlying body model to capture pose-induced deformations. While effective for tight-fitting clothing, this strategy requires predefined garment templates, limiting modeling flexibility and generalization. More critically, skinning-based methods struggle to represent loose garments that move independently and often deviate substantially from the body surface.

In prior work~\cite{Li23a,Li24a}, we addressed these limitations by introducing \emph{Implicit Sewing Patterns} (ISP), representing garments as collections of 2D panels with associated 3D surfaces. A learned deformation model allows these surfaces to deviate significantly from the body shape, enabling the modeling of loose garments. While effective, this approach tends to over-smooth the geometry, particularly in occluded regions that are not directly observable from the input image.

Moreover, the aforementioned methods are primarily designed for single-image reconstruction, which is insufficient for practical applications such as character animation, motion capture, and motion analysis, where temporal consistency is essential for both visual realism and physical plausibility. Applying single-frame methods independently to each frame of a video sequence produces severe temporal artifacts, including flickering and implausible garment motion, as these approaches fail to capture the highly nonlinear dynamics of clothing. Recent video-based approaches attempt to enforce temporal consistency, but they face notable limitations: body-driven methods~\cite{Qiu23a,Chen25a} struggle to capture large-scale and time-stable garment dynamics independent of the body, while approaches that learn garment-specific deformation fields~\cite{Guo24a,Dasgupta25} often over-smooth the geometry and inadequately model body-garment interactions.

In this paper, we introduce \ours, a unified diffusion-based framework for high-fidelity 3D garment reconstruction from both single images and video sequences. Our method leverages multiple diffusion processes to learn powerful garment shape priors capable of modeling complex geometries, completing occluded regions, and mapping 2D image observations to 3D and UV spaces to recover plausible garment shapes from static images. It also incorporates a spatio-temporal diffusion framework, integrating spatial priors with temporal dynamics through test-time guidance. This enables \ours{} to reconstruct dynamic garments with high geometric accuracy while preserving fine surface details and ensuring temporal consistency across video frames.

We proposed an early version, \dmapS{}, in a conference paper~\cite{Li25a}, but it was only designed for single images. In this journal paper, we extend it to \dmapD{} to also handle video sequences. This is far from a trivial extension. On one hand, the framework must ensure temporal consistency for long videos despite limited GPU memory. On the other hand, it must maintain the high-fidelity of per-frame reconstruction without suffering from the over-smoothing common in dynamic models~\cite{Guo24a,Dasgupta25}. To address this, our extended method builds a spatio-temporal framework that carefully leverages both spatial and temporal knowledge, together with advanced test-time guidance mechanisms.
The main contributions beyond the conference version are summarized as follows:
\begin{itemize}
\item We propose a spatio-temporal diffusion framework that explicitly decouples spatial and temporal modeling. The pre-trained spatial garment priors are reused without costly fine-tuning, while a lightweight, plug-and-play temporal module is trained to capture garment dynamics, enabling high-fidelity and temporally consistent 4D garment reconstruction.
\item We introduce a test-time guidance strategy that enforces long-range temporal consistency under limited GPU memory by blending learned garment priors with realistic constraints, preserving both spatial detail and temporal smoothness in long video sequences.
\item We develop analytic projection-based constraints that preserve visible garment geometry while enforcing spatio-temporal consistency in occluded regions.
\end{itemize}
We conduct extensive experiments on benchmark datasets and in-the-wild videos. Although trained exclusively on synthetically simulated cloth data, \ours{} generalizes well to real-world imagery and consistently outperforms existing approaches in terms of both geometric fidelity and temporal consistency. 
{The code and pre-trained models are publicly available at} \url{https://github.com/kasvii/DMap}.
% !TEX root = ../main.tex
% !TEX spellcheck = en-US

\section{Related Work}
\label{sec:related}

\subsection{Static Garment Reconstruction.}
Reconstructing clothed humans from single images has progressed significantly. Learning-based methods~\cite{Danerek17, Bhatnagar19, Jiang20d} predict vertex displacements and deform predefined garment templates into the target pose, while optimization-based methods~\cite{Casado22, Liu23b} refine vertex positions to fit estimated normal maps for fine details. However, the reliance on garment templates limits the flexibility and generalization of these methods. The methods of~\cite{Corona21, Moon22, Li22c} employ Signed Distance Functions (SDFs) to recover more diverse garment geometries. However, since SDFs represent watertight surfaces, modeling non-watertight garments requires enclosing them within thin volumes, which compromises geometric fidelity and hinders downstream refinements. In~\cite{DeLuigi23, Guillard22b}, the SDFs are replaced by Unsigned Distance Functions (UDFs) to remove this limitation.  Unfortunately, training a network to produce an accurate UDF is often more difficult than training it to yield an SDF, often resulting in inaccurate reconstructed 3D meshes with unwarranted holes. 
Recent works model 3D garments from 2D sewing patterns. For example, GarmentCodeData~\cite{korosteleva2024garmentcodedata} represents rest garment shapes with multiple panels and stitch information. However, obtaining deformed garments still requires draping or fitting them onto the body, which may produce plausible results but does not necessarily ensure alignment with image observations.
ISP~\cite{Li23a} solves this by leveraging 2D sewing patterns and UV-based parameterization to represent garments in a more structured and interpretable manner. However, it struggles to capture large garment deformations. 
\dmapS{}~\cite{Li25a} builds upon ISP and combines it with a generative diffusion model to learn the possible shape distribution represented by UV positional maps, allowing the reconstruction of highly detailed garments undergoing large non-rigid deformations. However, naively applying these single-image methods to video frames without explicitly enforcing temporal consistency often leads to jittery garment motion and unnatural dynamics. This issue is exacerbated by frame-to-frame inconsistencies in the reconstructed geometry, particularly in unseen surface regions.

\subsection{Dynamic Garment Reconstruction.}
Garment reconstruction from videos presents unique challenges due to complex clothing dynamics.
In~\cite{Li23d, Qiu23a}, a temporal deformation field that transfers vertices of a canonical template mesh to individual video frames is learned. However, the implicit encoding of time yields only weak temporal consistency, and the reliance on a fixed template mesh limits generality. The methods in~\cite{Jiang22a,Jiang22b,Paudel24a,Guo23a,Guo25a,Weng22a,Wang22d,Jiang24a} model garments using neural implicit functions, producing visually compelling reconstructions across a wide range of garment types. However, these methods fuse the body and clothes into a single entity, which limits their usefulness for downstream tasks such as garment editing and virtual try-on. 
Recent works separate garment and body representations for clothed human reconstruction. REC-MV~\cite{Qiu23a} and D$^3$-Human~\cite{Chen25a} rely on body-driven garment deformation, while MonoCloth~\cite{jin2026monocloth} adopts a 3D Gaussian Splatting representation~\cite{kerbl20233d} to reconstruct cloth-decoupled human avatars from monocular videos, which can produce high-quality reconstruction results. However, these methods still rely on body-driven deformation and Linear Blend Skinning (LBS), making it difficult to capture large and temporally stable displacements of loose garments away from the body.
ReLoo~\cite{Guo24a} introduces transferable virtual bones and global sequence optimization to improve temporal smoothness, but sacrifices per-frame accuracy and reconstruction details. NGD~\cite{Dasgupta25} models garment deformation with a neural Jacobian field and captures large motions of loose garments, but lacks sufficient modeling of body-garment interactions and mutual constraints.
With advances in sewing-pattern datasets such as GarmentCodeData~\cite{korosteleva2024garmentcodedata} and methods such as SPNet~\cite{lim2023spnet}, DressCode~\cite{he2024dresscode}, and ChatGarment~\cite{bian2025chatgarment}, a natural alternative is to first estimate the sewing pattern, drape the garment mesh onto the human body, and then fit it to image observations. Although such a pipeline can yield plausible animated garment meshes, it often has difficulty aligning the reconstructed geometry with the input observations. We provide an experimental analysis of this alternative pipeline in the Supplementary Material.
Overall, existing methods still struggle to provide the required balance between temporal consistency and spatial reconstruction accuracy under real-world physical interactions, which is precisely what our \ours{} is designed to address.

\subsection{Diffusion Models.}
Diffusion models~\cite{Ho20a, Song20c} learn complex data distributions through a forward-reverse noise process and have achieved remarkable success in generating high-quality and diverse images. Building on this foundation, recent works~\cite{Esser23a, Singer22a, Ho22a, Luo23a, Guo23b} extend image diffusion models to the temporal domain, enabling video processing and synthesis. In the context of garment reconstruction, diffusion-based shape priors have recently been explored~\cite{Guo24,Li24b}, leveraging UV-space parameterizations to model garment geometry. However, these methods require point clouds as 3D garment measurements and do not account for body-garment interaction during reconstruction. In contrast, our method reconstructs 3D garments directly from monocular 2D images while explicitly modeling both the garment and the underlying body. Furthermore, we extend spatial diffusion with a temporal module capable of processing video sequences. Instead of directly adopting video diffusion methods, which are typically restricted to short clips due to memory limitations, we introduce a testing-time guidance that incorporates realistic constraints, including long-range temporal consistency and accurate 2D-3D alignment, so that our approach overcomes the limitations of existing video diffusion methods and delivers high-fidelity and temporally consistent garment reconstructions across long video sequences.
% !TEX root = ../main.tex
% !TEX spellcheck = en-US

% !TEX root = ../top.tex
% !TEX spellcheck = en-US

\begin{figure*}[ht!]
    \centering
    \includegraphics[width=.98\textwidth]{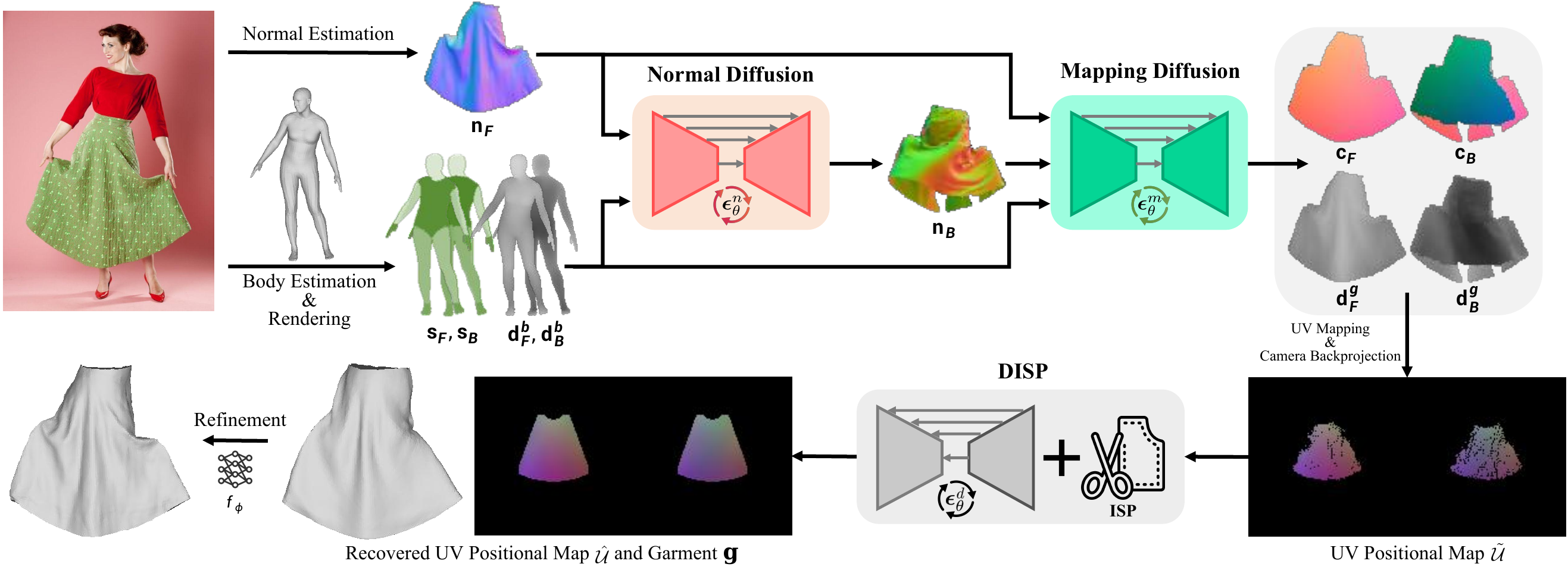}
    \vspace{-2mm}
    \caption{\textbf{Pipeline}.  Given an image of a clothed person, we first estimate the front normal $\bn_F$ of the target garment, and the SMPL body model which is used to render the body part segmentation ($\bs_F$, $\bs_B$) and depth ($\bd_F^b$, $\bd_B^b$) images. The back normal $\bn_B$ of the garment is estimated subsequently by the diffusion model $\boldsymbol{\epsilon}_{\theta}^n$. We then predict the UV-coordinate ($\bc_F$, $\bc_B$) and the depth ($\bd_F^g$, $\bd_B^g$) images from the garment normal and body estimations with the mapping model $\boldsymbol{\epsilon}_{\theta}^m$. The incomplete UV positional map $\Tilde{\mathcal{U}}$ is produced from them using the camera backprojection. Finally, we fit $\Tilde{\mathcal{U}}$ to DISP to recover the complete UV positional map $\hat{\mathcal{U}}$ and the corresponding garment mesh $\bg$, which is further improved by the refinement.}
    \vspace{-2mm}
    \label{fig:pipe}
\end{figure*}

\section{Garment Representation Model}
\label{sec:disp}
In this section, we introduce our garment representation models, ISP and DISP, which provide the reconstruction prior for both single-image and video-based garment reconstruction, as described in Sec.~\ref{sec:recon} and Sec.~\ref{sec:recon_vid}. 
Implicit Sewing Pattern (ISP)~\cite{Li23a} models canonical 3D garment surfaces through a 2D parameterization in a unified UV space. To capture the multiple plausible deformations induced by body motion and garment dynamics, DISP extends ISP by incorporating diffusion models to learn garment deformations in the positional maps, as illustrated by the gray network in Fig.~\ref{fig:pipe}.
The ISP/DISP representation is key because diffusion models can learn deformation priors more effectively and efficiently in regular 2D domains than directly on high-dimensional irregular 3D meshes. Moreover, the unified UV space establishes a pixel-UV-3D mapping, which directly connects image observations with garment surface geometry. This mapping enables image-geometry alignment, spatio-temporal guidance, and detail fitting.
In the remainder of this section, we first introduce ISP and then describe how we incorporate a diffusion model into it.

\subsection{Implicit Sewing Patterns}
\label{sec:isp}

\parag{Formalization} 
Implicit Sewing Patterns (ISP)~\cite{Li23a} is a garment model based on the sewing patterns used in the fashion industry to design and manufacture clothes. A sewing pattern is made of several 2D panels along with stitch information for assembling them together. The 2D panels and the stitching are implicitly modeled using a 2D signed distance field (SDF) and a 2D label field, respectively.  For a specific garment, its corresponding latent code $\bz$, and a point $\bu$ in the 2D UV space $\Omega=[-1,1]^2$, the ISP model outputs the signed distance $s$ to the panel boundary and a label $l$ using a fully connected network $\mathcal{I}_{\Teta}$ as
\begin{equation} \label{eq:pattern}
    (s,l) = \mathcal{I}_{\Teta}(\bu,\bz) \; . 
\end{equation}
The zero crossing of the SDF defines the shape of the panel, with $s<0$ indicating that $\bu$ is within the panel and $s>0$ indicating that $\bu$ is outside the panel. The label $l$ encodes the stitch information, instructing which panel boundaries should be stitched together. To map the 2D sewing patterns to 3D surfaces, a UV parameterization function $\mathcal{A}_{\Phi}$ is learned to perform the 2D-to-3D mapping
\begin{equation}
    \bX = \mathcal{A}_{\Phi}(\bu,\bz) \; ,
\end{equation}
where $\bX\in\mathbb{R}^3$ represents the 3D position of $\bu$. In essence, ISP registers different garments onto a unified UV space and establishes the mapping functions between points in UV space and the 3D garment surfaces. The shape of SDF's 0-crossing defines the geometry of garment in its rest state. As ISP is a differentiable representation, we can easily fit a latent code $\bz$ to arbitrary masks or contours of the panels to recover the corresponding garment geometry.

\parag{Training}
Training ISP requires the 2D sewing patterns of rest-state 3D garments. However, they are not available in most garment datasets, e.g. CLOTH3D \cite{Bertiche20}. Following the garment flattening strategy of \cite{Li24a,Pietroni22}, we cut the garment mesh of CLOTH3D into front and back surfaces according to predefined cutting rules and then flatten them into 2D panels by minimizing an as-rigid-as-possible energy~\cite{Liu08c}.
For each garment in the dataset, a front and a back panel are generated as its sewing pattern. By using the paired 2D sewing patterns and their 3D meshes, we learn the weights of the ISP model $(\mathcal{I}_{\Teta}, \mathcal{A}_{\Phi})$ with the training procedure of \cite{Li23a}.

\subsection{Extending ISP with a Diffusion Model}
\label{sec:deformation}

For a specific garment, the UV parameterization function of ISP only produces a single UV positional map $\mathcal{U}_r$ to model its 3D shape in the rest state
\begin{equation}\label{eq:uv_rest}
    \mathcal{U}_r[u,v] = 
    \begin{cases}
        \mathcal{A}_{\Phi}(\bu,\bz), & \mbox{if } s_{\bu} \le 0\\
        \varnothing, & \mbox{if } s_{\bu} > 0
    \end{cases}
    \; ,
\end{equation}
where $\bu=(u,v)$, $s_{\bu}$ is the SDF value of $\bu$, $[\cdot,\cdot]$ denotes the standard array addressing and $\varnothing=(-1,-1,-1)$ indicates the region outside the panel. 
When dressed on the body, the garment can have various deformations due to the motion of the underlying body, which is not able to be modeled by ISP solely. Inspired by \cite{Guo24,Li24b}, we first introduced \dmapS{} in a SIGGRAPH paper~\cite{Li25a}, which incorporates a diffusion model into ISP to capture these possible deformations by generating plausible UV maps.

Given the deformed garments worn on the body whose rest states are modeled by ISP as Eq. \ref{eq:uv_rest}, we write the corresponding UV maps
\begin{equation}\label{eq:uv_map}
    \mathcal{U}[u,v] = 
    \begin{cases}
        \bV, & \mbox{if } s_{\bu} \le 0\\
        \varnothing, & \mbox{if } s_{\bu} > 0
    \end{cases}
    \; ,
\end{equation}
where $\bV\in \mathbb{R}^3$ is the corresponding position on the deformed mesh surface for the UV point $\bu=(u,v)$. Each $\mathcal{U}$ represents a specific deformed shape for a particular garment. We use a diffusion model~\cite{Ho20} $\boldsymbol{\epsilon}^d_{\boldsymbol{\theta}}$ to learn the distribution of plausible deformations represented by $\mathcal{U}$. 

For each garment sample, we generate its UV map $\mathcal{U}$ according to Eq. \ref{eq:uv_map}, along with a panel mask $\mathcal{M}$ as
\begin{equation}\label{eq:uv_mask}
    \mathcal{M}[u,v] = 
    \begin{cases}
        1, & \mbox{if } s_{\bu} \le 0\\
        0, & \mbox{if } s_{\bu} > 0
    \end{cases}
    \; .
\end{equation}
$\mathcal{M}$ depicts the panel shape, which itself encodes the 3D geometry of the canonical rest garment. We concatenate $\mathcal{U}$ and $\mathcal{M}$ along the channel dimension to form the training samples and train the network $\boldsymbol{\epsilon}^d_{\boldsymbol{\theta}}$ on them.
After training, the diffusion model and ISP together form the garment model DISP.

\section{Static Garment Reconstruction}
\label{sec:recon}

Monocular images provide only partial 2D observations of visible regions. To complete the missing information in occluded areas and enable globally consistent reconstruction, we propose \dmapS{}, a framework first introduced in [20], that combines three diffusion schemes trained to
(1) learn a garment shape prior,
(2) infer image information for occluded garment regions, and
(3) map 2D image to UV and 3D spaces to recover plausible 3D shapes align with the learned prior.

\subsection{Observations from Images}
\label{sec:observations}
Recent advances in segmentation~\cite{Kirillov23,Ravi24}, normal estimation~\cite{Bae24,Khirodkar25} and human mesh recovery~\cite{Goel23a,Stathopoulos24} can be used to extract accurate observations from an image of a clothed person. In this manner, 
\dmapS{}~\cite{Li25a} first segments the target garment using \cite{Kirillov23,Li20k} and estimates its normals $\bn_F$ using \cite{Khirodkar25}. To model the body underneath, we use SMPL \cite{Loper15},  which relies on two sets of parameters $(\beta,\theta)$ to describe the body shape and pose respectively. The SMPL parameters are estimated from the image by \cite{Goel23a, shin2024wham} to infer the 3D body shape, which are then used to render front and back body part segmentations $\bs_F$ and $\bs_B$, along with front and back depth images $\bd_F^b$ and $\bd_B^b$, as shown in the top left of Fig. \ref{fig:pipe}.

To estimate the invisible normals, typically in the back, $\bn_B$ as shown in Fig. \ref{fig:pipe}, we use the estimated normal $\bn_F$ to guide a conditional diffusion model $\boldsymbol{\epsilon}_{\theta}^n$. The denoising process of $\boldsymbol{\epsilon}_{\theta}^n$ is conditioned on the visible normals $\bn_F$, the front and back segmentation images $\bs = (\bs_F, \bs_B)$, the body depth maps $\bd^b = (\bd_F^b, \bd_B^b)$. It is learned by minimizing the loss
\begin{equation}
    \mathcal{L} = \mathbb{E}_{t,\bN_B,\boldsymbol{\epsilon}}\| \boldsymbol{\epsilon} - \boldsymbol{\epsilon}_{\theta}^n\left( \sqrt{\bar{\alpha}_{t}}\bn_B+\sqrt{1-\bar{\alpha}_{t}}\boldsymbol{\epsilon}, \bn_F, \bs, \bd^b, t \right) \|^2 \; .
\end{equation}
The conditioning images of the garment and body provide information to generate plausible normals for the back of the garment. As will be shown in our experiments, the back normal estimation $\bn_B$ provides additional constraints to regularize the garment fitting process, which improves reconstruction fidelity.

\subsection{Mapping from Pixel Space to UV Space and 3D Space}
\label{sec:mapping}
% !TEX root = ../top.tex
% !TEX spellcheck = en-US

\begin{figure}[ht!]
    \centering
    \includegraphics[width=.48\textwidth]{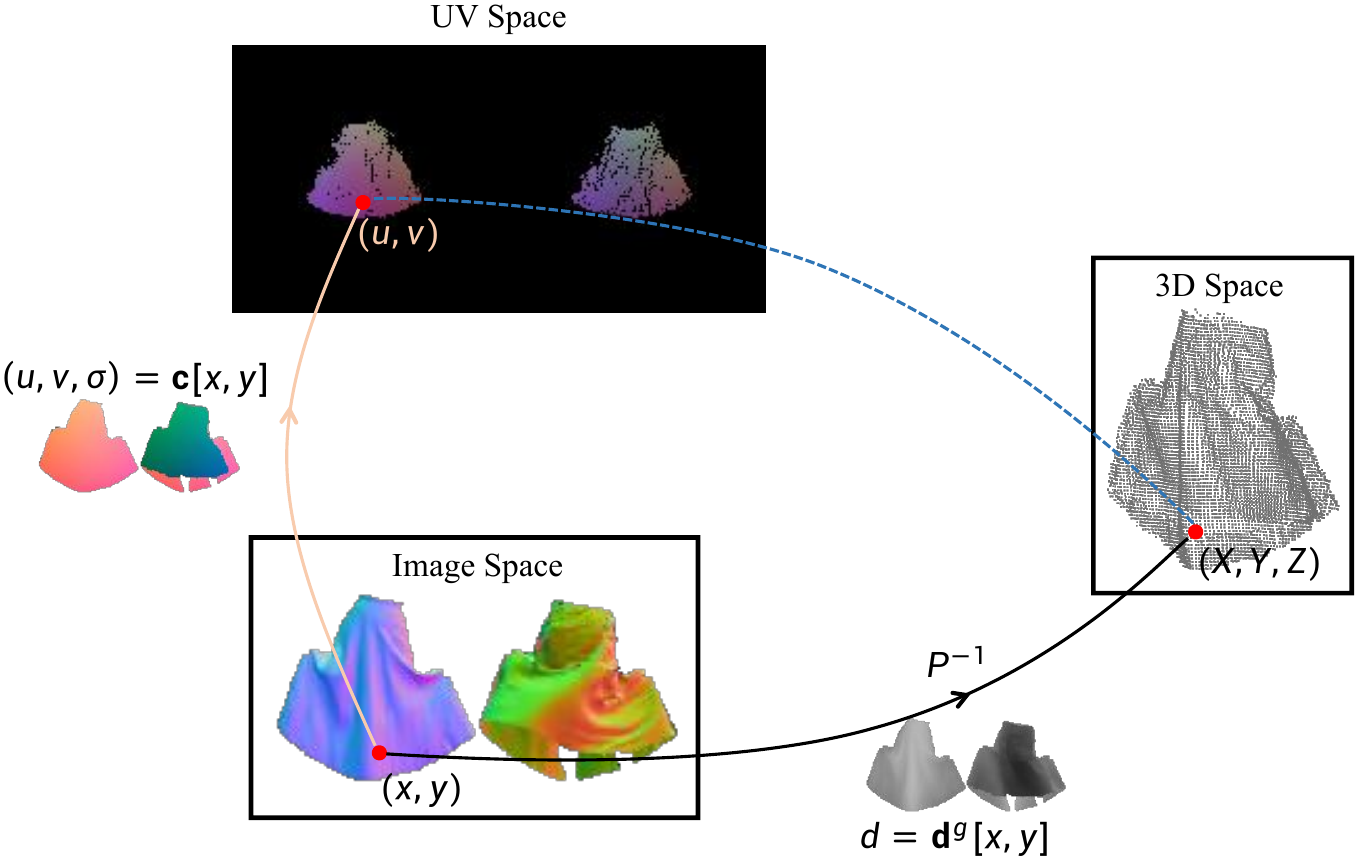}
    \vspace{-2mm}
    \caption{\textbf{Mapping between pixel, 3D, and UV spaces.} The pixel $(x,y)$ is mapped to $(X,Y,Z)$ in the 3D space using the estimated depth $d$ and the camera backprojection $P^{-1}$, and to $(u,v)$ in the UV space using the estimated UV coordinates $(u,v,\sigma)$. The dash line indicates that $(X,Y,Z)$ and $(u,v)$ are connected indirectly through $(x,y)$.}
    \label{fig:mapping}
    % \vspace{-2mm}
\end{figure}

The normal estimations provide observations in pixel space, while the garment model DISP is learned in the UV space of the garment panels, and the garment surface resides in the 3D space. To reconstruct 3D garments using our DISP, it is thus necessary to connect these three different spaces. To this end, \dmapS{} introduce a mapping function that translates image observations from the pixel space to both the UV space and the 3D space, as illustrated by Fig. \ref{fig:mapping}.

\parag{To 3D Space} 

Since the depth and surface normal are closely related in terms of 3D geometry, we estimate the garment depth image $\bd^g$ from normal estimations $\bn = (\bn_F, \bn_B)$ conditioned on the body depth $\bd^b$. For the foreground pixel $(x,y)$, its absolute depth value is $d=\bd^g[x,y]$. By leveraging the camera projection $P$, we can have the 3D coordinate $(X,Y,Z)$ for each pixel
\begin{equation}\label{eq:bproj}
    (X,Y,Z) = P^{-1}(x,y,d)\; ,
\end{equation}
where $P^{-1}$ denotes the camera backprojection. Through Eq. \ref{eq:bproj}, we establish the mapping from pixel space to 3D space.

\parag{To UV Space} 
Given the normal estimation $\bn$, we train a network $\boldsymbol{\epsilon}_{\theta}^m$ to predict a UV-coordinate image $\bc$ conditioned on the body part segmentation $\bs$. The pixel value of $\bc$ is 
\begin{equation}\label{eq:uv}
    \bc[x,y]  = (u,v,\sigma) \; , 
\end{equation}
where $(u,v)$ is the predicted coordinate on the UV space of the panel for pixel $(x,y)$, $\sigma$ indicates whether it belongs to the front ($\sigma > 0$) or the back ($\sigma < 0$) panel. Through Eq. \ref{eq:uv}, we establish the mapping from the pixel space to the UV space.

By assembling the results of UV and 3D mapping of Eq. \ref{eq:uv} and Eq. \ref{eq:bproj}, we can get a UV map $\Tilde{\mathcal{U}}$, where
\begin{equation}
    \Tilde{\mathcal{U}}[u,v] = P^{-1}(x,y,d) = (X,Y,Z)\; .
\end{equation}
For the positions on $\Tilde{\mathcal{U}}$ without projected points, we simply set their values to $\varnothing$. We also compute a mask $\Tilde{\mathcal{M}}$ with
$\Tilde{\mathcal{M}}[u,v] = 1$ at where a pixel is projected, and $\Tilde{\mathcal{M}}[u,v] = 0$ otherwise. Due to occlusions, both $\Tilde{\mathcal{U}}$ and $\Tilde{\mathcal{M}}$ are incomplete. In the next section, we will complete them by fitting to the DISP priors.

\parag{Training} 
We learn the mapping function in an image-to-image translation fashion with a conditional diffusion model $\boldsymbol{\epsilon}_{\theta}^m$. For the normal estimation $\bn_F$ and $\bn_B$, $\boldsymbol{\epsilon}_{\theta}^m$ is trained to predict their UV-coordinate image $\bc_F$ and $\bc_B$, and depth images $\bd_F^g$ and $\bd_B^g$ jointly. 
The denoising process of $\boldsymbol{\epsilon}_{\theta}^m$ is conditioned on the estimated normals of the front and the back $\bn = (\bn_F, \bn_B)$, the segmentation images $\bs = (\bs_F, \bs_B)$, the body depth maps $\bd^b = (\bd_F^b, \bd_B^b)$, and is learned by minimizing the loss
\begin{equation}
    \mathcal{L} = \mathbb{E}_{t,\bm_0,\boldsymbol{\epsilon}}\| \boldsymbol{\epsilon} - \boldsymbol{\epsilon}_{\theta}^m\left( \sqrt{\bar{\alpha}_{t}}\bm_0+\sqrt{1-\bar{\alpha}_{t}}\boldsymbol{\epsilon}, \bn, \bs, \bd^b, t \right) \|^2 \; ,
\end{equation}
where $\bm_0=[\bc_F, \bc_B, \bd_F^g, \bd_B^g]$. After the training, we assemble the results for both the front and the back to produce the UV map $\Tilde{\mathcal{U}}$ and the mask $\Tilde{\mathcal{M}}$. Compared with only using the front result, this provides more observations and constraints for the fitting, resulting in a reconstruction with higher quality for both the visible and invisible parts.

\subsection{Fitting}
\label{sec:fitting_dmap}
The incomplete panel mask $\Tilde{\mathcal{M}}$ and the incomplete UV map $\Tilde{\mathcal{U}}$ provide partial information of the garment geometry and deformation, respectively. To recover a complete garment from them, we leverage the prior of DISP. We first recover the complete panel mask to recover the garment geometry in its rest shape, and then recover the complete UV map for the deformation. 
However, very fine wrinkles may still be smoothed due to the synthetic-to-real domain gap and the low-frequency bias of neural networks. Specifically, the model learns garment priors mainly from synthetic data, whereas real images may contain different wrinkle, texture, and lighting patterns, while neural networks tend to favor global garment structures over high-frequency wrinkle details. To mitigate these limitations and further improve reconstruction accuracy, we introduce a post-optimization step that aligns the garment more closely with image observations.

% !TEX root = ../top.tex
% !TEX spellcheck = en-US

\begin{figure}[ht!]
    \centering
    \includegraphics[width=.48\textwidth]{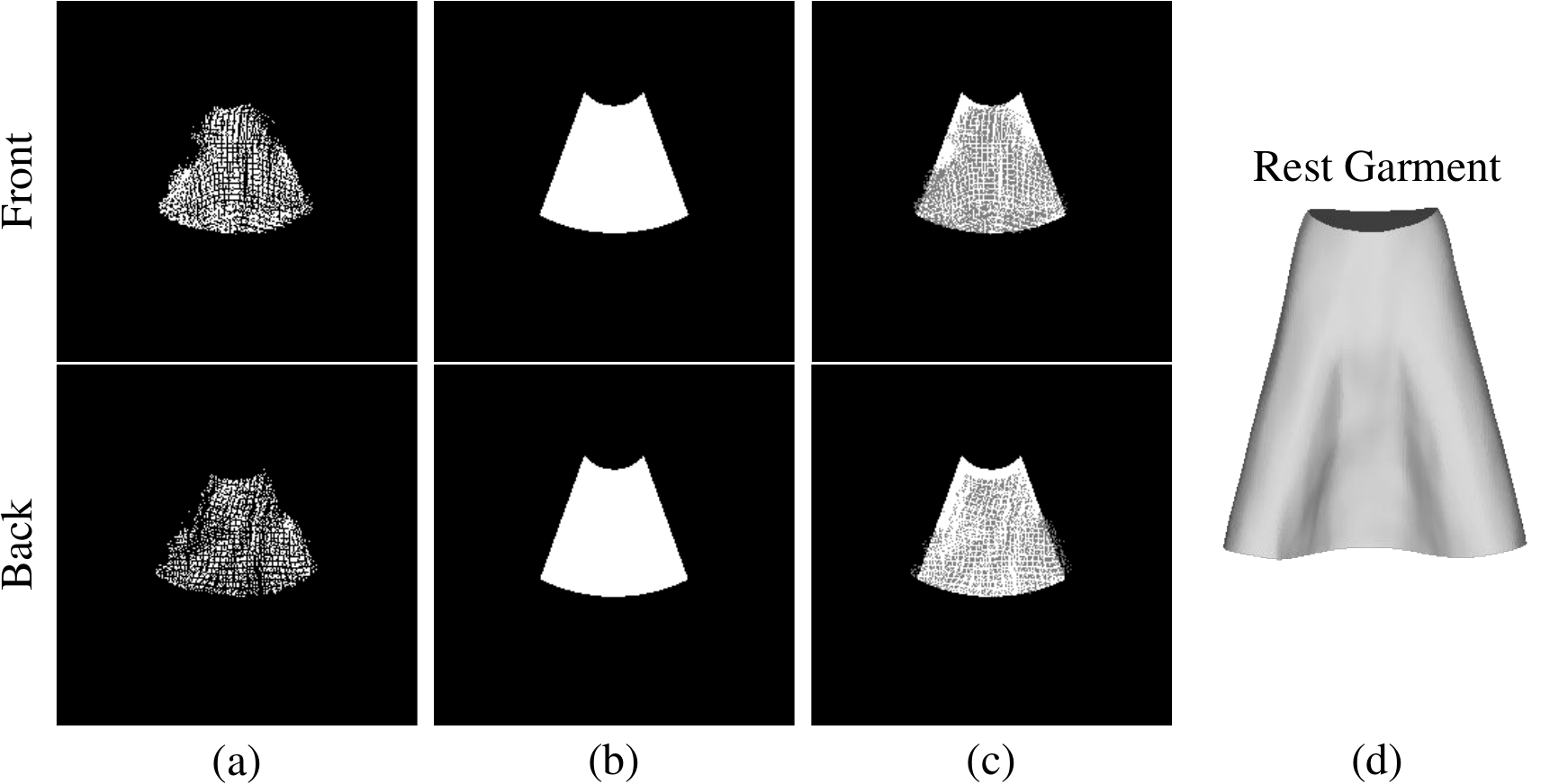}
    \vspace{-2mm}
    \caption{\textbf{Recovering garment rest geometry}. Given (a) the incomplete panel mask $\Tilde{\mathcal{M}}$, we fit (b) the complete panel mask $\mathcal{M}$ by Eq. \ref{eq:z}. (c) shows the overlay of $\Tilde{\mathcal{M}}$ in gray and $\mathcal{M}$ in white. (d) is the corresponding rest-state garment mesh $\bar{\bg}$ for (b).}
    \label{fig:mask}
\end{figure} 

\subsubsection{Recovering the Rest Geometry.} 
\label{sec:recoverPanel}
To recover the garment rest geometry represented by the 2D panel shape from $\Tilde{\mathcal{M}}$, we optimize the latent code $\bz$ of Eq.~\ref{eq:pattern} so that its corresponding patterns match $\Tilde{\mathcal{M}}$ as well as possible. The optimization objective is
\begin{equation} \label{eq:z} 
    \mathcal{L}(\bz) = \sum\limits_{\scalebox{.7}{$\bu\in\mathcal{M}_+$}}ReLU(s_\bu(\bz)) - \lambda_{area}\sum\limits_{\scalebox{.7}{$\bu\in\Omega$}}s_\bu(\bz) + \lambda_\bz||\bz||_2\; ,
\end{equation}
where $\mathcal{M}_+=\{\bu|\tilde{\mathcal{M}}_\bu=1, \bu\in\Omega\}$, $s_\bu(\bz)$ is the SDF value of $\bu$ computed by ISP, and $\lambda_{area}$ and $\lambda_\bz$ are the weighting constants. The first item in Eq. \ref{eq:z} ensures that the projected UV points are within the panel, while the second one penalizes large panel area to make the panel contours surround the non-zero points of $\tilde{\mathcal{M}}$ as closely as possible. This optimization produces an optimal latent code $\bz^*$ that we can use to infer a complete panel mask $\mathcal{M}$ and a rest-state garment mesh $\bar{\bg}$ as shown in Fig. \ref{fig:mask}.

\subsubsection{Recovering Deformed Geometry.} 
\label{sec:recoverUV}
The diffusion model $\boldsymbol{\epsilon}^d_{\boldsymbol{\theta}}$ of DISP learns the distribution of plausible deformations represented by UV maps. To recover the full UV map $\mathcal{U}$, we use the partial UV map $\tilde{\mathcal{U}}$ and the recovered panel mask $\mathcal{M}$ as the manifold guidance~\cite{Chung22a,Chung22b} in the reverse diffusion process of $\boldsymbol{\epsilon}^d_{\boldsymbol{\theta}}$:
\begin{small}
\begin{align}
    \nabla_{\bx_t}\log p(\bx_t|\tilde{\mathcal{U}}, \tilde{\mathcal{M}}, \mathcal{M}) &\simeq -\frac{\boldsymbol{\epsilon}^d_{\boldsymbol{\theta}}(\bx_t, t)}{\sigma_t}-\rho\nabla_{\bx_t}\mathcal{L}(\hat{\bx}_0, \tilde{\mathcal{U}}, \tilde{\mathcal{M}}, \mathcal{M}) \;  ,  \nonumber \\
    \hat{\bx}_0 &= \frac{1}{\sqrt{\bar{\alpha}_t}}\bx_t - \sqrt{\frac{1-\bar{\alpha}_t}{\bar{\alpha}_t}}{\boldsymbol{\epsilon}^d_{\boldsymbol{\theta}}}(\bx_t, t)\;  \; ,  \label{eq:guided_loss}
\end{align}
\end{small}
\hspace{-0.5em}where $\rho$ is the guidance step size. $\mathcal{L}$ is the function that measures the difference between the generated and the given UV maps and panel masks
\begin{equation} \label{eq:guidance} 
    \mathcal{L}(\hat{\bx}_0, \tilde{\mathcal{U}}, \tilde{\mathcal{M}}, \mathcal{M}) = \lVert\tilde{\mathcal{M}}*(\hat{\mathcal{U}} - \tilde{\mathcal{U}})\rVert _2 + \lVert\hat{\mathcal{M}} - \mathcal{M}\rVert _1 \; ,
\end{equation}
where $\hat{\bx}_0=[\hat{\mathcal{U}}, \hat{\mathcal{M}}]$, $\hat{\mathcal{U}}$ and $\hat{\mathcal{M}}$ refer to the generated UV map and panel mask respectively, and $*$ denotes the element-wise multiplication. With the generated UV map $\hat{\mathcal{U}}$, we update the vertices of $\bar{\bg}$ to get the recovered mesh $\bg$ as shown in the bottom-left of Fig. \ref{fig:pipe}.

\subsubsection{Garment Refinement} 
\label{sec:post}
As the diffusion model learns the shape distribution from the garment simulation data, which is generated with limited materials, external forces, and body motions, when handling in-the-wild images that are out-of-distribution, it can produce inaccurate UV maps by Eq. \ref{eq:guided_loss} and result in an inaccurate garment mesh that does not align with the images. To further improve the reconstruction accuracy, we refine the recovered mesh $\bg$ in Sec. \ref{sec:recoverUV} by optimizing its vertex positions to align it with the image observations. The loss function we use is
\begin{small}
\begin{equation} \label{eq:postrefine} 
    \mathcal{L} \!= \!\lambda_m\mathcal{L}_{mask} \! + \! \lambda_d\mathcal{L}_{depth}  \!+ \! \lambda_n\mathcal{L}_{normal}  \!+ \!\lambda_u\mathcal{L}_{uv} + \lambda_p\mathcal{L}_{phys} ,
\end{equation}
\end{small}
\hspace{-0.4em}where $\lambda_m$, $\lambda_d$, $\lambda_n$, $\lambda_u$ and $\lambda_p$ are scalar weights. $\mathcal{L}_{mask}$, $\mathcal{L}_{depth}$ and $\mathcal{L}_{normal}$ penalize the difference between the rendered front and back masks, depth and normal of the mesh $\bg$ and their corresponding estimation, respectively. $\mathcal{L}_{uv}$ ensures the corresponding UV map $\hat{\mathcal{U}}$ of $\bg$ aligns with the partial UV observations $\tilde{\mathcal{U}}$ by
\begin{equation} \label{eq:luv} 
    \mathcal{L}_{uv} = \lVert\tilde{\mathcal{M}}*(\hat{\mathcal{U}} - \tilde{\mathcal{U}})\rVert _2.
\end{equation}
$\mathcal{L}_{phys}$ contains a set of physics-based mesh regularization \cite{Narain12,Santesteban22} using the recovered rest-state garment $\bar{\bg}$ as the reference
\begin{equation} \label{eq:lphys} 
    \mathcal{L}_{phys} = \mathcal{L}_{strain} + \mathcal{L}_{bend} + \mathcal{L}_{gravity} + \mathcal{L}_{collision} ,
\end{equation}
where $\mathcal{L}_{strain}$ is the membrane strain energy caused by the deformation, $\mathcal{L}_{bend}$ is the bending energy resulting from the folding of adjacent faces, $\mathcal{L}_{gravity}$ is the gravitational potential energy and $\mathcal{L}_{collision}$ is the penalty for body garment collision. 

However, directly optimizing the vertex positions using Eq. \ref{eq:postrefine} will lead to a suboptimal solution, as vertices are not strongly coupled.
Inspired by \cite{Ulyanov18,Gadelha21}, we introduce a multi-layer perceptron (MLP) for the optimization to update vertices by a neural displacement field. To be specific, we initialize an MLP network $f$ with learnable parameters $\phi$. Given the vertex $V$ of mesh $\bg$ and its canonical position $\bar{V}$ on $\bar{\bg}$, the network $f_\phi$ predicts its displacement by
\begin{equation} \label{eq:nn} 
    \Delta V = f_\phi(V,\bar{V}) .
\end{equation}
We use the updated vertex position $V+\Delta V$ to compute the loss of Eq. \ref{eq:postrefine} and compute the gradient with respect to $\phi$ for its learning. Since neural networks tend to learn low-frequency functions \cite{Rahaman19}, the result after this step is a bit smooth. To further recover fine surface details, we perform an additional refinement step by directly optimizing the garment mesh vertices with Eq. \ref{eq:postrefine}.

% !TEX root = ../main.tex
% !TEX spellcheck = en-US

\begin{figure*}[t]
    \centering
    \includegraphics[width=\linewidth]{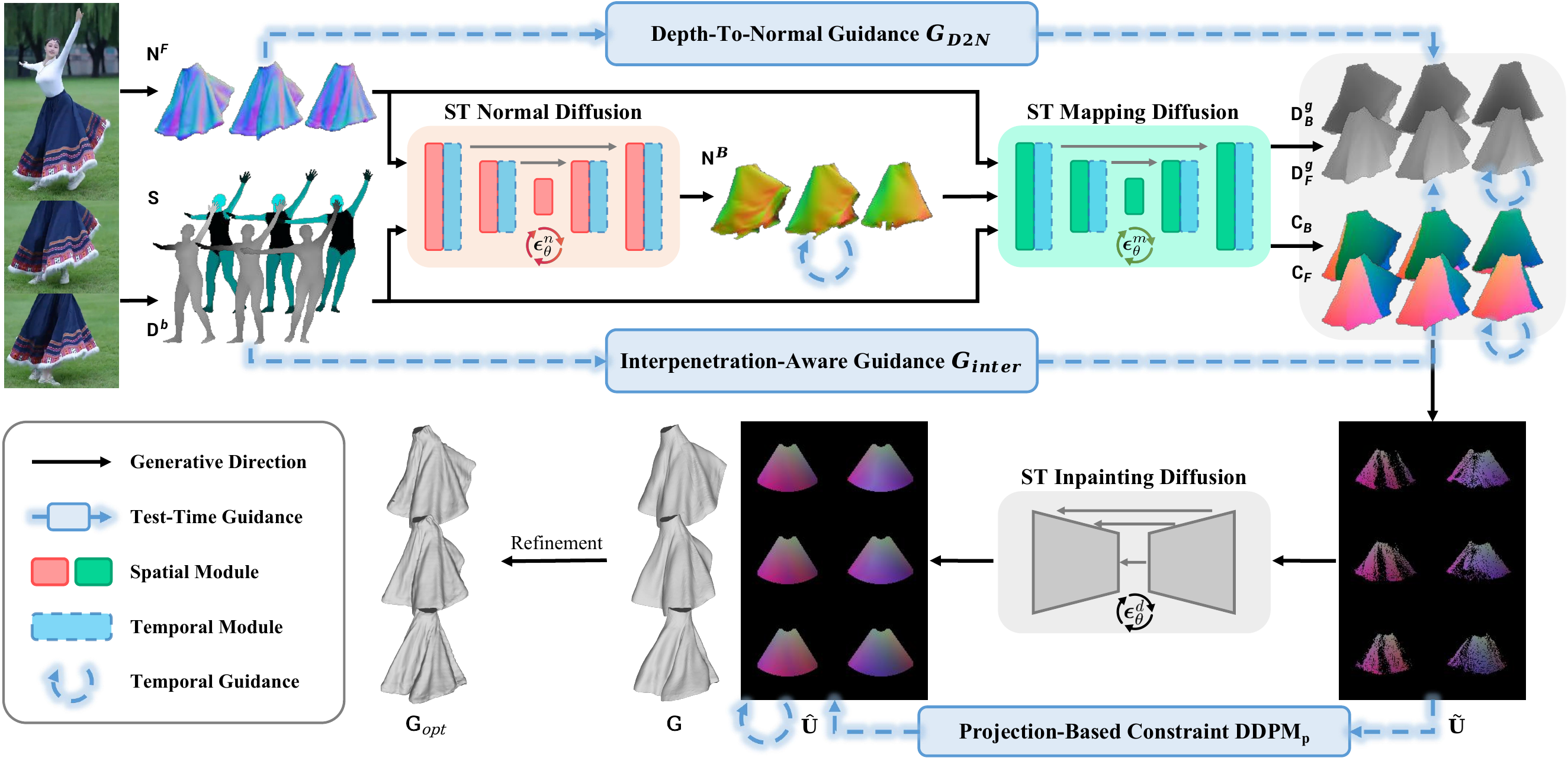}
    \vspace{-4mm}
    \caption{{\bf Processing a video sequence.} 
    Given a set of images with extracted body segmentations $\mathbf{S}$, body depths $\mathbf{D}^b$, and garment normals $\mathbf{N}^F$, our method produces a sequence of garment meshes $\mathbf{G}$ in three steps. 
    First, the back-view normals $\mathbf{N}^B$ of the garment are inferred. By design, our method ensures these predictions are temporally consistent. 
    Second, a mapping network estimates the 2D/3D positions of each pixel, where the 2D positions $(\mathbf{C}_F,\mathbf{C}_B)$ are in a reference \textit{pattern} space, and the 3D positions are represented as depth maps $(\mathbf{D}_F^g,\mathbf{D}_B^g)$. We introduce novel guidance on this generation to match normal estimations and prevent intersection with the body.
    Finally, this mapping is unwrapped into a partial 2D pattern $\tilde{\mathbf{U}}$ where pixel value encodes the 3D position, and our temporal inpainting diffusion completes these partial observation\yy{s} into a full garment sequence while ensuring the partial constraints are respected.}
    \label{fig:pipeline}
\end{figure*}

\section{Dynamic Garment Reconstruction}
\label{sec:recon_vid}

While \dmapS{}~\cite{Li25a} from Sec.~\ref{sec:recon} delivers good results from single images, due to the stochastic nature of diffusion, there is no reason for results generated independently in consecutive video frames to be consistent. Thus, frame-by-frame reconstruction does no yield realistic animations. To remedy this, we extend it into \dmapD{} that enforces consistency using a spatio-temporal diffusion framework, as shown in Fig. \ref{fig:pipeline}. The main difficulty in doing so is that spatio-temporal diffusion typically requires large amounts of memory. Thus, only short video snippets can be handled at any one time. We address this by building temporal diffusion modules with across- and within-subsequence guidance to enforce temporal consistency beyond short clips. We also incorporate an analytic projection-based constraint that faithfully preserves the visible garment geometry while enforcing spatial and temporal consistency in occluded areas. We discuss these below. 

\begin{figure*}[t]
    \centering
    \includegraphics[width=\linewidth]{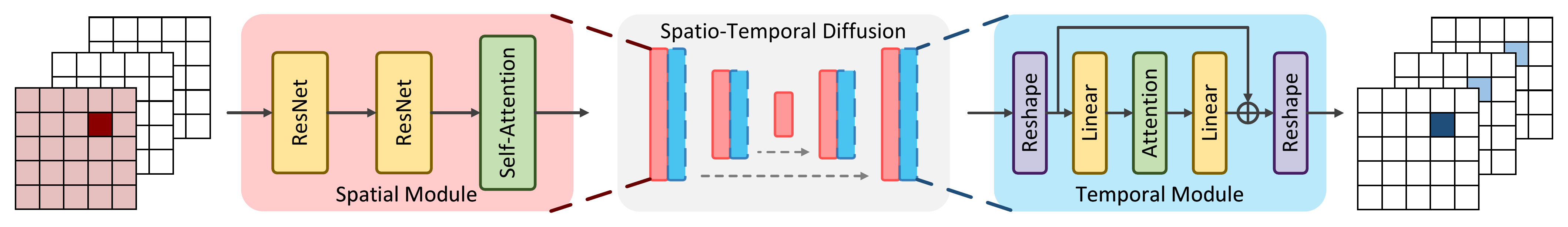}
    \vspace{-5mm}
    \caption{\textbf{Architecture of the spatio-temporal diffusion model.} The model decouples spatial and temporal knowledge into separate modules: the spatial module (red) captures per-frame spatial structure, while the temporal module (blue) models per-pixel motion across frames. Together, these components form the foundation for high-fidelity and temporally smooth 4D garment reconstruction.}
    \label{fig:block}
    \vspace{-2mm}
\end{figure*}

\subsection{Spatio-Temporal Diffusion}
\label{sec:ST-diffusion}
To generate a temporally consistent reconstruction over the entire video sequence, we introduce a spatio-temporal diffusion model for sequential generation of back-view normal map and 2D-to-3D mapping. As shown in Fig.~\ref{fig:block}, the model decouples spatial and temporal modeling through an alternating design: the spatial module captures per-frame geometric structure, while the temporal module estimates pixel-wise motion over time. Together, they provide a unified basis for high-fidelity and temporally coherent 4D garment reconstruction.
\subsubsection{Spatial Module}
The spatial module inherits both the architecture and weights of \dmapS{}, enabling us to reuse pre-trained spatial priors without the need for costly fine-tuning. As illustrated on the left of Fig.~\ref{fig:block}, the module consists of convolutional and self-attention layers. Specifically, it computes self-attention across pixels within individual frames by processing a tensor of shape $\mathbb{R}^{B' \times H \times W \times C}$, where the effective batch size $B' = B \times T$ combines the original batch size $B$ and the number of frames $T$. This design makes the module focuses exclusively on extracting spatial structures.
\subsubsection{Temporal Module}
To extend the model's capability to the temporal domain, we design a plug-and-play temporal module that is seamlessly inserted after each spatial module. This interleaved architecture ensures that the network captures temporal evolution hierarchically alongside spatial features. Distinct from the spatial module, the temporal module replaces convolutional layers with linear layers and computes self-attention across different frames rather than spatial positions, as shown on the right of Fig.~\ref{fig:block}. In practice, this is achieved by reshaping the spatial output into a tensor of shape $\mathbb{R}^{B'' \times T \times C}$, where $B'' = B \times H \times W$ represents the flattened spatial dimensions, enabling the network to capture temporal dependencies and enforce motion consistency throughout the sequence.
\subsubsection{Training}
Our spatio-temporal diffusion is the framework for sequential back-view normal estimation and 2D-to-3D mapping. During training, we decouple spatial and temporal learning. First, the spatial module is pre-trained on single frames in \dmapS{}, allowing it to directly inherit the learned spatial priors. Then, we freeze the spatial module and train only the temporal module, allowing it to focus on learning temporal motion priors. The training objective minimizes the noise prediction error
\begin{equation}
\mathcal{L}_{ST} = \mathbb{E}_{\boldsymbol{\epsilon}, t} \left[ || \boldsymbol{\epsilon} - \boldsymbol{\epsilon}_{\boldsymbol{\theta}}(\mathbf{x}_t, \mathbf{c}, t) ||^2_2 \right] \; ,
\end{equation}
where $\boldsymbol{\epsilon} \sim \mathcal{N}(\mathbf{0}, \mathbf{I})$ denotes Gaussian noise and $t$ is the diffusion timestep. The conditioning input $\mathbf{c}$ is task-specific.

For spatio-temporal~(ST) normal diffusion $\boldsymbol{\epsilon}_{\boldsymbol{\theta}}^n$, we set
\[
\mathbf{c}_n = \left[\bN^F, \bS, \bD^b\right],
\]
where $\bN^F = \{\bn_F^{t}\}_{t=1}^{T}$ denotes the sequence of front-view garment normal maps, $\bS = (\bS_F, \bS_B)$ represents the front and back body part segmentations with $\bS_{F/B} = \{\bs_{F/B}^{t}\}_{t=1}^{T}$, and $\bD^b = (\bD^b_F, \bD^b_B)$ are the corresponding body depth images with $\bD^b_{F/B} = \{\bd_{F/B}^{b,t}\}_{t=1}^{T}$. The model predicts the back-view normal sequence $\bN^B = \{\bn_B^{t}\}_{t=1}^{T}$.

For ST mapping diffusion $\boldsymbol{\epsilon}_{\boldsymbol{\theta}}^m$, the conditioning input is
\[
\mathbf{c}_m = \left[\bN^F, \bN^B, \bS, \bD^b\right],
\]
from which the model jointly predicts the garment UV-coordinate images $\bC = (\bC_F, \bC_B)$ and the garment depth images $\bD^g = (\bD_F^g, \bD_B^g)$ as shown in the top-right of Fig.~\ref{fig:pipeline}.

\subsection{Test-Time Guidance for Long Sequences}
\subsubsection{Guidance in ST Normal Diffusion}
\label{sec:st_n_diff}
ST normal diffusion aims to estimate the sequence of back-view normal maps $\bN^B$, which is particularly challenging. The back view is entirely unobserved in monocular videos, requiring the model to generate geometry that is not only spatially consistent with the visible front-view normals, but also temporally coherent over long video sequences.
In practice, processing an entire video sequence at once is computationally infeasible. We therefore divide long videos into shorter subsequences, which inevitably introduces temporal discontinuities at subsequence boundaries. To address this issue, we introduce training-free temporal guidance during inference, which enforces temporal consistency both across adjacent subsequences and within each subsequence. This guidance enables stable and smooth back-view normal estimation over long video sequences without modifying the trained diffusion model.
For across-subsequence consistency, we enforce the overlapping regions of consecutive subsequences to be identical at each denoising time step
\begin{equation}
\mathcal{G}_{across} = \left|\left| \bN'^{B}_{(T-N+1):T} - \hat{\bN}^{B}_{1:N} \right|\right|^2_2 \; ,
\end{equation}
where $N$ is the overlapping length between adjacent subsequences, $\bN'^{B}$ denotes the back-view normal maps generated for the previous clip, and $\hat{\bN}^{B} = \hat{\mathbf{x}}_{0|t}$ denotes the posterior mean (\ie the estimated clean back-view normal maps) for current clip at time step $t$.
In addition, to enforce consistency within each subsequence, we introduce velocity and acceleration losses to transfer the temporal information from the overlapping part $\hat{\bN}^{B}_{1:N}$ to the remaining part $\hat{\bN}^{B}_{N+1:T}$. They are given by
\begin{equation}
\mathcal{G}_{within} = \mathcal{G}_{vel} + \mathcal{G}_{acc},
\end{equation}
\begin{equation}
    \mathcal{G}_{{vel}} = \frac{1}{T-1} \sum_{f=2}^{T} \left\| \hat{\bN}^{B}_f - \hat{\bN}^{B}_{f-1} \right\|_2^2, 
\end{equation}
\begin{equation}
    \mathcal{G}_{{acc}} = \frac{1}{T-2} \sum_{f=3}^{T} \left\| \hat{\bN}^{B}_f - 2\hat{\bN}^{B}_{f-1} + \hat{\bN}^{B}_{f-2} \right\|_2^2.
\end{equation}
By using $\mathcal{G}_{temporal} = \lambda_{a}\mathcal{G}_{across} + \lambda_{w}\mathcal{G}_{within}$ as the manifold guidance~\cite{Chung22a}, the sample $\mathbf{x}_{t-1}$ is updated toward improved temporal consistency
\begin{equation}
    \mathbf{x}_{t-1} = \underbrace{\text{DDPM}(\mathbf{x}_t, \boldsymbol{\epsilon}^n_{\boldsymbol{\theta}}(\mathbf{x}_t, \mathbf{c}_n, t))}_{\text{sampling}} - \underbrace{\gamma\nabla_{\mathbf{x}_t}\mathcal{G}_{temporal}}_{\text{temporal guidance}},
\end{equation}
where $\lambda_a$ and $\lambda_w$ are weighting coefficients, and $\gamma$ is the guidance strength parameter.

\subsubsection{Guidance in ST Mapping Diffusion}
\label{sec:st_m_diff}
During the transformation from 2D image space to UV and 3D spaces in ST mapping diffusion, it is crucial to both preserve spatial details and enforce temporal consistency across frames. To this end, we introduce multiple spatial and temporal guidance terms during denoising:
\begin{itemize}
\item  \textit{Depth-to-normal} guidance to align geometric details. We derive a map of normals $\tilde{\mathbf{N}}$ from the posterior mean depth map $\hat{\bD}^g$ estimated at time step $t$, by computing its partial derivatives with respect to the $x$ and $y$ directions. We compute it as
\begin{equation}
\tilde{\mathbf{N}}(x, y) = \frac{(-\bD_x, -\bD_y, 1)^\top}{\sqrt{\bD_x^2 + \bD_y^2 + 1}},
\end{equation} 
\begin{equation}
\bD_x = s_x\frac{\partial \hat{\bD}^g}{\partial x}\;, \quad
\bD_y = s_y\frac{\partial \hat{\bD} ^g}{\partial y}\;,
\end{equation}
where $s_x$ and $s_y$ are factors that match the gradients to the depth scale. We take depth-to-normal guidance loss to be  $ \mathcal{G}_{D2N} = 1-cos(\mathbf{N}, \tilde{\mathbf{N}})$, where $\mathbf{N}$ is the input normal map. Minimizing this term encourages $\tilde{\mathbf{N}}$ to match $\mathbf{N}$.
\item \textit{Interpenetration-aware} guidance to encourage the garment surface to remain outside the body.  We define the guidance loss as 
\begin{small}
\begin{equation}
\mathcal{G}_{inter} = \text{ReLU}\left(\bD^b_F - \hat{\bD}^g_F\right) + \text{ReLU}\left(\hat{\bD}^g_B - \bD^b_B\right) \; ,
\end{equation}
\end{small}
\hspace{-0.4em}where $\bD^b_F$ and $\bD^b_B$ are the front and back body depths, respectively. $\hat{\bD}^g_F$ and $\hat{\bD}^g_B$ are the front and back garment depths derived by the posterior mean at time step $t$. Minimizing this term prevents body-garment interpenetrations.
\item \textit{Temporal-consistency} guidance to enforce both across- and within-subsequence consistency.
We use the same guidance function $\mathcal{G}_{temporal}$ as  in Sec.~\ref{sec:st_n_diff}. 
\end{itemize}
The gradients of these guidance terms are injected into the denoising process by writing 
\begin{equation}
\begin{aligned}
    \mathbf{x}_{t-1} & =  
    \text{DDPM}\big(\mathbf{x}_t,\,
    \boldsymbol{\epsilon}^m_{\boldsymbol{\theta}}(\mathbf{x}_t, \mathbf{c}_m, t)\big) 
     -\gamma_n \nabla_{\mathbf{x}_t}\mathcal{G}_{D2N} \\
    &  -\gamma_i \nabla_{\mathbf{x}_t}\mathcal{G}_{inter}
      -\gamma_t \nabla_{\mathbf{x}_t}\mathcal{G}_{temporal}.
\end{aligned}
\end{equation}
where $\gamma_n$, $\gamma_i$, and $\gamma_t$ denote the strengths of these guidance, respectively.
As shown in the experiment section, all three guidance terms are essential to jointly achieve spatial accuracy, temporal consistency, and physical plausibility for 4D garment reconstruction.

Finally, we map the predicted $\hat{\bx}_0 = [\bC, \bD^g]$ to the UV space to obtain a partial positional map $\tilde{\bU}$ and a binary mask $\tilde{\bM}$, as shown in the bottom-right of Fig.~\ref{fig:pipeline}. These maps are then used in the next section to recover the full garment deformations.

\subsection{Projection-Based Constraint for Deformation Modeling}
\label{sec:inpaint_diffusion}
To model the completed garment deformation while maximally preserving the established observation $\tilde{\bU}$, we introduce a projection-based constraint, DDPM$_p$, inspired by~\cite{Wang22c}. This constraint enforces strong alignment between the generation process and the incomplete positional map $\tilde{\bU}$. We apply DDPM$_p$ on the deformation prior $\boldsymbol{\epsilon}^d_{\boldsymbol{\theta}}$ of DISP and enforce temporal smoothness as
\begin{equation}\label{eq:inpaint}
\begin{aligned}
    \mathbf{x}_{t-1} =\; &
    \text{DDPM}_p\!\left(
        \mathbf{x}_t,\,
        \tilde{\bU},\,
        \tilde{\bM},\,
        \boldsymbol{\epsilon}^d_{\boldsymbol{\theta}}(\mathbf{x}_t, t)
    \right) \\
     &- \gamma_v \nabla_{\mathbf{x}_t}\mathcal{G}_{vel}^d
      - \gamma_a \nabla_{\mathbf{x}_t}\mathcal{G}_{acc}^d \; , 
\end{aligned}
\end{equation}
where $\gamma_v$ and $\gamma_a$ are the guidance scales. $\text{DDPM}_p$ projects the estimate into observed and unobserved components using range-null space decomposition. To denoise  only on the unobserved part, we write
\begin{equation}\label{eq:ddnm}
\begin{aligned}
& \text{DDPM}_p(\mathbf{x}_t, \tilde{\bU}, \tilde{\bM}, \boldsymbol{\epsilon}^d_{\boldsymbol{\theta}}(\mathbf{x}_t, t))
= \; \frac{\sqrt{\bar{\alpha}_{t-1}} \beta_t}{1 - \bar{\alpha}_t} \hat{\mathbf{x}}_{0 \mid t} \\
& + \frac{\sqrt{\alpha_t} (1 - \bar{\alpha}_{t-1})}{1 - \bar{\alpha}_t} \mathbf{x}_t 
  + \sigma_t \boldsymbol{\epsilon}, \quad 
    \boldsymbol{\epsilon} \sim \mathcal{N}(0, \mathbf{I}) .
\end{aligned}
\end{equation}
\begin{equation}
\label{eq:x_0t}
\hat{\mathbf{x}}_{0 \mid t} = \tilde{\bM}^{\dagger} \tilde{\bU} + \left( \mathbf{I} - \tilde{\bM}^{\dagger} \tilde{\bM} \right) \mathbf{x}_{0 \mid t} \; ,
\end{equation}
where $\tilde{\bM}^{\dagger}$ is the pseudo-inverse of {the observation mask} $\tilde{\bM}$, $\mathbf{x}_{0 \mid t}$ is the estimated $\mathbf{x}_0$ at timestep $t$ following DDPM~\cite{Ho20a}. Intuitively, $\tilde{\bM}^{\dagger}$ projects $\mathbf{x}_{0 \mid t}$ to the range-space of $\tilde{\bM}$, satisfying the observation $\tilde{\bM} \hat{\mathbf{x}}_{0 \mid t} \equiv \tilde{\bU}$, while the $(\mathbf{I} - \tilde{\bM}^{\dagger} \tilde{\bM})$ operator projects $\mathbf{x}_{0 \mid t}$ to the null-space of $\tilde{\bM}$, which {does not affect the observation but determines whether $\hat{\mathbf{x}}_{0 \mid t}$ follows the prior distribution.} This projection-based constraint is crucial here, as it generates positional maps {$\hat{\bU} = \bx_0$} by maximally aligning the extracted observations $\tilde{\bU}$ and satisfies the learned diffusion priors that maintain harmony between the observed and unobserved regions,
resulting in a plausible and complete reconstruction. More detailed derivations and explanations can be found in Sec.~B of the Supplementary Material.

As in Sec.~\ref{sec:st_n_diff}, we introduce velocity and acceleration guidance $\mathcal{G}^d_{vel}$ and $\mathcal{G}^d_{acc}$ to ensure temporal smoothness. They are computed from the current frame and previous frames {$\hat{\bU}_{f-1}$/$\hat{\bU}_{f-2}$} as
\begin{equation}
    \mathcal{G}^d_{vel} = || \mathbf{x}_{0 \mid t} - {\hat{\bU}}_{f-1} ||_2^2, 
\end{equation}
\begin{equation}
    \mathcal{G}^d_{acc} = || \mathbf{x}_{0 \mid t} - 2\hat{\bU}_{f-1} + \hat{\bU}_{f-2} ||_2^2 \; .
\end{equation}
The update of Eq.~\ref{eq:inpaint} yields a sequence of complete positional maps, from which we can recover the mesh sequence $\mathbf{G}$ as illustrated in the bottom-left of Fig.~\ref{fig:pipeline}. 

\subsection{Refinement}
\label{sec:refine}
To bridge the domain gap between synthetic training data and real-world videos, we adopt the refinement strategy in Sec.~\ref{sec:post} and extend it to account for video dynamics by introducing a temporal optimization term
\begin{equation} 
\mathcal{L}_{{temporal}} = \mathcal{L}_{{vel}} + \mathcal{L}_{{acc}},
\end{equation} 
which penalizes irregular velocity and acceleration, and enforces temporal consistency in the reconstructed sequence.
% !TEX root = ../main.tex
% !TEX spellcheck = en-US

% !TEX root = ../main.tex
% !TEX spellcheck = en-US

\begin{table*}[t]
\centering
\caption{Quantitative comparison with SOTA methods. ``$\dagger$'' denotes models that use refinement. \textbf{Bold}: best; \underline{Underline}: second best.}
\vspace{-2mm}
\setlength\tabcolsep{2.8mm}
\begin{tabular}{l|ccc|ccc|ccc|ccc}
\toprule
\multirow{2}{*}{Method} & \multicolumn{3}{c|}{Skirt} & \multicolumn{3}{c|}{Trousers} & \multicolumn{3}{c|}{Tshirt} & \multicolumn{3}{c}{Jacket}\\
& CD~$\downarrow$ & NC~$\uparrow$ & IoU~$\uparrow$ &  CD~$\downarrow$ & NC~$\uparrow$ & IoU~$\uparrow$ & CD~$\downarrow$ & NC~$\uparrow$ & IoU~$\uparrow$ & CD~$\downarrow$ & NC~$\uparrow$ & IoU~$\uparrow$ \\
\midrule
SMPLicit~\cite{Corona21} & 5.80 & 0.34 & 53.72 & 1.46 & -0.12 & 74.53 & 2.77 & 0.09 & 58.64 & 2.40 & 0.03 & 54.81 \\
ISP~\cite{Li23a} & 4.18  & 0.78 & 68.10  & 1.53 & 0.78 & 75.39 & 1.36 & 0.82 & 68.70 & 1.69 & 0.79 & 71.31 \\
GaRec$\dagger$~\cite{Li24a} & 2.67  & 0.88 & \underline{96.33}
& 1.67 & 0.85 & 92.55 & 1.19 & 0.83 & 87.15 & 1.21 & 0.79 & \underline{91.40} \\
D$^3$-Human$\dagger$~\cite{Chen25a} & 3.64   & 0.78 & 76.42 & 1.80 & 0.79 & 82.49 & 1.79 & 0.79 & 78.51 & 1.71 & 0.76 & 56.60 \\
\midrule
\dmapS{}~(Ours)
& 2.26 & 0.89 & 88.80
& 1.36 & 0.84 & 86.23
& 1.06 & 0.83 & 80.20
& 1.29 & 0.83 & 78.28 \\
\dmapS{}$\dagger$~(Ours)
& 2.17 & 0.91 & 94.80
& \underline{0.85} & \textbf{0.88} & \underline{94.64}
& \underline{0.81} & \underline{0.88} & \underline{93.34}
& \underline{0.97} & \underline{0.83} & 89.85 \\
\dmapD{}~(Ours)
& \underline{1.62} & \textbf{0.93} & 90.93
& 1.21 & 0.83 & 88.72
& 0.92 & 0.86 & 91.95
& 1.11 & 0.85 & 81.75 \\
\dmapD{}$\dagger$~(Ours)  & \textbf{1.54} & \underline{0.92} & \textbf{96.73} & \textbf{0.83} & \underline{0.87} & \textbf{95.41} & \textbf{0.76} & \textbf{0.92} & \textbf{96.46} & \textbf{0.93} & \textbf{0.86} & \textbf{93.47} \\
\bottomrule
\end{tabular}
\label{tab:compare_sota}
\vspace{-2mm}
\end{table*}

% !TEX root = ../top.tex
% !TEX spellcheck = en-US

\begin{figure*}[ht!]
    \centering
    \includegraphics[width=.83\textwidth]{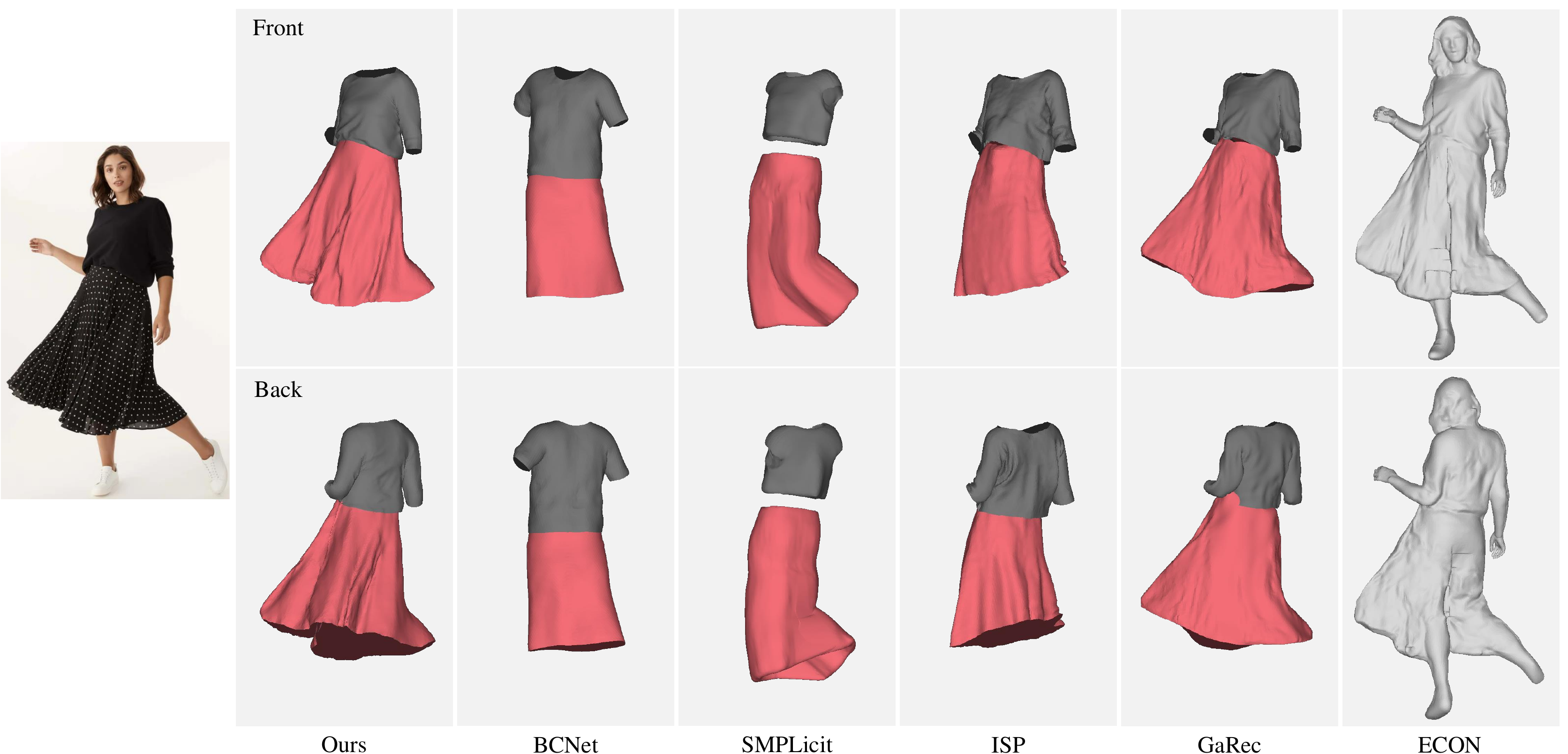}
    \vspace{-2mm}
    \caption{\textbf{Qualitative comparison on an in-the-wild image.} The top and bottom rows show the front and the back of the reconstructions produced by our method DMap, BCNet \cite{Jiang20d}, SMPLicit \cite{Corona21}, ISP \cite{Li23a}, GaRec \cite{Li24a} and ECON \cite{Xiu23}, respectively.}
    \label{fig:compare}
    \vspace{-2mm}
\end{figure*} 
\begin{figure*}[ht]
    \centering
    \includegraphics[width=0.88\linewidth]{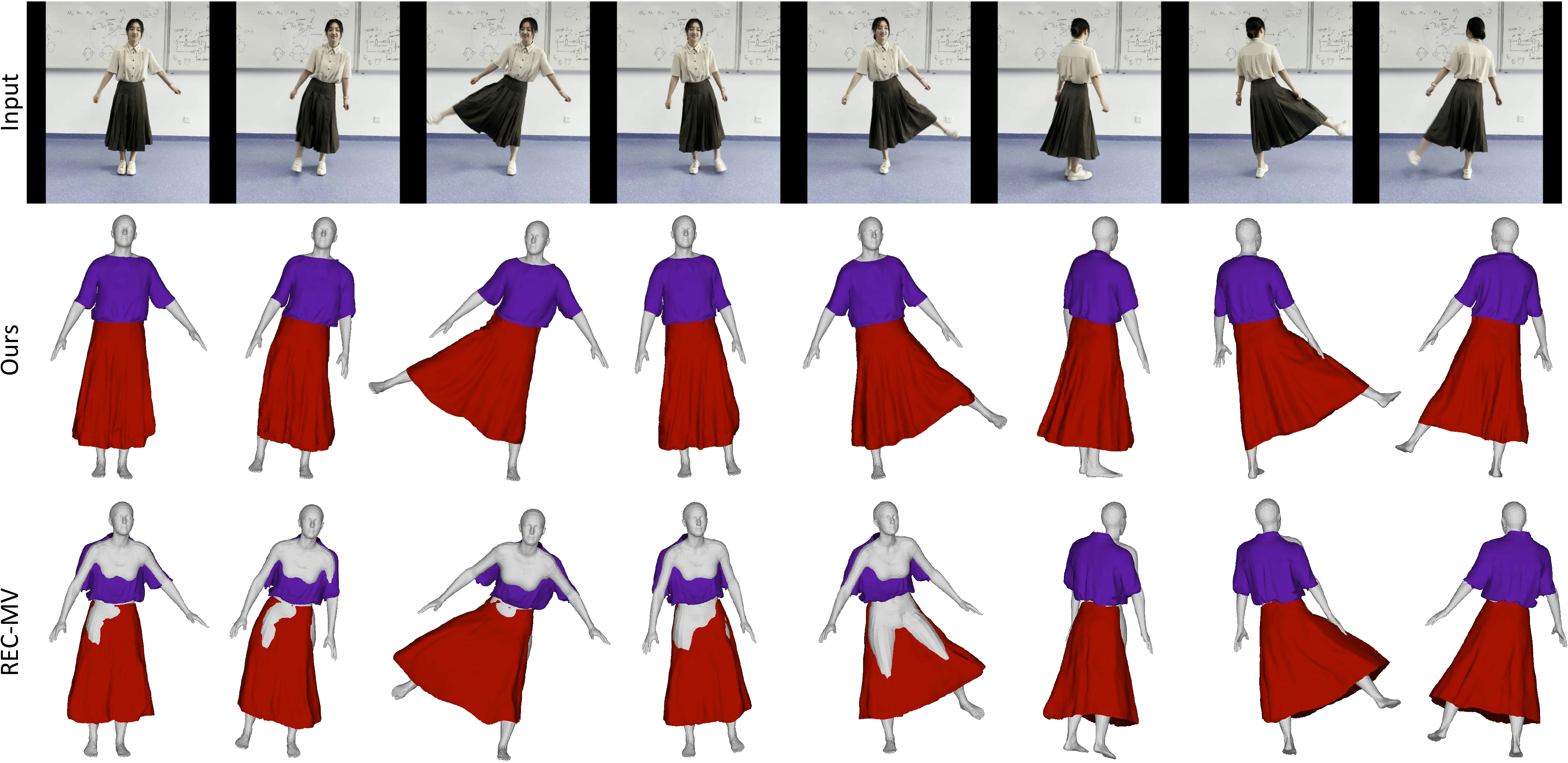}
    \vspace{-2mm}
    \caption{\textbf{Qualitative comparison on a video sequence.} Our method (second row) achieves finer geometric detail, better temporal consistency, and avoids the garment–body collisions frequently observed in REC-MV~\cite{Qiu23a} (bottom row), where the reconstructed garment often penetrates the body surface.}
    \label{fig:comp_vis}
    \vspace{-3mm}
\end{figure*}

\section{Experiments}

\subsection{Dataset and Evaluation Metrics}
\label{sec:data}
\subsubsection{Dataset}
Due to the intricate structure of garments, collecting real 3D data with complete geometry for them is extremely difficult. Instead, we use physics-based simulation to generate garment with realistic deformations in its interaction with the underlying body.
CLOTH3D~\cite{Bertiche20} is a synthetic dataset containing 3D garments simulated on animated SMPL bodies. 
For each clothing category, including shirt, open jacket, skirt, and trousers, we randomly select $33$ garment samples from CLOTH3D, with $30$ for training and $3$ for testing. Please refer to Fig.~K in the Supplementary Material for visualizations of the garment shapes. These samples cover diverse garment shapes, ranging from tight-fitting to loose-fitting garments. To construct complete outfit configurations, each sample is randomly paired with a compatible upper-body or lower-body garment. Therefore, each category contains $33$ garment pairs.
Beyond static garment shape, garment looseness is also reflected in dynamic deformation. For example, skirts may flare outward during spinning, while jackets may swing open due to inertia. Although CLOTH3D provides dynamic garment sequences, each garment is simulated using only a single motion sequence, which mainly consists of relatively simple motions such as walking, running, and jumping. To obtain richer garment dynamics, we re-simulate each garment-body pair using motion data from the AMASS~\cite{Mahmood19} dataset, including dancing, spinning, hopping, and other motions with large body movements.
Specifically, we concatenate $32$ AMASS motion sequences with interpolation for smooth transitions and simulate each garment-body pair on the resulting motion sequence, as shown in Supplementary Material Fig.~L. After downsampling by a factor of $4$, each garment-body sequence contains $8,000$ frames at $30$ fps. In total, across the four garment categories, we have $4 \times 33 \times 8{,}000 = 1{,}056{,}000$ frames.
The motion sequences are generated using Blender~\cite{Blender} and Marvelous Designer~\cite{MarvelousDesigner}. Additional pins are manually set for open jackets, skirts, and trousers to prevent sliding during simulation. For each body sample in a sequence, we randomly rotate it and render the front and back body-part segmentation and depth images. The corresponding garment sample is rotated by the same angle, and its front and back normal and depth images are rendered. Its front and back UV coordinate images are generated using the UV parameterization of ISP. For \dmapS{}, we subsample the training data by selecting one frame every five frames, whereas for \dmapD{}, we use consecutive frames for training without subsampling.

\subsubsection{Evaluation Metrics}
To evaluate the quality of garment reconstruction, we use the Chamfer Distance (CD) and the Normal Consistency (NC) between the ground truth and the recovered garment mesh, and the Intersection over Union (IoU) between the ground truth mask and the rendered mask of reconstructed garment mesh. Temporal consistency is measured by vertex acceleration (mm/frame$^2$) between consecutive frames. The quantitative comparison is conducted on the synthetic test set, while qualitative evaluation is performed on in-the-wild images and videos that include loose garments, diverse motions, and occlusions.

\begin{figure*}[ht!]
    \centering
    \includegraphics[width=1.\textwidth]{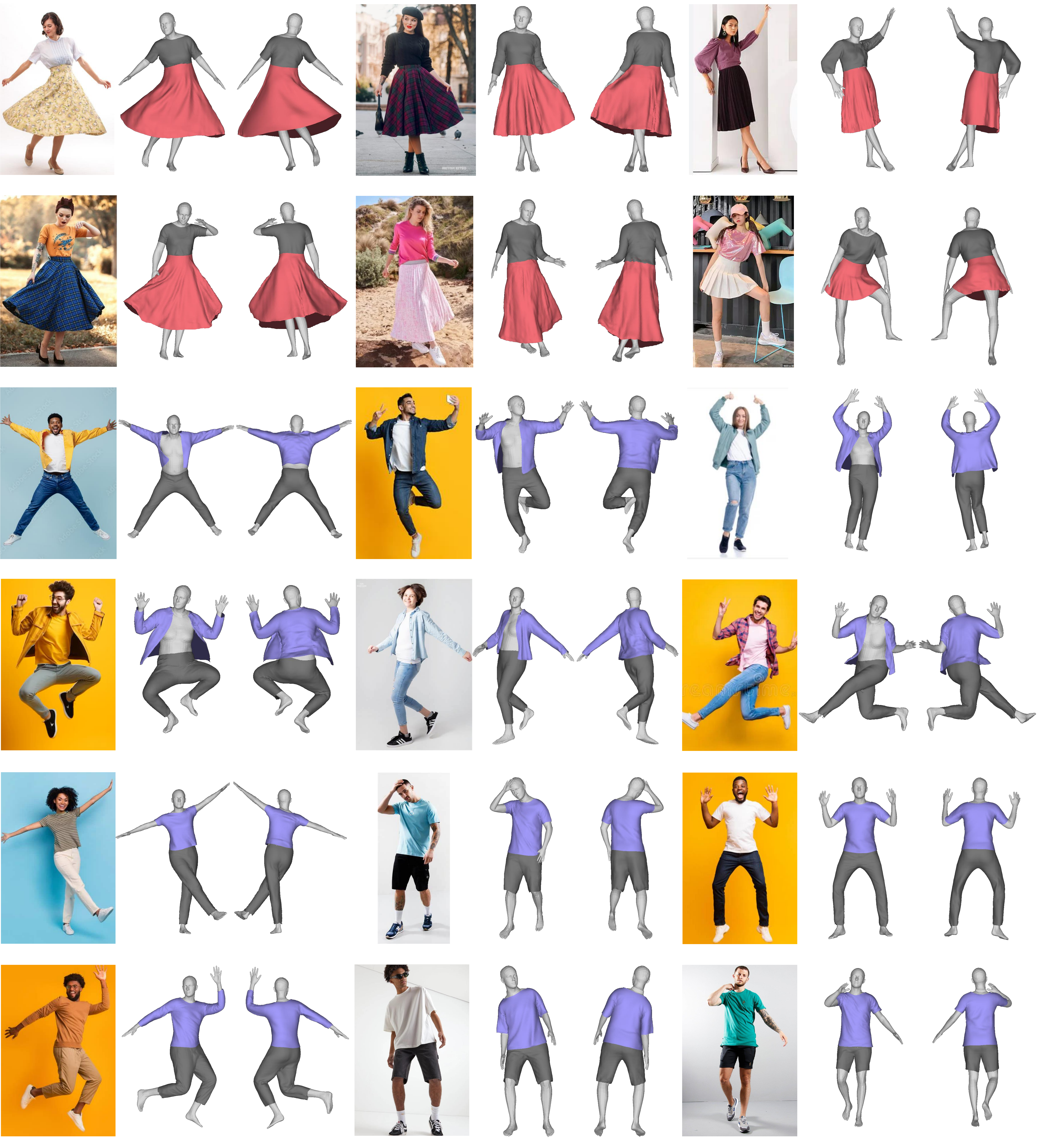}
    \vspace{-6mm}
    \caption{\textbf{Reconstruction for in-the-wild images.}  Our method can recover realistic 3D models for diverse garments in different shapes and deformations.}
    \label{fig:fitting}
    \vspace{-2mm}
\end{figure*} 
\vspace{-2mm}
\begin{figure*}[p]
    \centering
    \includegraphics[width=0.90\linewidth]{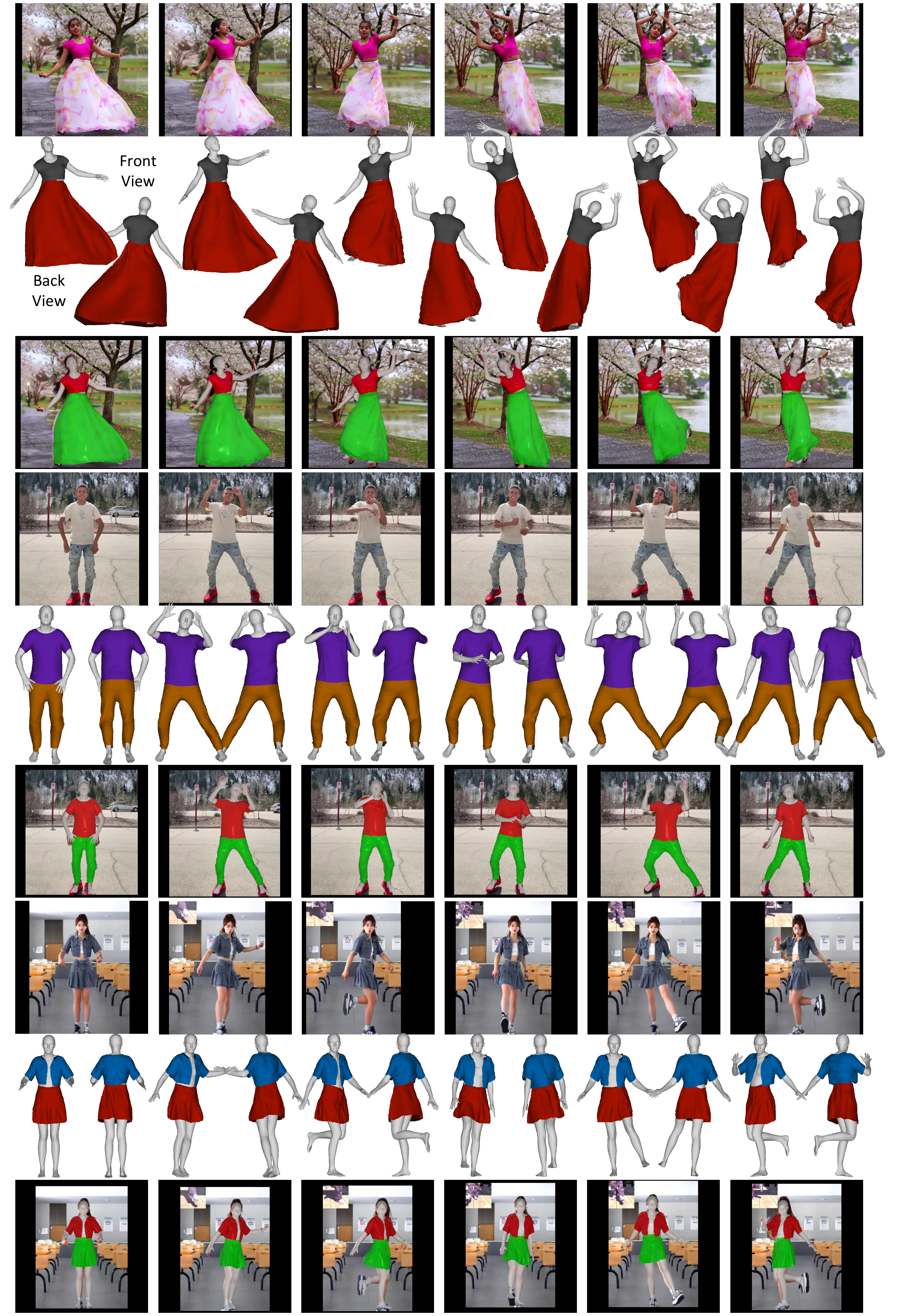}
    \vspace{-3mm}
    \caption{\textbf{Qualitative results on real-world videos.} For each sample, we show the input, the front- and back-view reconstructions, and the 2D renderings.}
    \label{fig:supp_self_vis}
    \vspace{-3mm}
\end{figure*}

\subsection{Reconstruction Results}
\subsubsection{Quantitative Comparison}
In Table \ref{tab:compare_sota}, we report the quantitative comparison between our methods and state-of-the-art approaches.
SMPLicit~\cite{Corona21}, ISP~\cite{Li23a}, and GaRec~\cite{Li24a} are image-based methods, whereas D$^{3}$-Human~\cite{Chen25a} is a video-based approach.
We first evaluate the inferred versions (\dmapS{}, \dmapD{}), obtained directly via diffusion inference. \dmapS{} already outperforms existing methods, while \dmapD{} achieves even higher accuracy by considering the spatio-temporal consistency jointly. We also present results for the refined versions (\dmapS{}$^\dagger$, \dmapD{}$^\dagger$), where an additional refinement stage (Sec.~\ref{sec:fitting_dmap} and~\ref{sec:refine}) is applied to further enhance the reconstruction quality.
Notably, for loose-fitting skirts, \dmapD{}$^\dagger$ delivers the best performance by a substantial margin, demonstrating the effectiveness of our spatio-temporal diffusion and refinement in capturing large garment deformations. Please see Sec. A of the Supplementary Material and the supplementary video for dynamic comparisons on video sequences.

\subsubsection{Qualitative Comparison}
\label{sec:qualitative_comp}
Fig. \ref{fig:compare} shows a qualitative comparison on an in-the-wild image for the static reconstruction. Methods such as BCNet\cite{Jiang20d}, SMPLicit, and ISP rely on body-driven skinning and therefore keep garments tightly attached to the body. ECON~\cite{Xiu23} and GaRec can recover garments that stand away from the body surface, but ECON merges the body and garment into a single watertight mesh, and GaRec ~\cite{Li24a} often yields flat surfaces with limited fold details. In contrast, our method uses back-normal estimation together with DISP priors to reconstruct garment meshes with realistic wrinkles and high-fidelity details on both the front and back.
Fig. \ref{fig:comp_vis} presents a qualitative comparison on a video sequence with the state-of-the-art video-based method REC-MV~\cite{Qiu23a}. Our method reconstructs garments with richer geometric detail and better temporal consistency, while also avoiding garment-body collisions that are clearly visible in the results of REC-MV.

% !TEX root = ../top.tex
% !TEX spellcheck = en-US

\begin{figure*}[ht!]
    \centering
    \includegraphics[width=.95\textwidth]{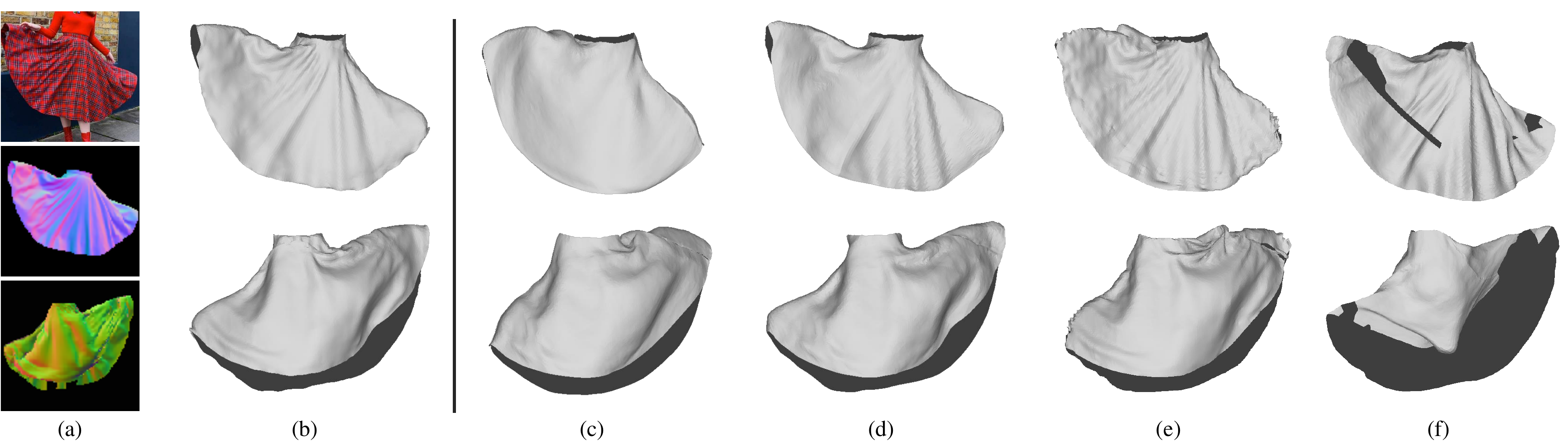}
    \caption{\textbf{Ablation of fitting strategy}. (a) The input image and its normal estimations for the front and back. (b) Our full reconstruction. (c) Reconstruction without post-refinement. (d) Reconstruction refined by optimizing only the neural displacement field. (e) Reconstruction refined by optimizing only the vertex positions. (f) Reconstruction using only the front normal.}
    \label{fig:ablation}
    \vspace{-0.25cm}
\end{figure*}

\subsubsection{Reconstructions on real-world images and videos}

Fig. \ref{fig:fitting} provides more results of our method applied to in-the-wild images, demonstrating its ability to produce realistic 3D meshes with fine details for both tight-fitting and loose-fitting garments.
Fig.~\ref{fig:supp_self_vis} provides additional qualitative results on real-world video sequences with various garment types, including T-shirts, skirts, trousers, and jackets. These results demonstrate that our method can handle a wide range of garments and human poses while ensuring physically plausible interactions between them. Notably, the rendered images closely align with the input frames, and the geometric details are consistent with the images, indicating that our method effectively extracts and preserves visual observations. Please refer to the supplementary video for dynamic results.

\begin{table}[t]
\centering
\setlength{\tabcolsep}{0.9mm}
\caption{Inference time. ``$\dagger$'' denotes models that use refinement.}
\vspace{-2mm}
\re{
\begin{tabular}{c l c c c c}
\toprule
  & Method & Time/Frame & Hardware & VRAM Usage & I/O\\
\midrule
\multirow{5}{*}{\rotatebox{90}{Image-based}} 
& SMPLicit [9] & 8 min & A100-80G & 3.8 GB & 1/1\\
& ISP [14] & 19 min & A100-80G & 2.3 GB & 1/1 \\
& GaRec$\dagger$ [15] & \textbf{1 min} & A100-80G & 12.0 GB & 1/1\\
& \dmapS{} & 7 min & A100-80G & 6.8 GB & 1/1 \\
& \dmapS{}$\dagger$ & 24 min & A100-80G & 6.8 GB & 1/1\\
\midrule
\multirow{4}{*}{\rotatebox{90}{Video-based}}
& REC-MV$\dagger$ [16] & 12 min & A100-80G & 19.6 GB &1/1\\
& D$^{3}$-Human$\dagger$ [17] & 13 min & RTX 4090-24G & 15.2 GB & 1/1\\
& \dmapD{} & \underline{3 min} & A100-80G & 74.1 GB & 10/10\\
& \dmapD{}$\dagger$ & 7 min & A100-80G & 74.1 GB& 10/10\\
\bottomrule
\end{tabular}
}
\label{tab:inference_time}
\vspace{-3mm}
\end{table}

\subsubsection{Inference Time}

Table~\ref{tab:inference_time} presents the runtime comparison, including the inference hardware and GPU memory usage. Inference is conducted on a single NVIDIA A100 GPU with 80GB VRAM. D$^{3}$-Human~\cite{Chen25a} is evaluated on an NVIDIA RTX 4090 GPU because it requires on-screen rendering and cannot be directly run on our headless A100 cluster.
\dmapD{} achieves lower computational cost than most competing methods, especially among video-based approaches. Unlike REC-MV~\cite{Qiu23a} and D$^{3}$-Human, which process frames independently, \dmapD{} processes $10$ frames jointly and outputs $10$ reconstructions in one inference pass. Although this leads to higher VRAM usage, the peak memory is bounded by the fixed subsequence length ($T=10$), rather than the full video length, enabling long-video reconstruction with fixed peak memory.
\dmapD{} takes $3$ minutes per frame for base inference and $7$ minutes with the optional refinement stage on a single A100-80G GPU. As shown in Table~\ref{tab:compare_sota}, the inference-only result already achieves competitive performance, providing a strong initialization that reduces the optimization burden during refinement. Consequently, the refined version still achieves a lower runtime than most baselines. This design offers a practical balance between efficiency and accuracy: the inference-only mode provides rapid reconstruction, whereas the refinement stage can be enabled when maximal geometric fidelity is desired.

\subsubsection{Temporal Consistency}
\begin{figure}[t]
    \centering
    \includegraphics[width=0.8\linewidth]{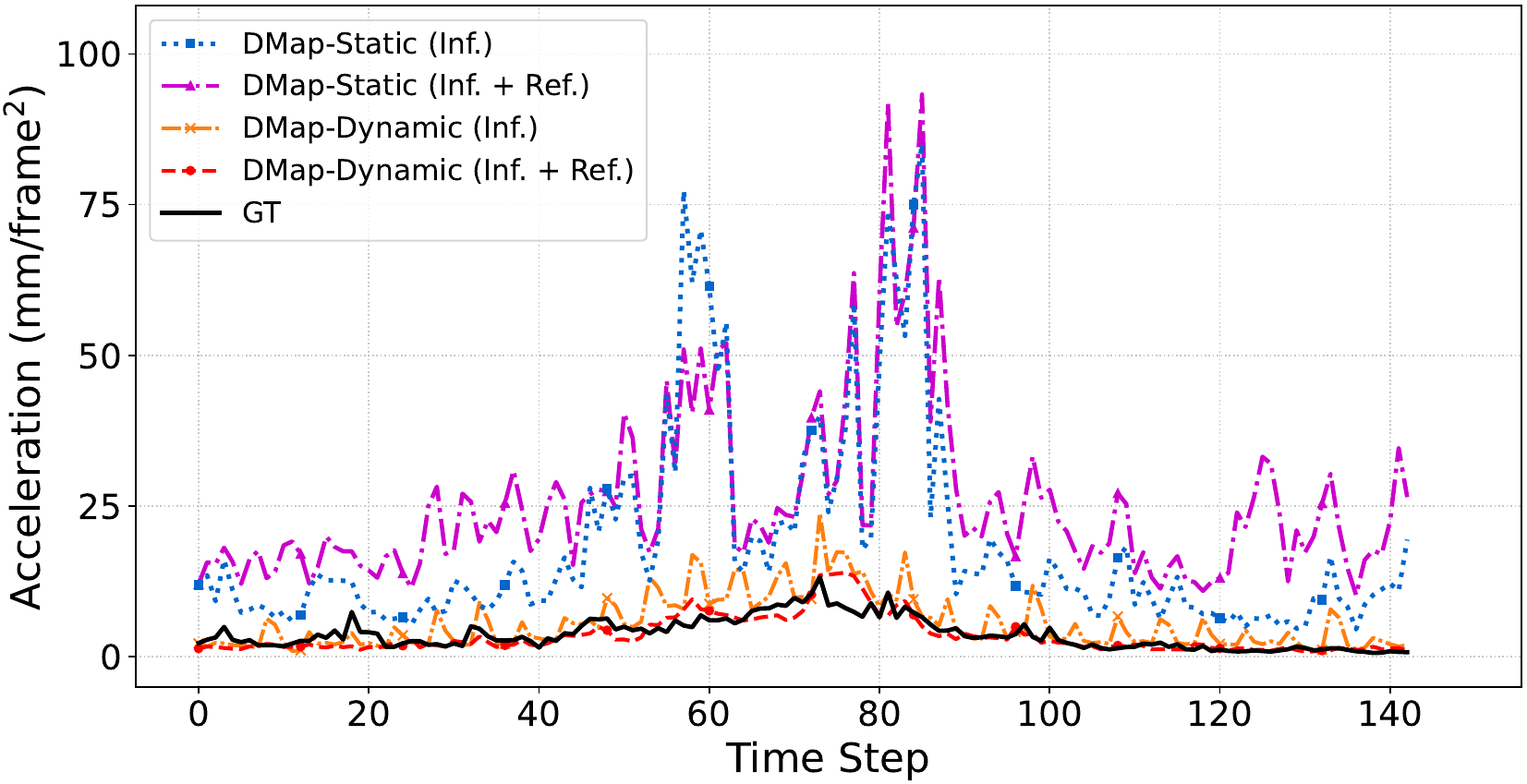}
    
    {\scriptsize \re{(a) Acceleration comparison on the skirt sequence.}}
    
    \vspace{0.5em}
    
    \includegraphics[width=0.8\linewidth]{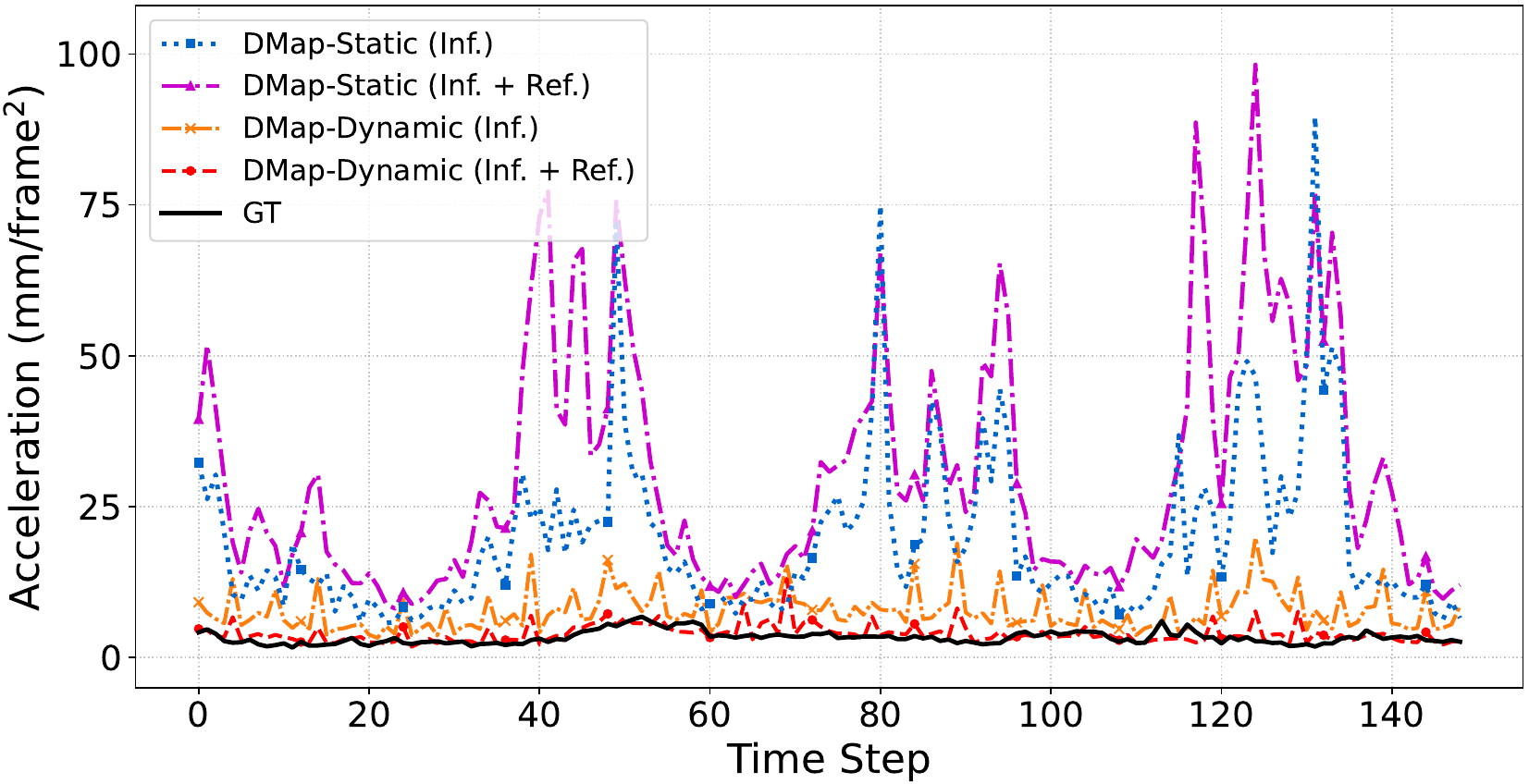}
    
    {\scriptsize \re{(b) Acceleration comparison on the jacket sequence.}}
    
    \vspace{-0.5em}
    \caption{\re{\textbf{Acceleration comparison.} ``Inf.'': inference; ``Ref.'': refinement.}}
    \label{fig:acceleration_curves}
    \vspace{-2mm}
\end{figure}

Although the refined \dmapS{} achieves lower per-frame errors than the non-refined \dmapD{} in several categories in Table~\ref{tab:compare_sota}, these metrics do not account for temporal consistency. To evaluate this, we compare the vertex acceleration curves of \dmapS{} and \dmapD{} on the test-set skirt and jacket sequences.
As shown in Fig.~\ref{fig:acceleration_curves}, both \dmapS{} variants exhibit larger acceleration values and stronger fluctuations, indicating visible temporal jitter. This is because \dmapS{} reconstructs and refines each frame independently: refinement improves frame-wise alignment but does not enforce temporal coherence. In contrast, \dmapD{} introduces spatio-temporal diffusion and temporal guidance, producing smoother results with lower acceleration values closer to the ground truth, as reported in Table~\ref{tab:acceleration_statistics}. Dynamic visualizations in the supplementary video further show that \dmapD{} produces smoother garment motion than the refined \dmapS{}.

% ==========================================
% ==========================================
\begin{table}[t]
    \centering
    \caption{\re{Quantitative acceleration statistics.}}
    \vspace{-2mm}
    \resizebox{\linewidth}{!}{
    \re{
    \begin{tabular}{l|rrrr}
        \toprule
        Method & Mean & Median & Min & Max \\
        \midrule
        DMap-Static (Inf.) & 18.232 & 11.994 & 4.546 & 85.132 \\
        DMap-Static (Inf. + Ref.) & 25.470 & 21.355 & 10.219 & 93.281 \\
        DMap-Dynamic (Inf.) & 5.588 & 3.939 & 0.945 & 23.843 \\
        DMap-Dynamic (Inf. + Ref.) & 3.497 & 2.122 & 0.869 & 13.967 \\
        GT & 3.724 & 2.846 & 0.622 & 13.193 \\
        \bottomrule
    \end{tabular}
    }
    }
    
    \vspace{0.4em}
    {\scriptsize \re{(a) Quantitative results on the skirt sequence.}}
    
    \vspace{0.8em}
    
    \resizebox{\linewidth}{!}{
    \re{
    \begin{tabular}{l|rrrr}
        \toprule
        Method & Mean & Median & Min & Max \\
        \midrule
        DMap-Static (Inf.) & 19.461 & 14.434 & 5.727 & 89.611 \\
        DMap-Static (Inf. + Ref.) & 30.359 & 24.567 & 8.125 & 98.236 \\
        DMap-Dynamic (Inf.) & 7.748 & 6.751 & 3.235 & 19.874 \\
        DMap-Dynamic (Inf. + Ref.) & 3.926 & 3.555 & 1.804 & 13.047 \\
        GT & 3.304 & 3.174 & 1.651 & 6.760 \\
        \bottomrule
    \end{tabular}
    }
    }
    
    \vspace{0.4em}
    {\scriptsize \re{(b) Quantitative results on the jacket sequence.}}
    
    \label{tab:acceleration_statistics}
    \vspace{-4mm}
\end{table}

\begin{table*}[t]
\centering
\setlength{\tabcolsep}{3.3mm}
\caption{Ablation study of the proposed spatial guidance in DMap-dynamic on the CLOTH3D dataset.}
\begin{tabular}{l|ccc|ccc|ccc|ccc}
\toprule
\multirow{2}{*}{Method} &
\multicolumn{3}{c|}{Skirt} &
\multicolumn{3}{c|}{Trousers} &
\multicolumn{3}{c|}{Tshirt} &
\multicolumn{3}{c}{Jacket} \\
& CD$\downarrow$ & NC$\uparrow$ & IoU$\uparrow$
& CD$\downarrow$ & NC$\uparrow$ & IoU$\uparrow$
& CD$\downarrow$ & NC$\uparrow$ & IoU$\uparrow$
& CD$\downarrow$ & NC$\uparrow$ & IoU$\uparrow$ \\
\midrule

inference w/o back view
& 2.67 & 0.90 & 83.62
& 1.90 & 0.81 & 79.09
& 1.61 & 0.81 & 78.31
& 2.00 & 0.73 & 70.46 \\

inference w/o $\mathcal{G}_{\mathrm{D2N}}$
& 1.97 & 0.92 & 90.67
& 1.53 & 0.82 & 86.53
& 1.24 & 0.84 & 85.73
& 1.56 & 0.82 & 72.55 \\

inference w/o $\mathcal{G}_{\mathrm{inter}}$
& 1.71 & 0.92 & 90.53
& 1.30 & 0.82 & 86.62
& 1.07 & 0.83 & 88.23
& 1.24 & 0.83 & 77.63 \\
\midrule
\textbf{inference}
& \underline{1.62} & \textbf{0.93} & 90.93
& \underline{1.21} & \textbf{0.83} & \underline{88.72}
& \underline{0.92} & \underline{0.86} & \underline{91.95}
& \underline{1.11} & \underline{0.85} & \underline{81.75} \\

\textbf{inference + refinement}
& \textbf{1.54} & \underline{0.92} & \textbf{96.73}
& \textbf{0.83} & \underline{0.87} & \textbf{95.41}
& \textbf{0.76} & \textbf{0.92} & \textbf{96.46}
& \textbf{0.93} & \textbf{0.86} & \textbf{93.47} \\

\bottomrule
\end{tabular}

\label{tab:ablation_components}
\vspace{-3mm}
\end{table*}

\vspace{-2mm}
\subsection{Ablation Study}
\label{sec:ablation}

\subsubsection{Effectiveness of Fitting Strategy}
\label{sec:ablation_fitting}
Fig. \ref{fig:ablation} presents the ablation study of our fitting method of \dmapS{} introduced in Sec.~\ref{sec:fitting_dmap}. As shown in Fig. \ref{fig:ablation}~(c), the initial reconstruction, without the post-refinement step described in Section \ref{sec:post}, fails to fully align with the input image shown in Fig. \ref{fig:ablation}~(a). Incorporating a neural displacement field to optimize the initial mesh improves reconstruction accuracy, as seen in Fig. \ref{fig:ablation}~(d). Further refinement by directly optimizing vertex positions enhances the wrinkle details, as illustrated in Fig. \ref{fig:ablation}~(b). However, applying post-refinement without first optimizing the neural displacement field (Fig. \ref{fig:ablation}~(e)) struggles to recover an accurate shape, as each vertex is optimized independently, leading to suboptimal results. Finally, Fig. \ref{fig:ablation}~(f) shows the outcome when only the front normal estimation is used throughout the fitting process described in Section \ref{sec:fitting_dmap}. The lack of constraints for the back surface results in unrealistic deformations on the back. The corresponding quantitative results are provided in the Supplementary Material Sec.~D(c).

\begin{figure}[t]
    \centering
    \includegraphics[width=\linewidth]{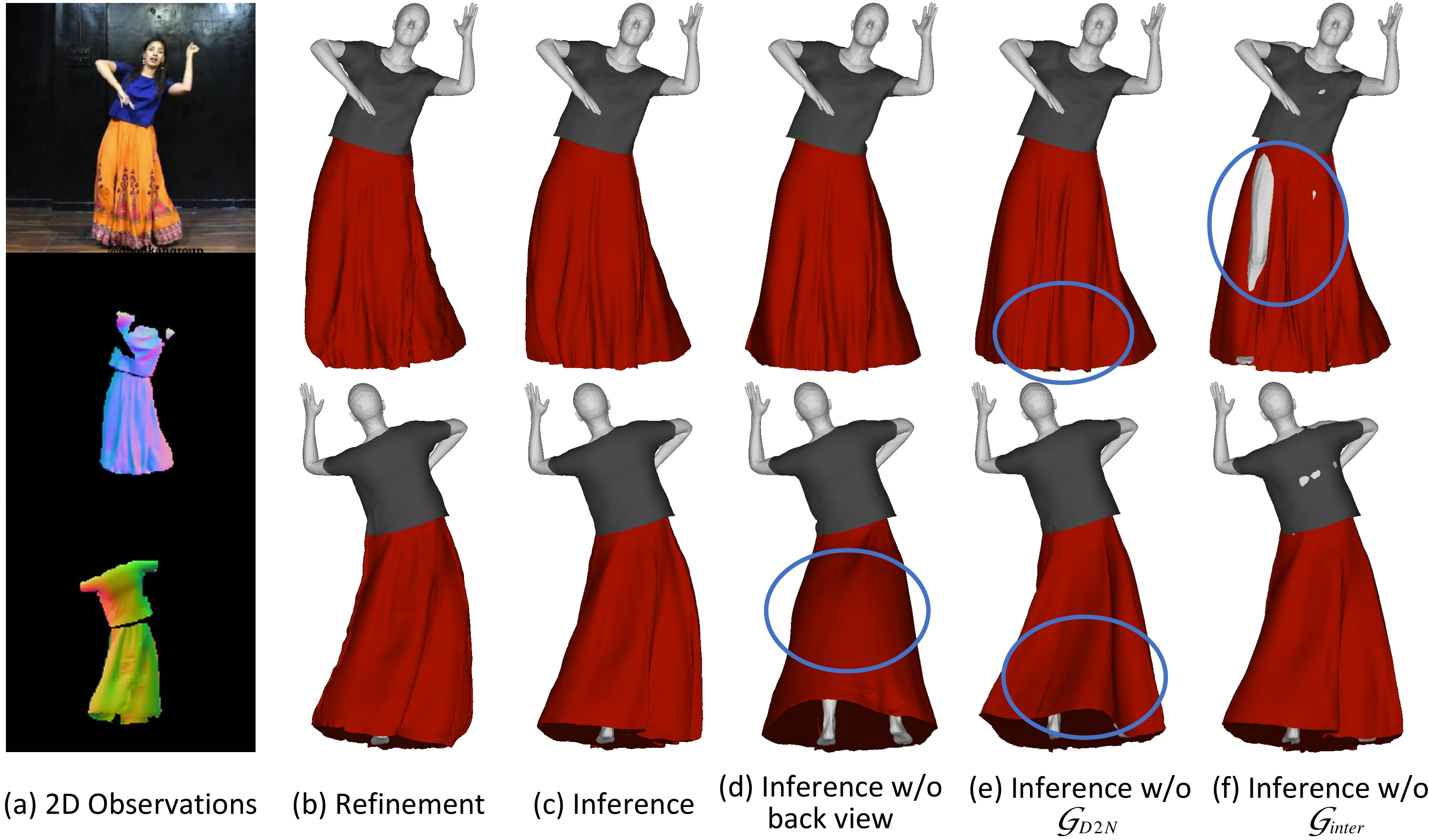}
    \vspace{-5mm}
    \caption{\textbf{Ablation on spatial guidance.} (a) 2D observations, including input image, front normal, and back normal. (b) Reconstruction after refinement. (c) Reconstruction after inference. (d) Inferred result without using generated back-view normals. (e) Inferred result without depth-to-normal guidance $\mathcal{G}_{D2N}$. (f) Inferred result without interpenetration-aware guidance $\mathcal{G}_{inter}$.}
    \label{fig:spatial_ablation}
    \vspace{-2mm}
\end{figure}
\begin{figure}[t]
    \centering
    \begin{minipage}[t]{0.48\linewidth}
        \centering
        \includegraphics[width=\linewidth]{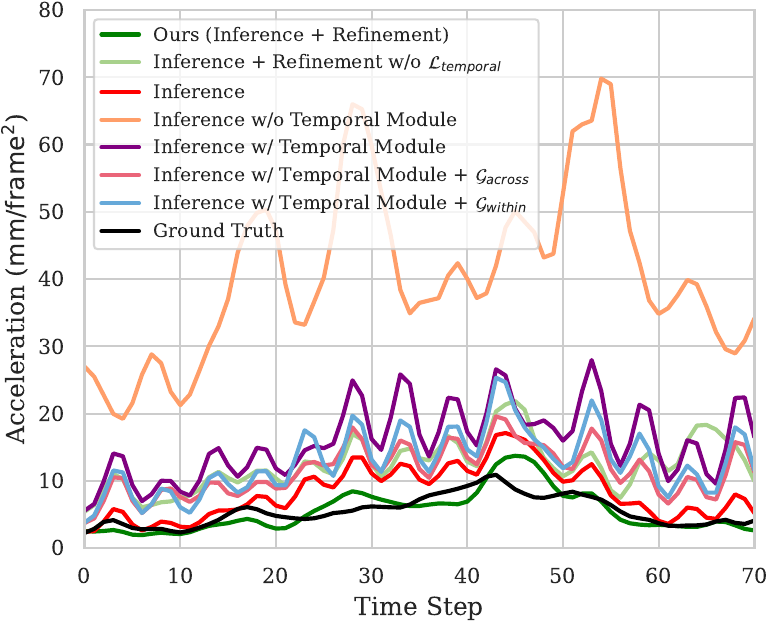}
        \caption{\textbf{Effectiveness of Temporal Components.} Vertex acceleration quantifies temporal consistency.}
        \label{fig:acc}
    \end{minipage}
    \hfill
    \begin{minipage}[t]{0.48\linewidth}
        \centering
        \includegraphics[width=\linewidth]{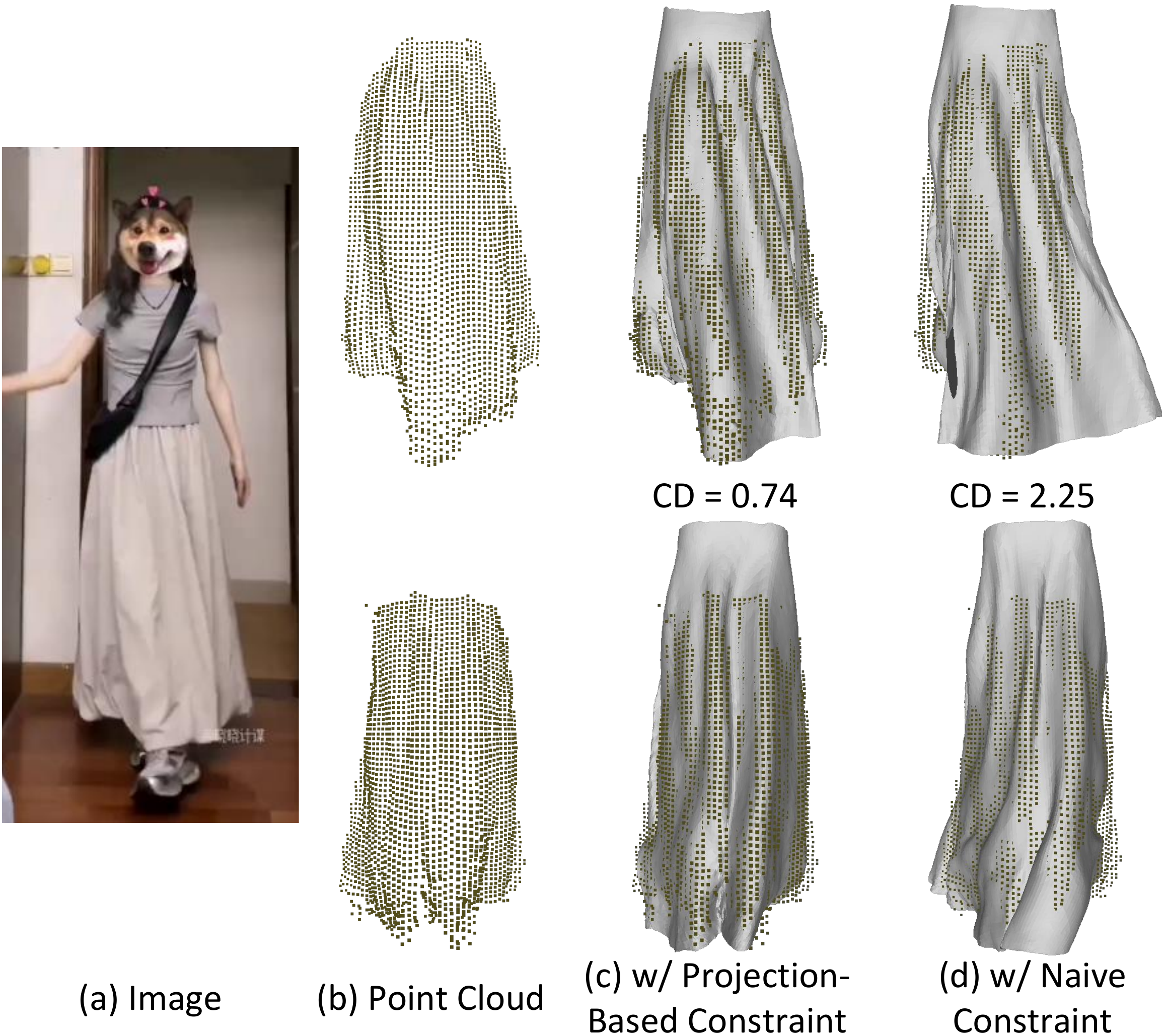}
        \caption{\textbf{Reconstructions using projection-based constraint and gradient-based guidance.}}
        \label{fig:hard_vs_soft}
    \end{minipage}
    \vspace{-2mm}
\end{figure}

\subsubsection{Effectiveness of Spatial Guidance}
Table~\ref{tab:ablation_components} and Fig.~\ref{fig:spatial_ablation} show the ablation study of the spatial guidance terms of \dmapD{} {introduced in Sec.~\ref{sec:st_m_diff}. As shown in Fig.~\ref{fig:spatial_ablation} (d), {without back-view guidance, the diffusion model tends to produce overly smooth garments that lack realistic draping. We hypothesize that, in the absence of both observations and guidance, the diffusion model is only able to generate an averaged garment mesh where all draping has been smoothed out.} The depth-to-normal guidance $\mathcal{G}_{D2N}$ encourages the depth to capture normal details. Fig.~\ref{fig:spatial_ablation} (e) shows that removing this guidance leads to a loss of fine details and misalignment with the normal observations. 
Finally, without the interpenetration-aware guidance $\mathcal{G}_{inter}$, the estimated garment collides with the body, as illustrated in Fig.~\ref{fig:spatial_ablation} (f). 

\subsubsection{Effectiveness of Temporal Components}
We evaluate temporal consistency by measuring vertex acceleration between consecutive frames.
Fig.~\ref{fig:acc} shows the acceleration curves on a synthetic skirt sequence and illustrates the effect of different temporal components: the temporal diffusion module, across-subsequence guidance ($\mathcal{G}_{across}$), within-subsequence guidance ($\mathcal{G}_{within}$), and the temporal loss used during refinement. The black curve denotes ground-truth acceleration.
Without the temporal module (orange), \dmapD{} reduces to an image-based method and exhibits large, frequent accelerations, indicating strong motion jitter. Incorporating the temporal module (purple) substantially reduces jitter, though inconsistencies remain at subsequence boundaries.
Across-subsequence guidance (pink) mitigates these boundary artifacts, and combining $\mathcal{G}_{across}$ with $\mathcal{G}_{within}$ (red) further improves temporal smoothness during inference.
Finally, the comparison between the deep and light green curves highlights the importance of the refinement-stage temporal loss $\mathcal{L}_{temporal}$ for achieving the highest level of temporal coherence.

\subsubsection{Effectiveness of Projection-Based Inpainting}
Fig.~\ref{fig:hard_vs_soft} compares the inpainted reconstructions from the incomplete positional map $\tilde{\bU}$ using our projection-based constraints and {a naive constraints that directly minimizing $|| \tilde{\bM}(\mathbf{x}_{0 \mid t} - \tilde{\bU})||_2^2$}.
For better visualization, we map the positional map to point clouds.
It can be observed that the reconstruction with projection constraint aligns more closely with the sparse input, thereby better preserving the accuracy and temporal consistency established in the previous stages. In contrast, the {naive} constraint provides weaker control over the generation process, which
introduces discrepancies and leads to spatial and temporal inconsistencies.

% !TEX root = ../top.tex
% !TEX spellcheck = en-US

\begin{figure}[t]
	\centering
		\begin{minipage}{.485\textwidth}
        \centering
        \includegraphics[width=.99\textwidth]{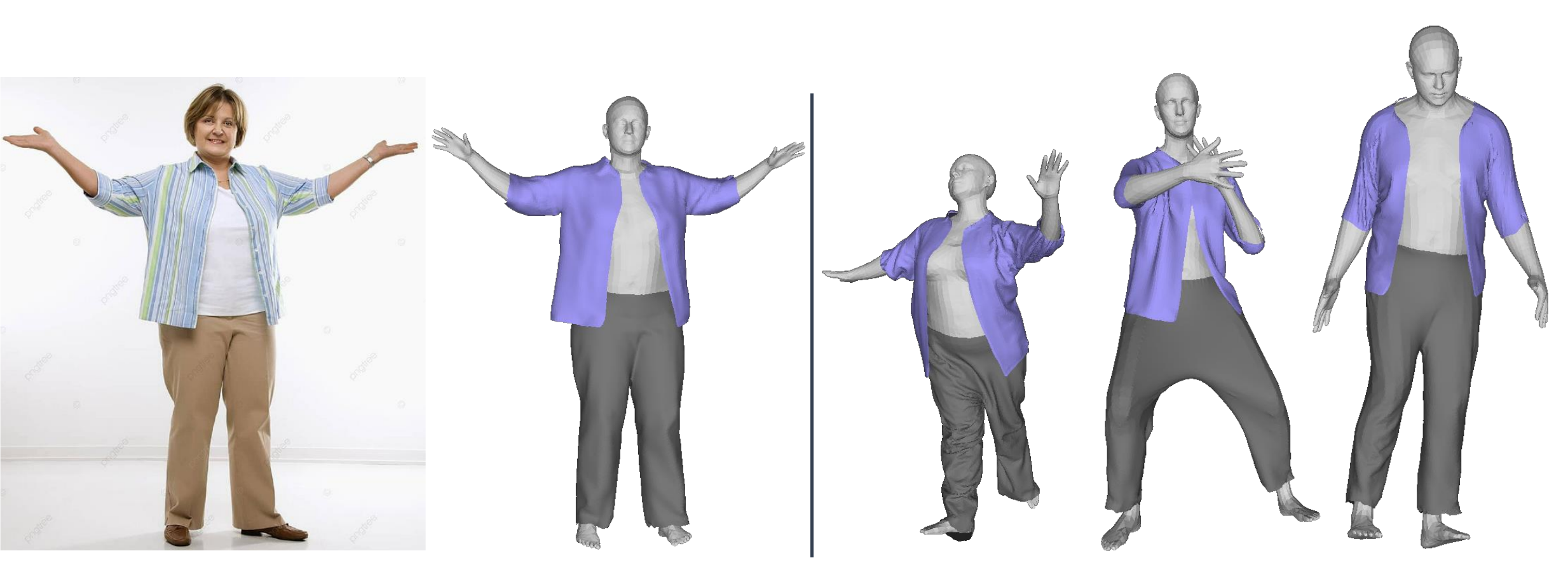}
        \vspace{-5mm}
        \caption{\textbf{Retargeting.} Left: The input image and our reconstructions. Right: Garments transfer to bodies with different poses and shapes.}
        \label{fig:repose}
    \end{minipage}
    \\
	\vspace{3mm}
	\begin{minipage}{.485\textwidth}
		\centering
        \includegraphics[width=.99\textwidth]{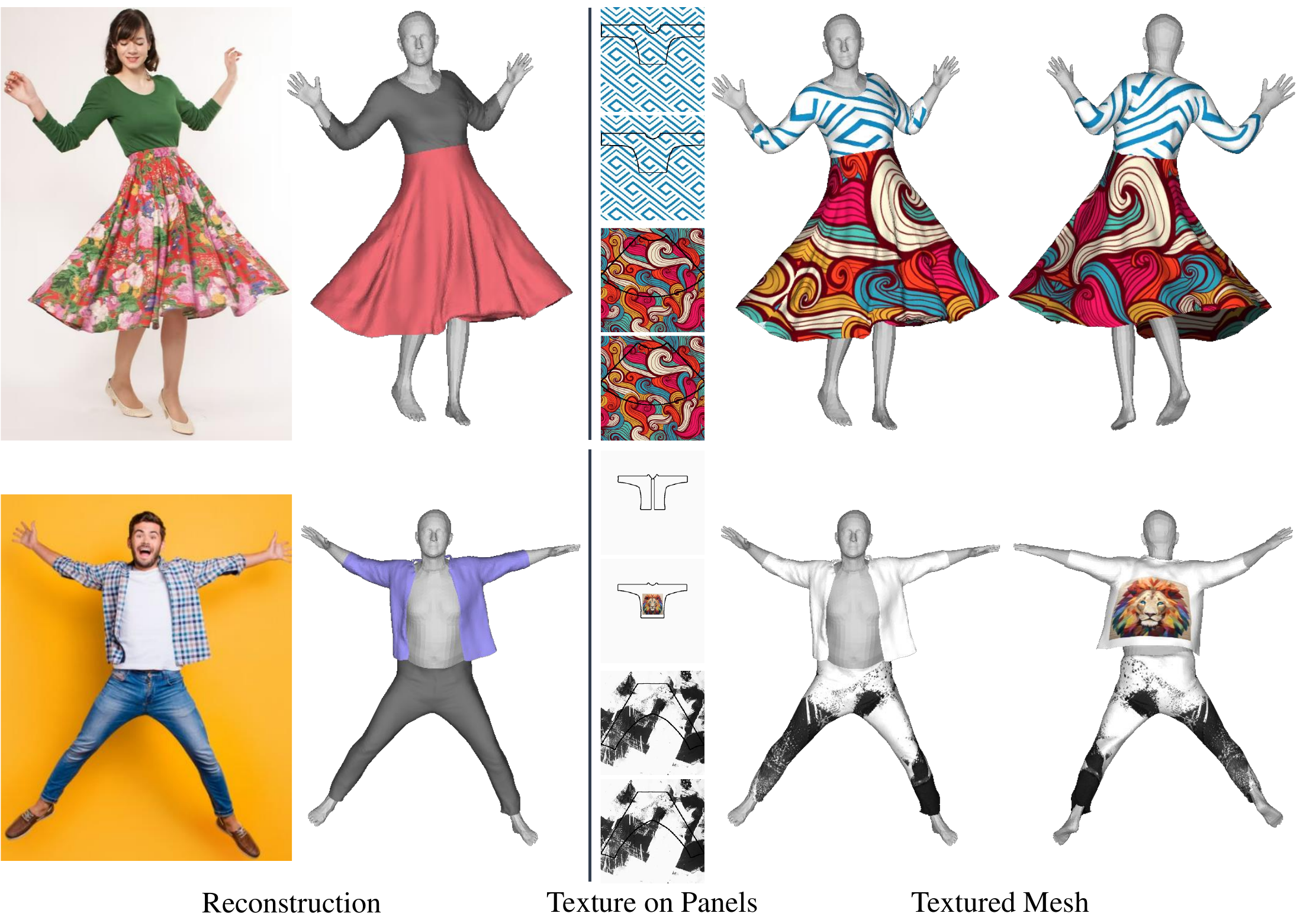}
        \vspace{-5mm}
        \caption{\textbf{Texture editing}. By drawing figures or patterns on the recovered 2D panels, we can directly edit the texture of the recovered garment mesh.}
        \label{fig:texture}
	\end{minipage}
    \vspace{-2mm}
\end{figure}

\subsection{Downstream Applications}

\parag{Retargeting} Since our method produces separate models for the garment and the underlying body, we can easily repose it on the new body. In Fig. \ref{fig:repose}, we show the retargeting results for the reconstructed open jacket and trousers by transferring them onto bodies with different poses and shapes. Accurate reconstruction of garments results in realistic retargeting. 

\parag{Texture Editing} Since we reconstruct both the 3D model and the corresponding 2D panels for garment, we can easily realize texture editing. As shown in Fig. \ref{fig:texture}, by painting patterns or drawing specific figures onto the recovered panels, the mesh will show the texture on the corresponding position.

\begin{figure}[hbt]
    \centering
    \includegraphics[width=\linewidth]{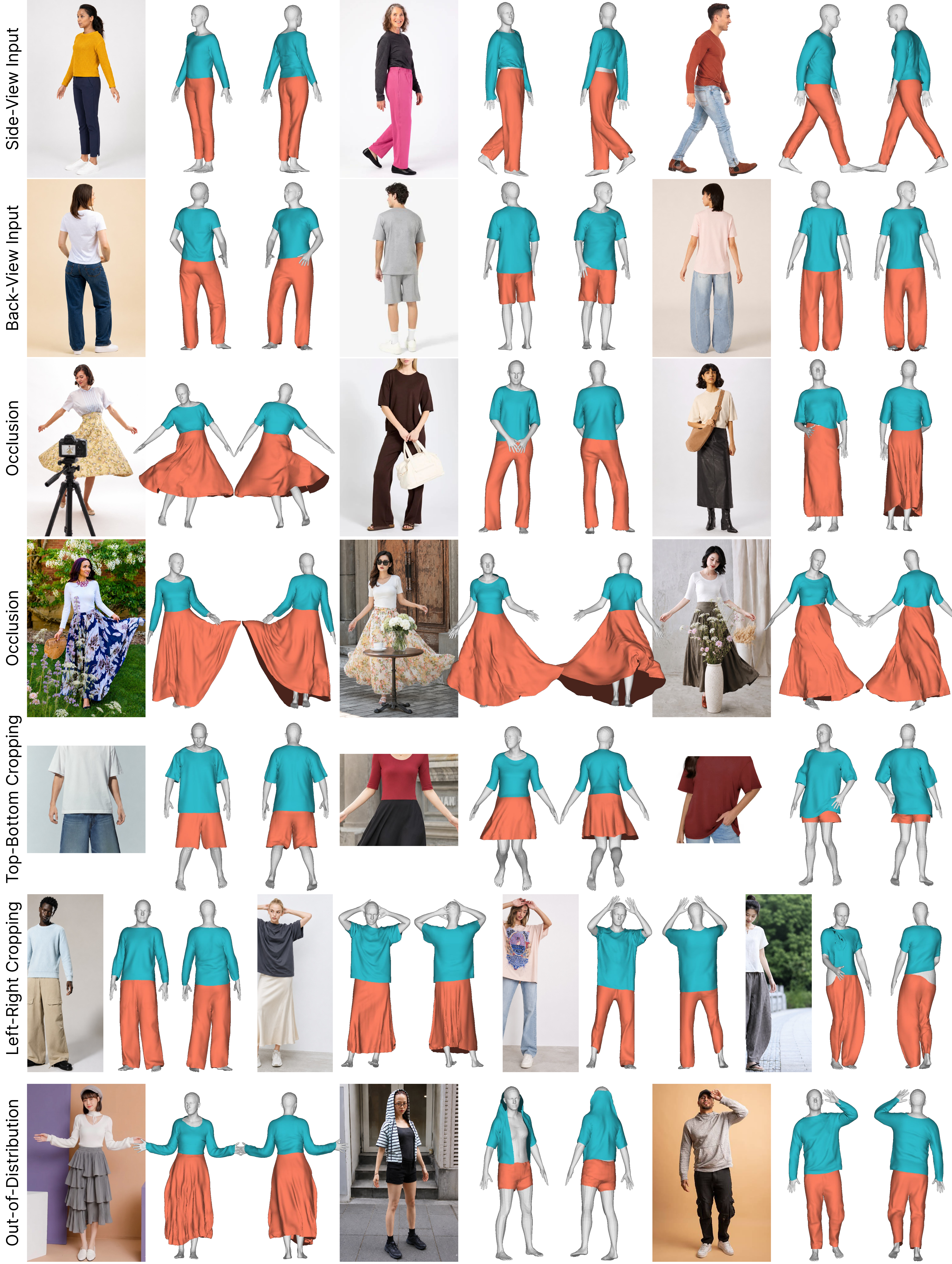}
    \vspace{-6mm}
    \caption{\re{Qualitative results under profile views, occlusions, cropping, and out-of-distribution garment types.}}
    \label{fig:re1_challenging_image}
    \vspace{-2mm}
\end{figure}
\begin{figure}[hbt]
    \centering
    \includegraphics[width=\linewidth]{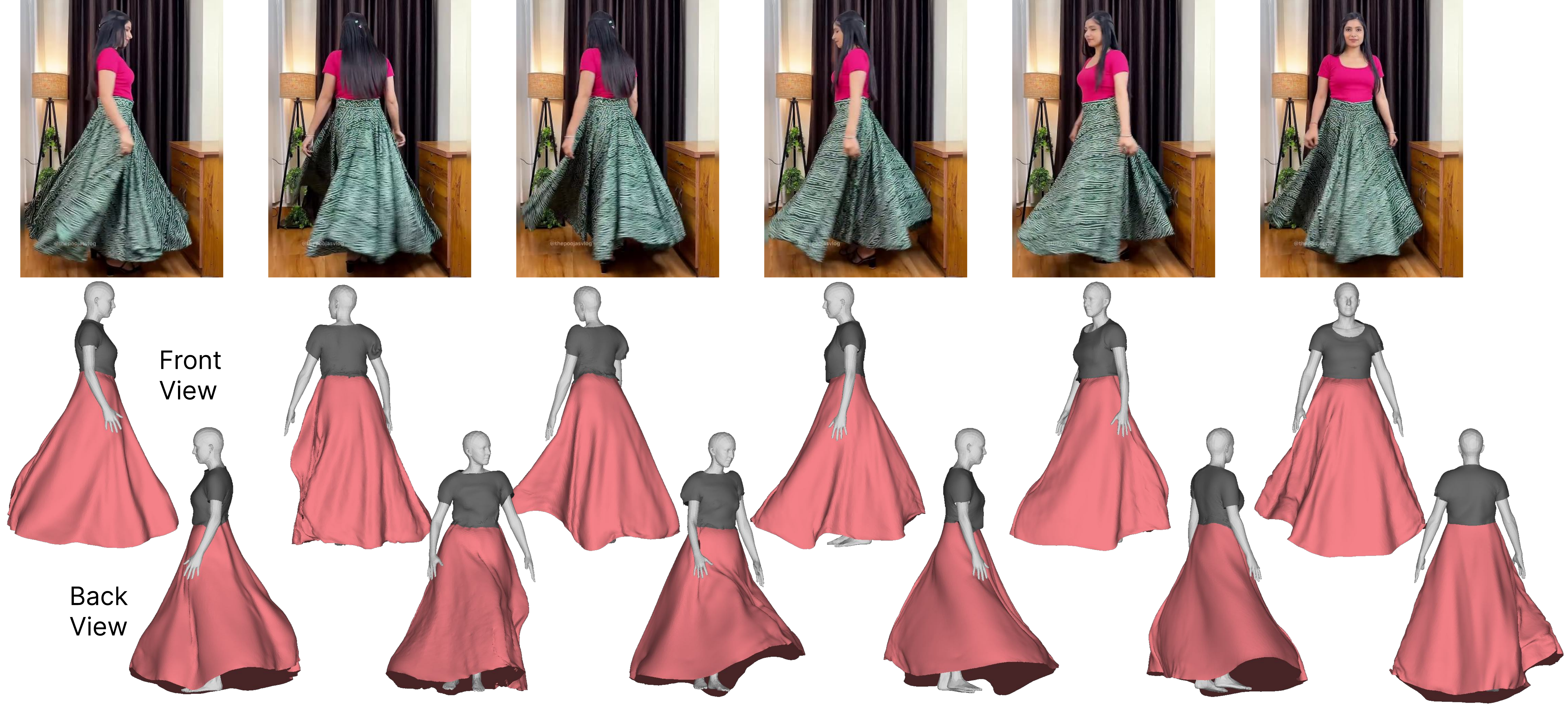}
    \vspace{-6mm}
    \caption{\re{Qualitative results on a profile-view and occluded video sequence.}}
    \label{fig:re1_challenging_video}
    \vspace{-3mm}
\end{figure}

\subsection{Challenging Cases}
As shown in Fig.~\ref{fig:re1_challenging_image} and Fig.~\ref{fig:re1_challenging_video}, we further evaluate our method on challenging cases, including side/back views, occlusions, cropping, and out-of-distribution garment structures. Our method handles profile-view inputs, completes invisible garment regions, and generalizes to naturally occurring partial occlusions. For challenging video frames, such as the T-shirt largely occluded by long hair in Fig.~\ref{fig:re1_challenging_video}, our spatio-temporal diffusion model leverages neighboring frames to perform accurate reconstruction.
We simulate severe garment truncation using top-bottom and left-right cropping. Under top-bottom cropping, our method completes truncated T-shirts into plausible full garments, but may preserve shortened trousers when the visible part remains consistent with a plausible garment prior. For left-right cropping, it generalizes to mild truncation but may fail under severe truncation due to insufficient body and garment information, where SMPL estimation can also become inaccurate. Addressing severe truncation with image completion by generative models before reconstruction could be a solution.
Finally, we evaluate out-of-distribution garment structures, including multi-layered ruffle skirts, hooded garments, and garments with pockets. Our method can approximate some geometric details, such as ruffle-like structures and pocket regions, but these garments are still represented as a single-layer surface in the current ISP/DISP representation. This limits its ability to faithfully reconstruct truly multi-layered or topologically complex garments. A promising direction is to extend ISP/DISP with a multi-panel sewing-pattern representation with separate UV domains and explicit seam connectivity.

% !TEX root = ../main.tex
% !TEX spellcheck = en-US

\section{Conclusion}

We have proposed a unified spatio-temporal framework for high-fidelity 3D and 4D garment generation from monocular visual input. When using single images as input, it delivers accurate static reconstructions even for loose-fitting clothing without requiring predefined templates. To this end, it uses UV-based sewing patterns to bridge the gap between 2D image observations and the 3D geometry of loose-fitting garments and relies on diffusion schemes to infer the parts of the garments that cannot be seen. 
When processing videos, our method extends the single-image diffusion scheme to a spatio-temporal formulation with test-time guidance that enforces long-range temporal consistency. For the more challenging spatio-temporal modeling of unobserved regions, we introduce analytic projection-based constraints that faithfully preserve visible garment geometry while enforcing spatial and temporal consistency in occluded areas.
Extensive experiments on benchmarks and real-world imagery demonstrate that our method consistently outperforms existing approaches on both static and dynamic garments, achieving high fidelity and strong temporal consistency.

\textbf{Limitations and Future Work:} Modeling multi-layered garments beyond the standard setting of our current framework. As shown in Fig.~\ref{fig:re1_multilayer}, our method can reconstruct geometric details such as ruffles and layered regions, but these details are still represented on a single-layer garment surface. This limitation comes from the current ISP/DISP representation, which parameterizes garments as single-layer surfaces in a unified UV space. 
A promising future direction is to extend ISP/DISP with a multi-panel sewing-pattern representation, where garment panels from different layers have their own 2D UV domains and explicit seam connectivity. Such a representation could better handle complex garment structures, including ruffles, layered skirts, and pockets.

\begin{figure}[hbt]
    \centering
    \includegraphics[width=\linewidth]{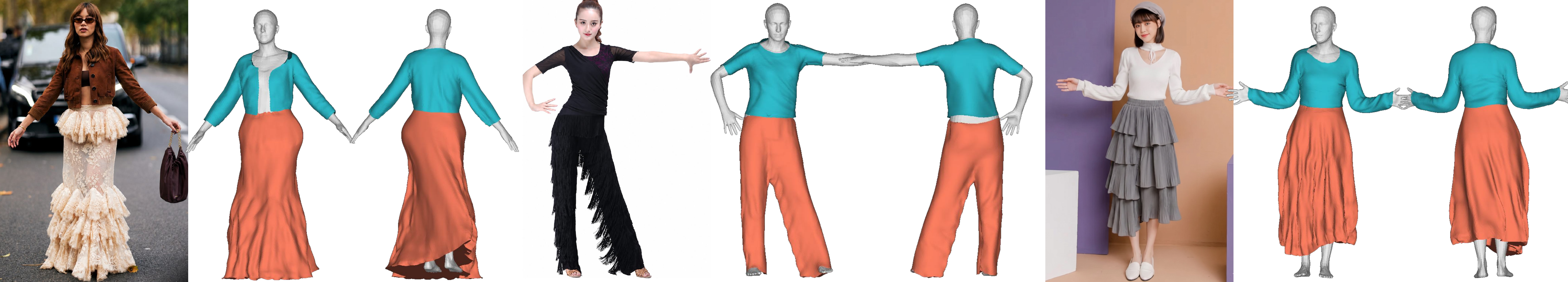}
    \vspace{-5mm}
    \caption{\re{Reconstruction results on multi-layered structures.}}
    \label{fig:re1_multilayer}
    \vspace{-2mm}
\end{figure}

\section*{Acknowledgments}
This work was supported in part by the Swiss National Science Foundation grant and by the Metaverse Center Grant from the MBZUAI Research Office.

{\small
\bibliographystyle{ieee_fullname}
\bibliography{bib/string,bib/reference}
}

\clearpage

\section{Supplementary Material}

\setcounter{figure}{0}
\renewcommand{\thefigure}{\Alph{figure}}
\renewcommand{\theHfigure}{supp.\arabic{figure}}

\setcounter{table}{0}
\renewcommand{\thetable}{\Alph{table}}
\renewcommand{\theHtable}{supp.\arabic{table}}

% \setcounter{page}{1}
% \pagenumbering{arabic}

\noindent This supplementary material contains the following parts:

(A) Additional Qualitative Results.

(B) Range-Null Space Decomposition and DDPM$_p$.

(C) Recovering Rest Geometry with Accumulated Masks.

(D) Implementation Details.

(E) Tight- to Loose-Fitting Variations.

(F) Dataset Visualization.

(G) Supplementary Video {`supplementary\_video.mp4'}.

\subsection{Additional Qualitative Results}
\label{subsec:supp_visial_results}
\parag{Results of Pattern-Draping-Fitting Method}
Estimating sewing patterns, draping the garment onto the body, and then fitting it to the depth map is a viable alternative. To test this, we used ChatGarment~\cite{bian2025chatgarment}, a recent sewing pattern estimation method, to estimate the sewing pattern. We then draped the garment mesh onto the body and fitted it to the predicted depth images. As shown in Fig.~\ref{fig:supp_R1_1_image} and Fig.~\ref{fig:supp_R1_1_video}, this pipeline can generate plausible garments, but the results are less accurate than ours and exhibit weaker alignment with the images, especially for loose-fitting garments such as skirts and open jackets.
The inaccuracy has four main causes: 1) the estimated sewing pattern may be inaccurate; 2) the draped garment driven by the human body may not capture the actual garment dynamics observed in the input; 3) fitting is difficult because the correspondences between image pixels and mesh vertices are ambiguous; and 4) when applied to videos, frame-wise errors can accumulate and introduce temporal jitter.

\begin{figure}[htbp]
    \centering
    \includegraphics[width=\linewidth]{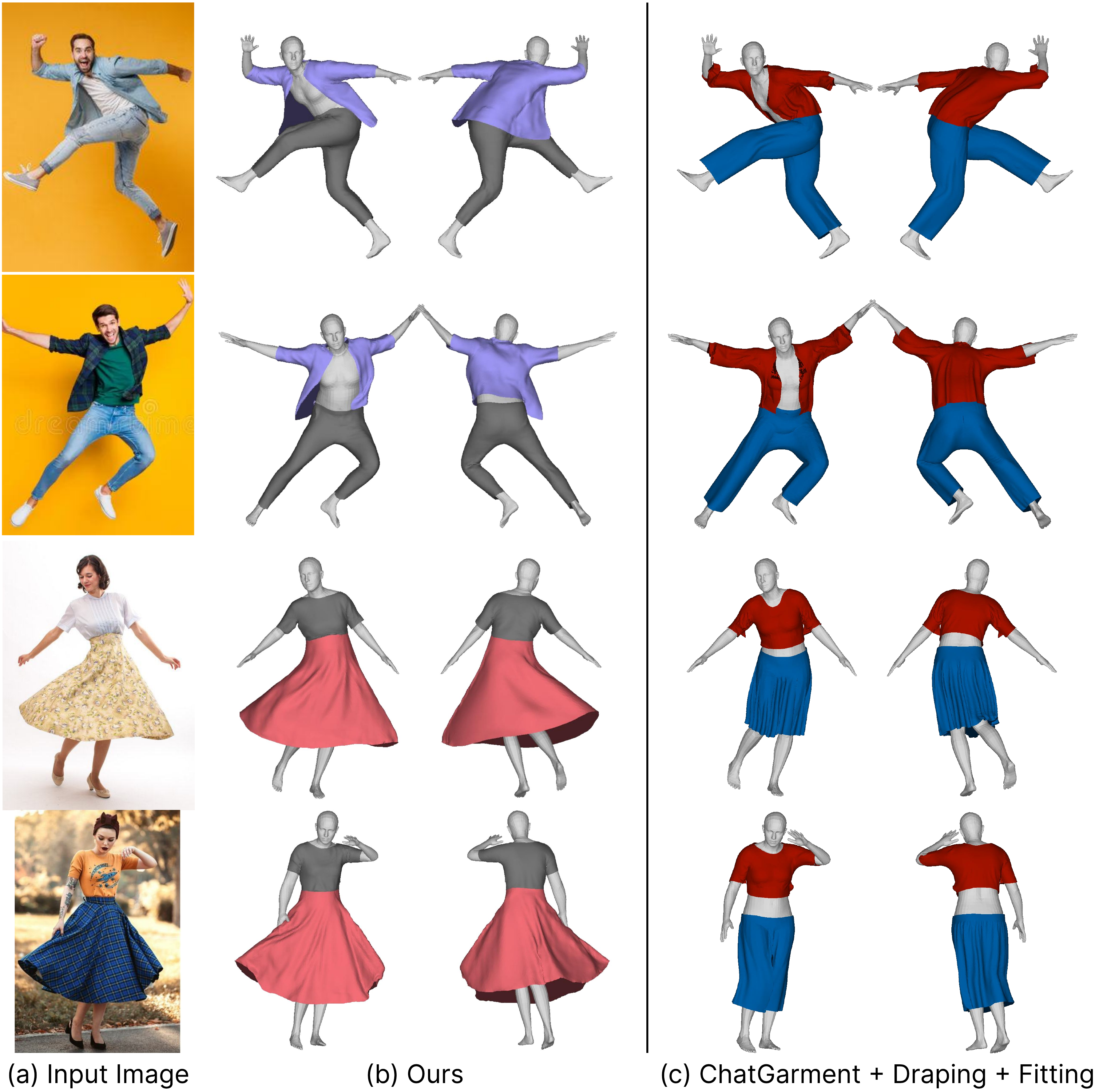}
    \caption{Qualitative comparison with the pattern-draping-fitting method on images.}
    \label{fig:supp_R1_1_image}
\end{figure}

\begin{figure*}[t]
    \centering
    \includegraphics[width=\linewidth]{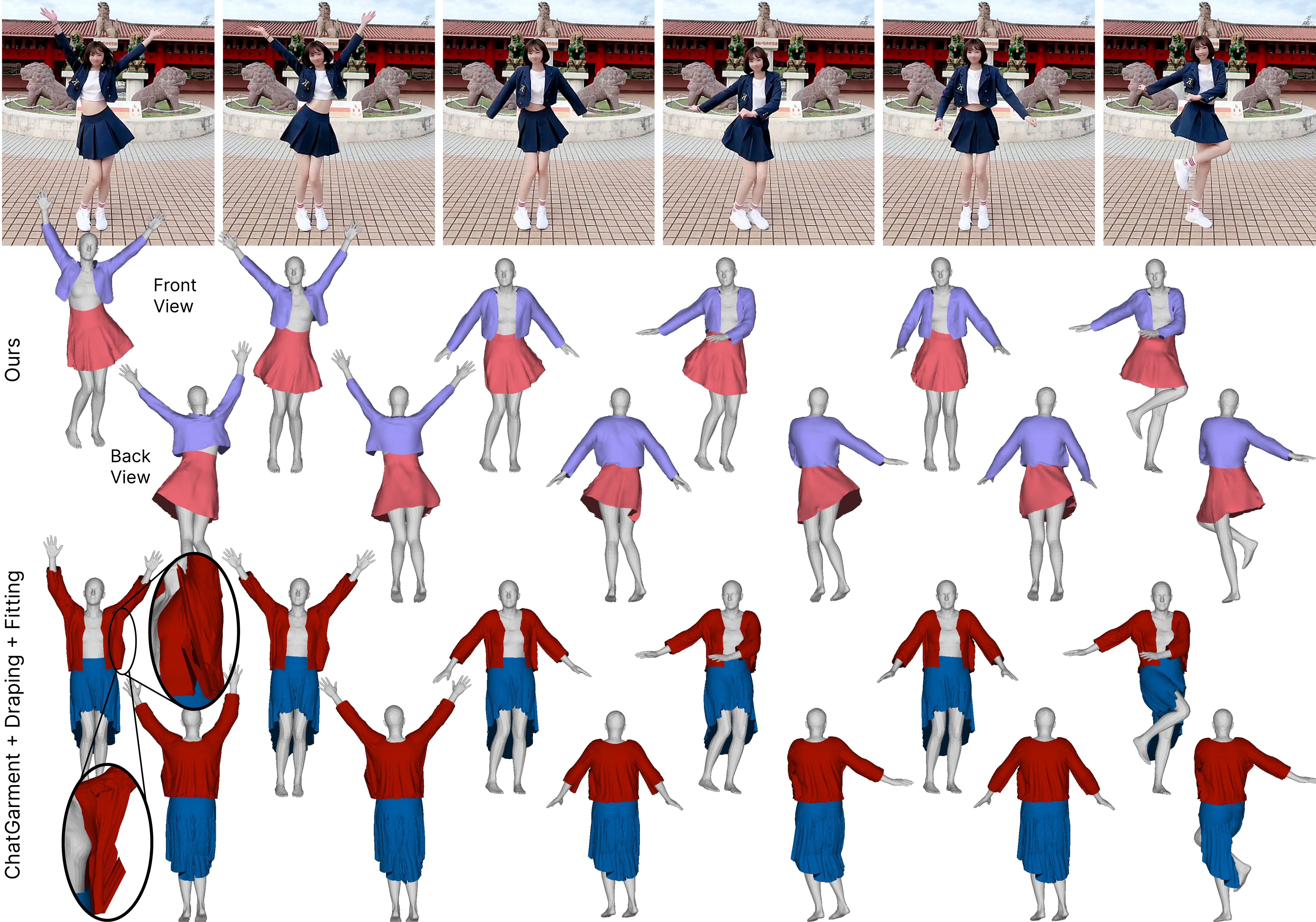}
    \caption{Qualitative comparison with the pattern-draping-fitting method on a video sequence.}
    \label{fig:supp_R1_1_video}
\end{figure*}

\parag{Static Reconstruction}
In Fig. \ref{fig:garec}, we show additional comparison with GaRec~\cite{Li24b}. \dmapS{} can capture high-fidelity surface details and complex deformations, while GaRec~\cite{Li24b} produces results overly smooth and lack realistic wrinkles. In Fig. \ref{fig:supp_fitting}, we show more reconstruction results produced by \dmapS{} for in-the-wild images. As illustrated, \dmapS{} accurately reconstructs garments, capturing intricate deformations and fine surface details.

% !TEX root = ../top.tex
% !TEX spellcheck = en-US

\begin{figure}[ht!]
    \centering
    \includegraphics[width=.48\textwidth]{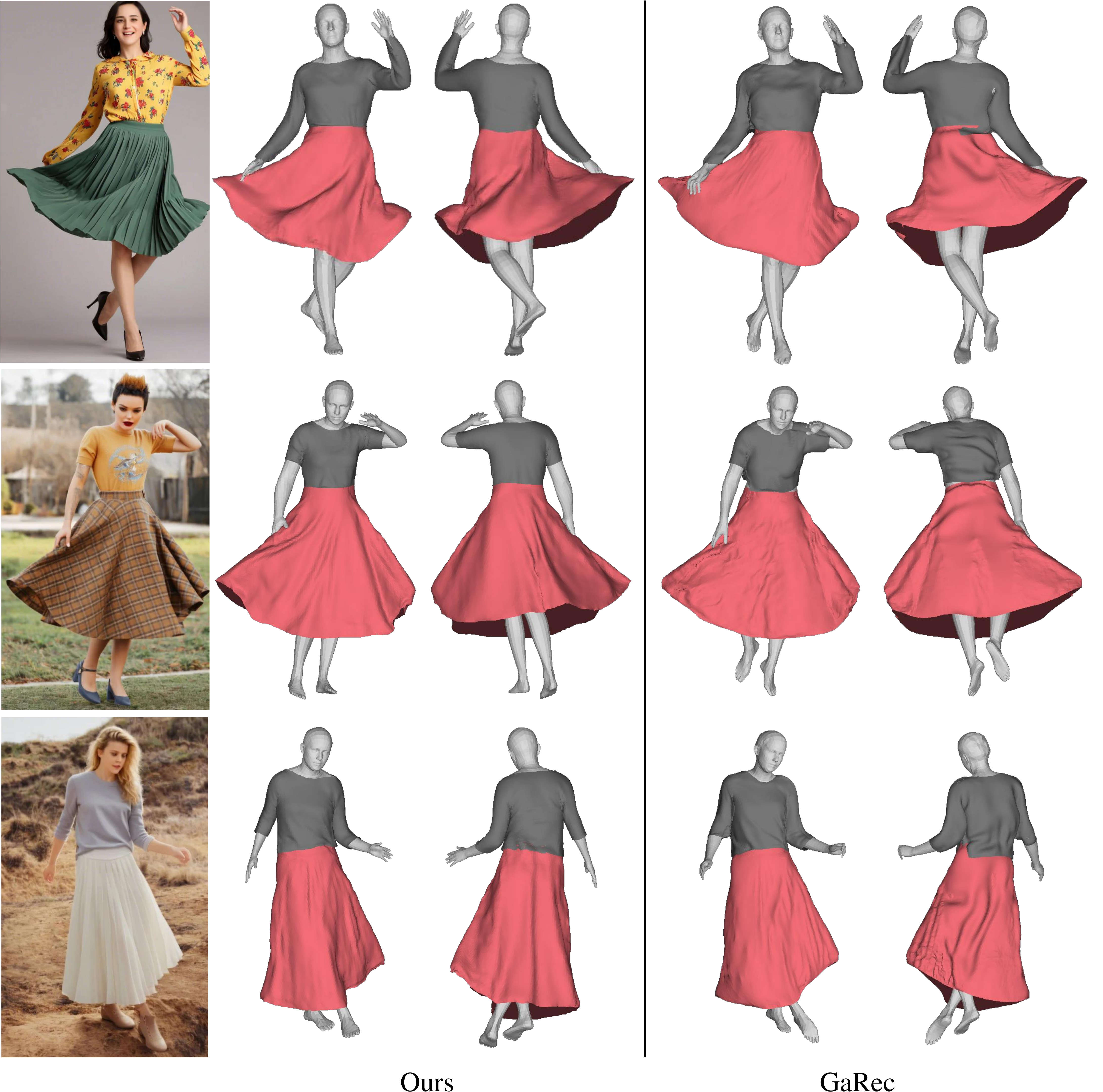}
    %\vspace{-0.35cm}
    \caption{Comparison between \dmapS{} and GaRec~\cite{Li24b}. %Input images courtesy of Dreamina.
    }
    \label{fig:garec}
    %\vspace{-0.45cm}
\end{figure} 

\begin{figure*}[ht!]
    \centering
    \includegraphics[width=.98\textwidth]{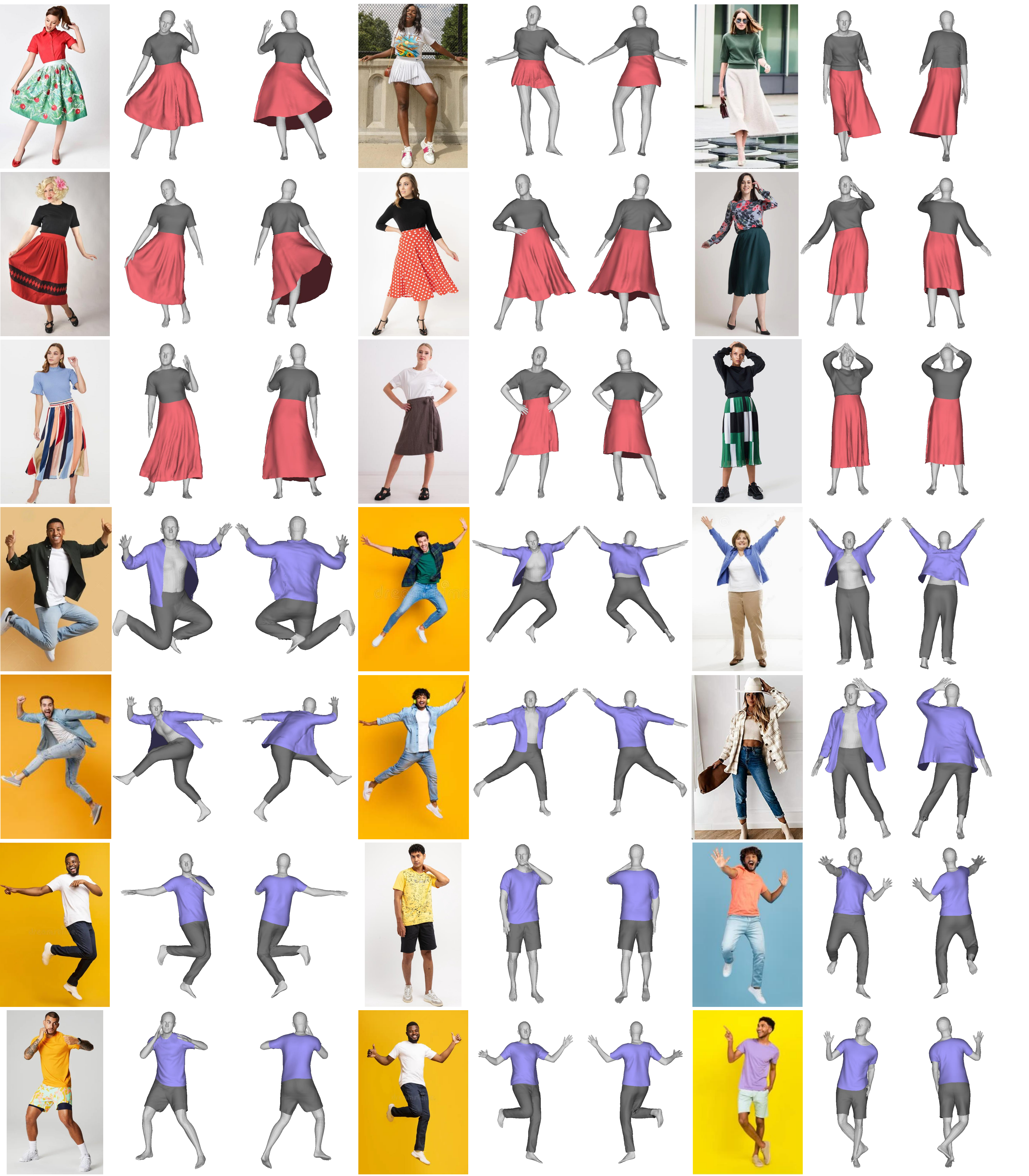}
    %\vspace{-0.35cm}
    \caption{\textbf{More reconstruction results for in-the-wild images.} \dmapS{} can handle both the tight-fitting and the loose-fitting garments and recover high-fidelity 3D meshes for them.}
    \label{fig:supp_fitting}
    \vspace{-0.45cm}
\end{figure*} 
\begin{figure*}[p]
    \centering
    \includegraphics[width=.8\linewidth]{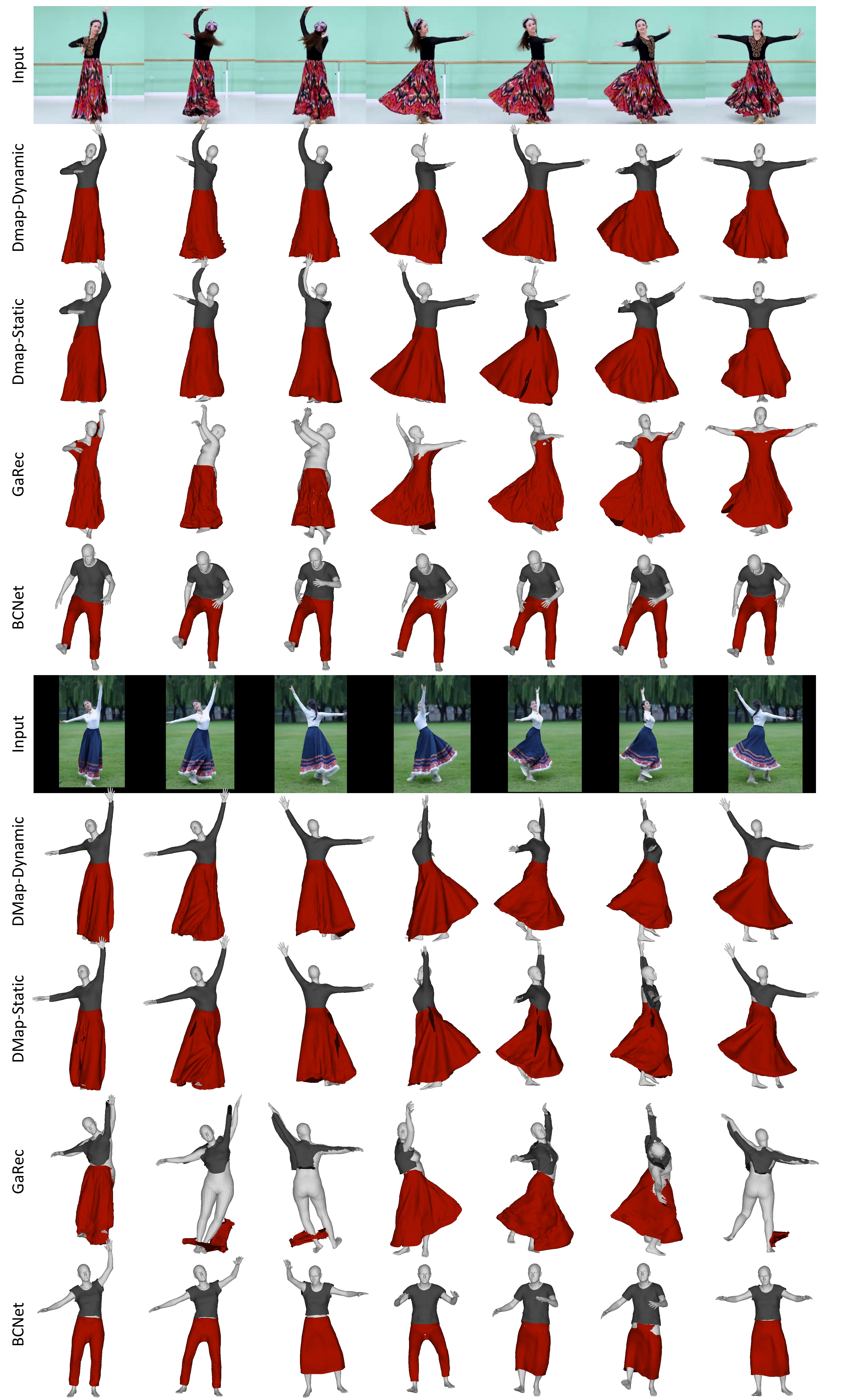}
    \vspace{-3mm}
    \caption{\textbf{Additional qualitative comparison for in-the-wild videos.} \dmapD{} achieves both geometric accuracy and temporal consistency.}
    \label{fig:supp_comp_vis}
\end{figure*}

\parag{Dynamic Reconstruction}
Fig.~\ref{fig:supp_comp_vis} presents additional qualitative comparisons on real-world video sequences with previous methods, including BCNet~\cite{Jiang20d}, GaRec~\cite{Li24a}, and \dmapS{}. These examples are particularly challenging due to fast motions, loose-fitting garments, and occlusions. BCNet is limited to frontal views and struggles to reconstruct accurate garment types or body poses. GaRec often fails under back or side views, lacking spatiotemporal consistency. Although \dmapS{} achieves high per-frame accuracy, it also fails to model spatio-temporal consistency, producing jittery results as illustrated in the supplementary video. 
Besides, it tends to produce visual artifacts during garment motion. In contrast, \dmapD{} achieves the best overall performance, combining per-frame accuracy, physical plausibility and temporal consistency.
Please refer to the supplementary video for dynamic comparisons.

\parag{Visualization of Error}
Fig. \ref{fig:supp_cd2} and \ref{fig:supp_cd1} visualize the spatial error distribution on reconstructed garment meshes, measured using one-directional Chamfer Distance. Our method achieves lower error across the entire surface, whereas competing methods exhibit significant errors, particularly in regions farther from the underlying body.
\begin{figure*}[ht!]
    \centering
    \includegraphics[width=.95\textwidth]{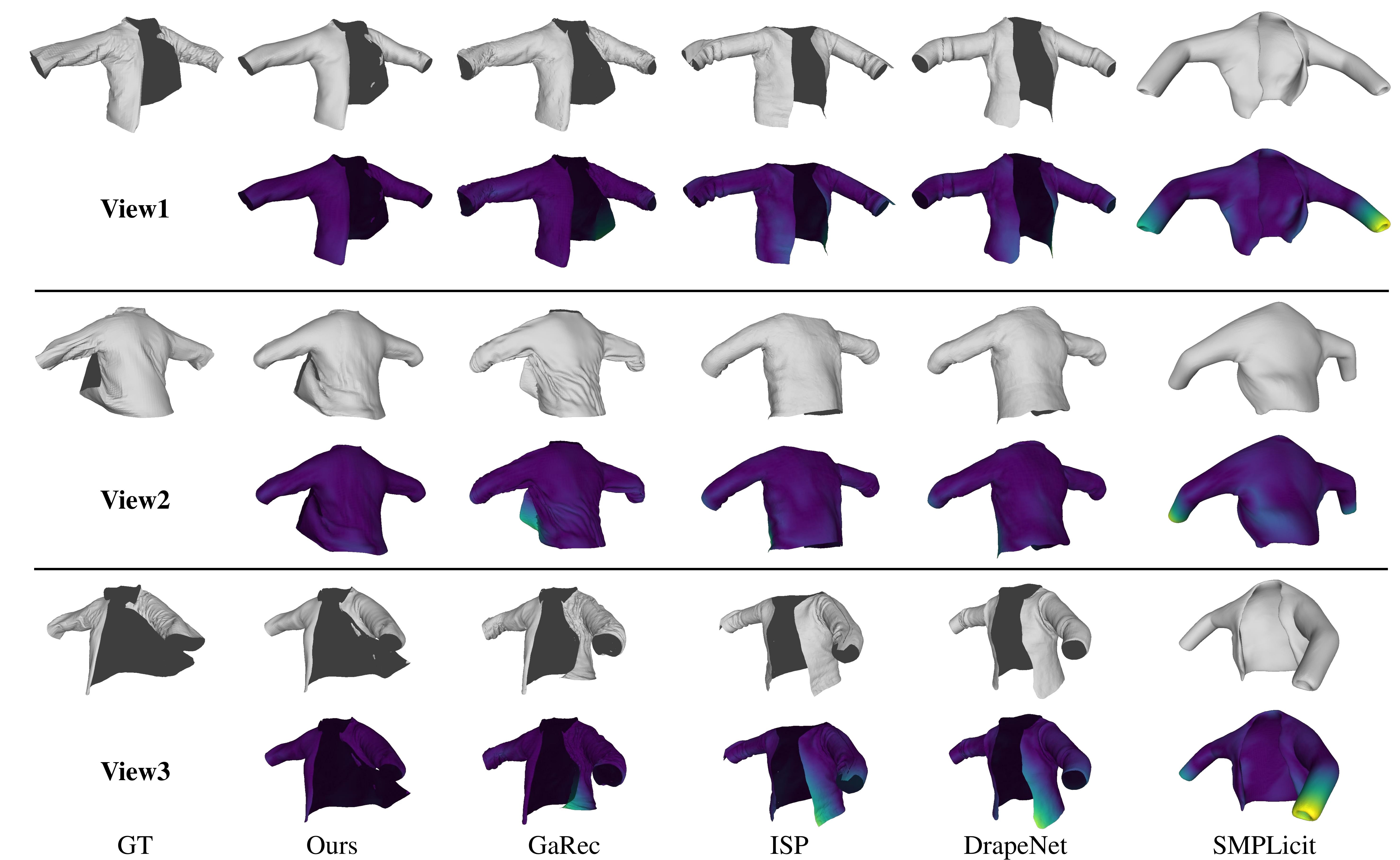}
    \caption{\textbf{Visualization of error distribution on the reconstructed jacket mesh.} The bright colors indicate large errors.}
    \label{fig:supp_cd2}
\end{figure*} 
\begin{figure}[h!]
    \centering
    \includegraphics[width=.48\textwidth]{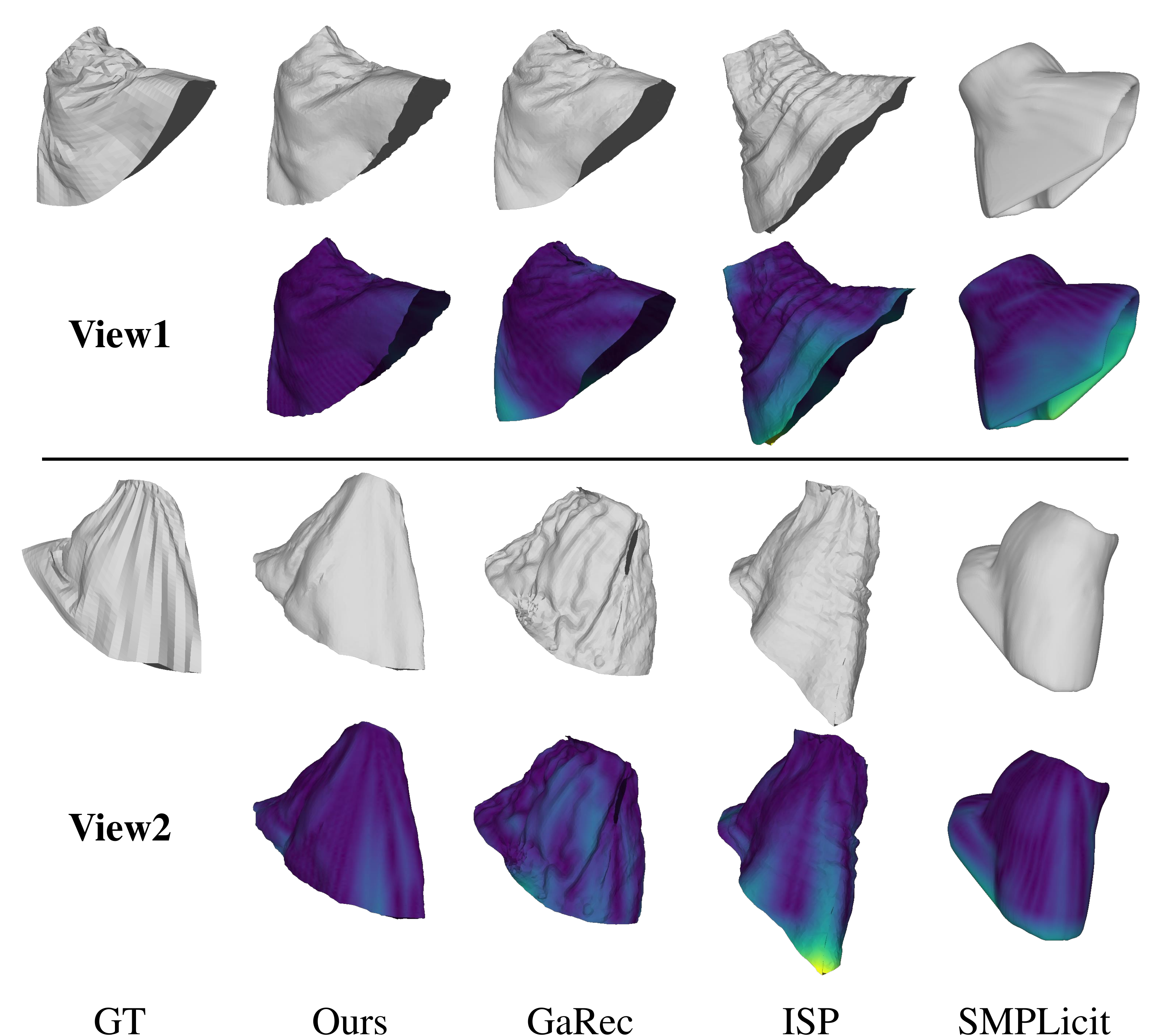}
    \caption{\textbf{Visualization of error distribution on the reconstructed skirt mesh.} The bright colors indicate large errors.}
    \label{fig:supp_cd1}
\end{figure}

\subsection{Range-Null Space Decomposition and DDPM$_p$}
Given a linear operator $\mathbf{A}$, any sample $\bx$ can be split into range- and null-space parts of $\mathbf{A}$:
\begin{equation}
    \bx \equiv \underbrace{\mathbf{A}^{\dagger} \mathbf{A} \bx}_{\text{range-space part}} + \underbrace{(\mathbf{I} - \mathbf{A}^{\dagger} \mathbf{A}) \bx}_{\text{null-space part}},
\end{equation}
where $\mathbf{A}^\dagger$ is the pseudo-inverse of $\mathbf{A}$. The matrices $\mathbf{A}$ and $\mathbf{A}^\dagger$ exhibit useful projection properties. Specifically, $\mathbf{A}^\dagger \mathbf{A}$ projects $\bx$ onto the range space of $\mathbf{A}$, since $\mathbf{A} \mathbf{A}^\dagger \mathbf{A} \mathbf{x} \equiv \mathbf{A} \mathbf{x}$. Conversely, $(\mathbf{I} - \mathbf{A}^\dagger \mathbf{A})$ projects $\bx$ onto the null space of $\mathbf{A}$, as $\mathbf{A} (\mathbf{I} - \mathbf{A}^\dagger \mathbf{A}) \mathbf{x} \equiv \mathbf{0}$.

Inspired by~\cite{Wang22c}, which uses range-null space decomposition in the DDPM~\cite{Ho20a} denoising process to solve the inverse problems in 2D images, we introduce such projection-based constraint in the positional representation and temporal guidance to recover the garment deformations by completing positional map from the partial observation $\tilde{\bU}$ and its corresponding mask $\tilde{\bM}$. We aim for the final output $\mathbf{x}_0$ to both satisfy the partial observation constraint $\tilde{\mathbf{M}} \mathbf{x}_0 \equiv \tilde{\mathbf{U}}$ and remain plausible with temporal consistency. 
The denoising step is written as:
\begin{equation}\label{eq:inpaint_sup}
\scalebox{0.81}{$
    \mathbf{x}_{t-1} = \text{DDPM}_p(\mathbf{x}_t, \tilde{\bU}, \tilde{\bM}, \boldsymbol{\epsilon}^d_{\boldsymbol{\theta}}(\mathbf{x}_t, t)) - \gamma_v\nabla_{\mathbf{x}_t}\mathcal{G}_{vel}^d - \gamma_a\nabla_{\mathbf{x}_t}\mathcal{G}_{acc}^d \; ,$
    }
\end{equation}
where $\gamma_v$ and $\gamma_a$ are the guidance scales. $\text{DDPM}_p$ projects the estimate into observed and unobserved components using range-null space decomposition. Specifically, during the denoising process, we denote $\mathbf{x}_{0|t}$ as the predicted clean sample at timestep $t$ computed from the noisy sample $\mathbf{x}_t$:
\begin{equation}
\label{eq:x_0t_sup}
\mathbf{x}_{0 \mid t} = \frac{1}{\sqrt{\bar{\alpha}_t}} \left( \mathbf{x}_t - \sqrt{1 - \bar{\alpha}_t} \cdot \boldsymbol{\epsilon}^d_{\boldsymbol{\theta}}(\mathbf{x}_t, t) \right),
\end{equation}
where $\bar{\alpha}_t$ is the scale factor. This equation retains equivalence with the original DDPM~\cite{Ho20a}. To constrain the final generation to satisfy the partial observation $\tilde{\bM} \mathbf{x}_0 \equiv \tilde{\bU}$, we modify $\bx_{0 \mid t}$ by fixing the range-space as $\tilde{\bM}^{\dagger} \tilde{\bU}$ and leaving the null-space unchanged:
\begin{equation}
\label{eq:hat_x_0t_sup}
\hat{\mathbf{x}}_{0 \mid t} = \tilde{\bM}^{\dagger} \tilde{\bU} + \left( \mathbf{I} - \tilde{\bM}^{\dagger} \tilde{\bM} \right) \mathbf{x}_{0 \mid t} \; ,
\end{equation}
where $\tilde{\bM}^{\dagger}$ is the pseudo-inverse of the observation mask $\tilde{\bM}$. Intuitively, $\tilde{\bM}^{\dagger}$ projects $\mathbf{x}_{0 \mid t}$ to the range-space of $\tilde{\bM}$, satisfying the observation $\tilde{\bM} \hat{\mathbf{x}}_{0 \mid t} \equiv \tilde{\bU}$, while $(\mathbf{I} - \tilde{\bM}^{\dagger} \tilde{\bM})$ projects $\mathbf{x}_{0 \mid t}$ to the null-space of $\tilde{\bM}$, which does not affect the observation but determines whether $\hat{\mathbf{x}}_{0 \mid t}$ follows the prior distribution. 

Then, DDPM$_p$ rewrites the sampling formulation from posterior distribution $p\left(\mathbf{x}_{t-1} \mid \mathbf{x}_t, \mathbf{x}_0\right)$ of DDPM by replacing the $\mathbf{x}_{0}$ with it estimate $\hat{\mathbf{x}}_{0 \mid t}$, which is written as:
\begin{equation}
\label{eq:ddpm_p_sup}
\scalebox{0.78}{$
\text{DDPM}_p(\mathbf{x}_t, \tilde{\bU}, \tilde{\bM}, \boldsymbol{\epsilon}^d_{\boldsymbol{\theta}}(\mathbf{x}_t, t)) = \frac{\sqrt{\bar{\alpha}_{t-1}} \beta_t}{1 - \bar{\alpha}_t} \hat{\mathbf{x}}_{0 \mid t}
	+ \frac{\sqrt{\alpha_t} (1 - \bar{\alpha}_{t-1})}{1 - \bar{\alpha}_t} \mathbf{x}_t + \sigma_t \boldsymbol{\epsilon},$
}
\end{equation}
where $\boldsymbol{\epsilon} \sim \mathcal{N}(0, \mathbf{I})$, and $\alpha$, $\bar{\alpha}$, $\beta$ are scales factors predefined in DDPM~\cite{Ho20a}. The property of DDPM$_p$ is that the range-space component $\tilde{\bM}^{\dagger} \tilde{\bU}$ remains fixed, while only the null-space component $( \mathbf{I} - \tilde{\bM}^{\dagger} \tilde{\bM} ) \mathbf{x}_{0 \mid t}$ undergoes denoising to eliminate disharmony between the two components. By iteratively denoising using Eqs.~\ref{eq:inpaint_sup},~\ref{eq:x_0t_sup},~\ref{eq:hat_x_0t_sup},~\ref{eq:ddpm_p_sup}, we obtain a final result $\bx_{0}$ that satisfies the partial observation $\tilde{\bM} \bx_0 \equiv \tilde{\bU}$ while remaining within the distribution $\bx_0 \sim q(\bx)$. This projection-based constraint is crucial here, 
as it generates positional maps $\hat{\bU}$ by maximally aligning the extracted observations $\tilde{\bU}$ and satisfies the learned diffusion priors that maintain harmony between the observed and unobserved regions, resulting in a plausible and complete reconstruction.

\subsection{Recovering Rest Geometry with Accumulated Partial Masks for Dynamic Reconstruction}
\label{sec:accumulate_mask}

\begin{figure*}[hbt]
    \centering
    \includegraphics[width=0.8\linewidth]{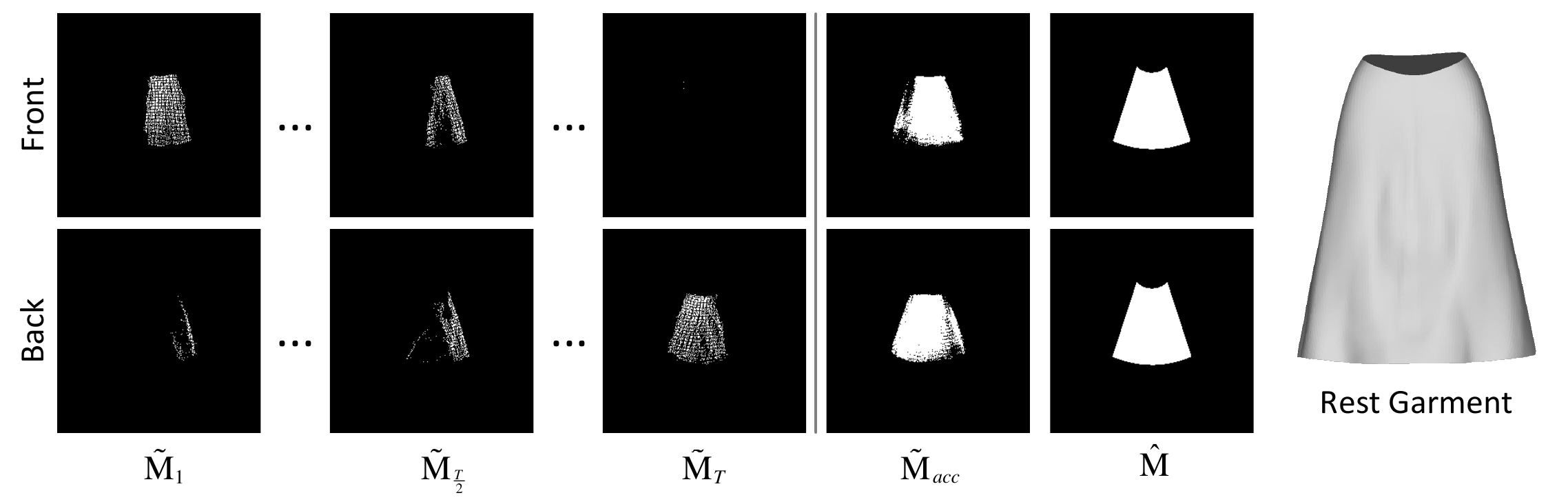}
    \caption{Recovering the rest garment using accumulated partial masks in \dmapD{}.}
    \label{fig:supp_mask}
\end{figure*}

For the dynamic reconstruction, instead of using $\tilde{\bM}$ from a single image as introduced in Sec.~\ref{sec:recoverPanel}, we accumulate the partial masks across all frames to form a more complete representation $\tilde{\bM}_{acc} = {\bigvee}_{f=1}^{T} \tilde{\bM}_f$ in \dmapD{}, since the observed masks from different frames offer globally complementary coverage.

We optimize the latent code of the garment prior so that the decoded panel patterns closely match $\tilde{\bM}_{acc}$ (Eq.~\ref{eq:z}). The resulting optimal latent code is subsequently used to infer the complete panel mask $\hat{\bM}$ and the corresponding rest garment mesh, as shown in Fig.~\ref{fig:supp_mask}.

\subsection{Implementation Details}
\parag{Networks and Training}
Following \cite{Li24a,Li23a}, we implement the pattern parameterization network $\mathcal{I}_{\Theta}$ and the UV parameterization network $\mathcal{A}_{\Phi}$ of ISP as MLPs with the latent code $\bz$ of size 128. $\mathcal{I}_{\Theta}$ and $\mathcal{A}_{\Phi}$ are trained jointly for 9000 iterations with a batch size of 50. 

For the diffusion models $\boldsymbol{\epsilon}^d_{\boldsymbol{\theta}}$, $\boldsymbol{\epsilon}_{\theta}^n$ and $\boldsymbol{\epsilon}_{\theta}^m$ of \dmapS{}, we use a U-Net architecture \cite{Ronneberger15} with attention blocks in the middle. They all have 6 down-sampling blocks and 6 up-sampling blocks. 
The numbers of out channels of each block are [128, 128, 256, 256, 512, 512]. 
The output of $\boldsymbol{\epsilon}_{\theta}^d$ is the concatenated front and back UV maps and panel maps with the dimensions of $128\times256\times4$, and the input is the noise images having the same size as the output. 
The output of $\boldsymbol{\epsilon}_{\theta}^n$ is the concatenation of an estimated normal image for the garment back and a mask indicating foreground pixels with the size of $192\times192\times4$. The input of $\boldsymbol{\epsilon}_{\theta}^n$ is the concatenation of [$\bn_F$, $\bs_F$, $\bs_B$, $\bd_F^b$, $\bd_B^b$] and the noise image, whose size is $192\times192\times11$.
The output of $\boldsymbol{\epsilon}_{\theta}^m$ is the estimation of UV coordinates ($\bc_F$, $\bc_B$) and depth ($\bd_F^g$, $\bd_B^g$) with the size of $192\times192\times8$, and the input is the concatenation of [$\bn_F$, $\bn_B$, $\bs_F$, $\bs_B$, $\bd_F^b$, $\bd_B^b$] and the noise image, having the size of $192\times192\times18$.
$\boldsymbol{\epsilon}_{\theta}^d$ is trained for 100 epochs on frames sampled every 5 frames, with a learning rate of 1e-4, a batch size of 64, and $T=1000$ steps. We apply random rotations to the mesh represented as UV maps for augmentation. 
$\boldsymbol{\epsilon}_{\theta}^n$ and $\boldsymbol{\epsilon}_{\theta}^m$ are trained for 200 epochs on the subsampled set of frames, with a learning rate of 1e-4, a batch size of 162, and $T=1000$ steps. 

The temporal modules of \dmapD{} for $\boldsymbol{\epsilon}_{\theta}^n$ and $\boldsymbol{\epsilon}_{\theta}^m$ each consist of two linear layers around a temporal attention block. Both spatio-temporal diffusion models are trained on consecutive frames without subsampling, using a sequence length of $T=10$, a batch size of 8, a learning rate of $1\times10^{-4}$, and 500 warm-up steps. Training is performed with the Adam optimizer~\cite{Kingma15a} on 4 NVIDIA A100 GPUs with 80GB VRAM, and each spatio-temporal diffusion model takes about 100 hours to train.

The neural displacement field $f_\phi$ introduced in Sec. \ref{sec:post} is a 4-layer MLP with Softplus activations in-between. The output channels of each layer is [256, 256, 256, 3]. For each garment, we optimize a separate randomly-initialized MLP.

% !TEX root = ../top.tex
% !TEX spellcheck = en-US

\begin{table}[ht]
      \caption{The values of hyper-parameters.}
      \label{tab:param}
    \begin{center}
    \begin{tabular}{c | c  }
    \toprule
      Eq. \ref{eq:z} & $\lambda_{area}=0.5, \lambda_{\bz}=0.02$  \\
      Eq. \ref{eq:guided_loss} & $\rho=20$  \\
      Eq. \ref{eq:postrefine} & $\lambda_m=1, \lambda_d=2,  \lambda_n=2, \lambda_u=100, \lambda_p=1$ \\
    \bottomrule
    \end{tabular}
      \end{center}
\end{table}
\begin{table}[t]
\centering
\setlength{\tabcolsep}{6.2mm}
\caption{Ablation study. w/o $N_B$ means optimization without using the back normal estimation $N_B$. 
$f_\phi$ denotes optimization using the neural displacement field, and $V$ denotes optimization conducted directly on vertex positions.}
\vspace{1mm}
\begin{tabular}{l|ccc}
\toprule
& CD $\downarrow$ & IoU $\uparrow$ & NC $\uparrow$ \\
\midrule
w/o $N_B$           & 1.67 & 90.04 & 0.78 \\
\midrule
w/o $f_\phi$, w/o $V$ & 1.45 & 92.67 & 0.81 \\
w/o $f_\phi$, w/ $V$  & 1.32 & 93.34 & 0.82 \\
w/ $f_\phi$, w/o $V$  & 1.27 & 94.71 & 0.82 \\
\midrule
\textbf{Ours}        & \textbf{1.21} & \textbf{95.32} & \textbf{0.83} \\
\bottomrule
\end{tabular}
\label{tab:ablation_param}
\vspace{-2mm}
\end{table}

\parag{Inference} 
We follow the DDPM formulation~\cite{Ho20a} with 1000 denoising timesteps. During sampling of \dmapD{}, we introduce multiple guidance terms as detailed in Sec.~\ref{sec:recon_vid}, with weights set to $\lambda_a = 20$, $\lambda_w = 1$, $\gamma = 50$, $\gamma_n = 5$, $\gamma_i = 50$, $\gamma_t = 500$, $\gamma_v = 0.5$, and $\gamma_a = 0.5$.
For spatio-temporal diffusion, we adopt a subsequence-based inference strategy. Specifically, we divide the input video into overlapping 10-frame subsequences with a 5-frame overlap between adjacent subsequences, so each inference pass processes only a short segment. Temporal consistency over the full video is maintained by enforcing consistency within each subsequence and across adjacent subsequences. If the last subsequence contains fewer than 10 frames, we pad it with the last frame.
This strategy keeps the peak memory consumption determined by the subsequence length ($T=10$), rather than the total video length. Therefore, longer videos increase the total processing time but not the peak GPU memory usage, enabling the same inference procedure to be applied to videos of arbitrary~length.
Inference is conducted on a single NVIDIA A100 GPU with 80GB VRAM.
At inference time on real-world videos of clothed humans, we employ Sapiens~\cite{Khirodkar25} to estimate front-view garment normal maps and WHAM~\cite{shin2024wham} to predict human body meshes. These meshes are then rendered into body segmentations and depths, which serve as conditioning inputs.

\begin{figure*}[p]
    \centering
    \includegraphics[width=0.9\linewidth]{fig/revision1/R2_2_image_lowres.pdf}
    \caption{\re{\textbf{Clothing looseness variation.} Columns correspond to garment categories from left to right: T-shirt, trousers, jacket, and skirt. In each column, examples of the same garment category are ordered from top to bottom from tight-fitting to loose-fitting.}}
    \label{fig:supp_tight2loose}
\end{figure*}
\begin{figure*}[hbt]
    \centering
    \includegraphics[width=0.85\linewidth]{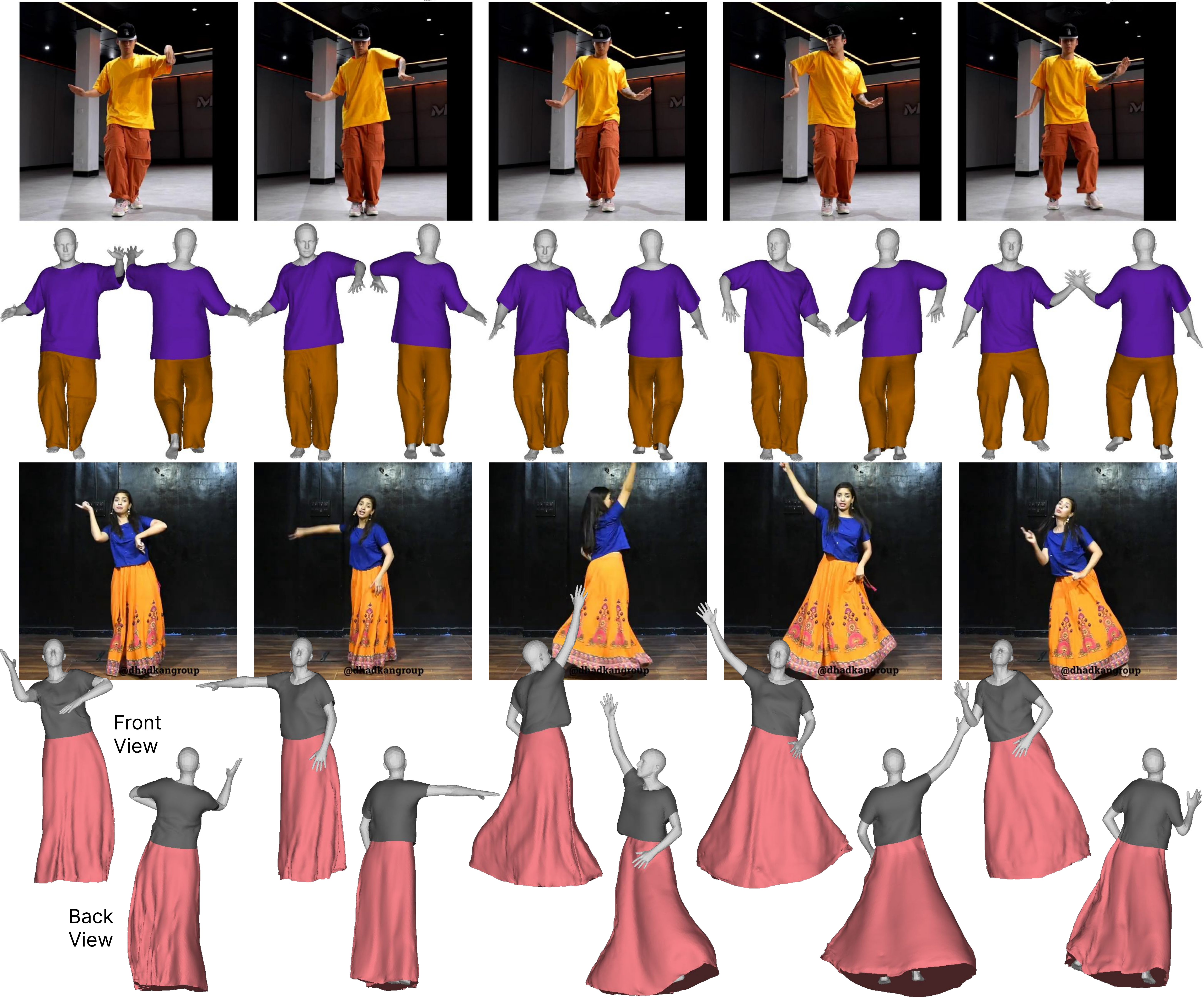}
    \caption{\re{\textbf{Loose-garment reconstruction from Internet videos}. First row: loose-fitting T-shirt and trousers. Second row: skirt.}}
    \label{fig:supp_looseshirt_vid}
\end{figure*}

\parag{Fitting}
For the hyper-parameters introduced in Sec. \ref{sec:fitting_dmap}, we summarize their values in Table \ref{tab:param}. For the regularization term $\mathcal{L}_{phys}$ introduced in Eq. \ref{eq:lphys}, we adopt the default material parameters from \cite{Santesteban22} which are general enough to model most casually-worn garments. In Table~\ref{tab:ablation_param}, we show the quantitative results for the ablation study introduced in Sec.~\ref{sec:ablation_fitting}. The first row shows that omitting the back normal estimation significantly reduces reconstruction accuracy. The comparison between rows 2-5 demonstrates the effectiveness of our refinement strategy, which first optimizes the mesh using network $f_\phi$ before direct vertex optimization.

\subsection{Tight- to Loose-Fitting Variations}
As shown in Fig.~\ref{fig:supp_tight2loose}, we further evaluate our method on tight- to loose-fitting variations for each garment type. For T-shirts, we include sleeveless tops, tight tops, loose T-shirts, long-sleeve shirts, and oversized T-shirts. For trousers, we include tight shorts, tapered pants, cropped pants, straight-leg pants, and wide-leg pants. For jackets, we include examples ranging from cropped jackets to trench coats. For skirts, we include examples ranging from mini skirts to long skirts. We also provide video reconstruction results for loose-fitting garments, especially combinations of loose T-shirts and trousers, as shown in Fig.~\ref{fig:supp_looseshirt_vid}. These results show that our method can reconstruct common garments across a wide range of tight- to loose-fitting shapes. This ability benefits from the ISP/DISP representation, which models 3D garment surfaces in a unified UV space and formulates garment deformations as positional maps. Such a regular 2D domain enables the subsequent diffusion model to directly learn the mapping from image observations to 3D garment deformations in both spatial and temporal dimensions, rather than relying solely on body-driven deformation.

\begin{figure*}[p]
    \centering
    \includegraphics[width=0.95\linewidth]{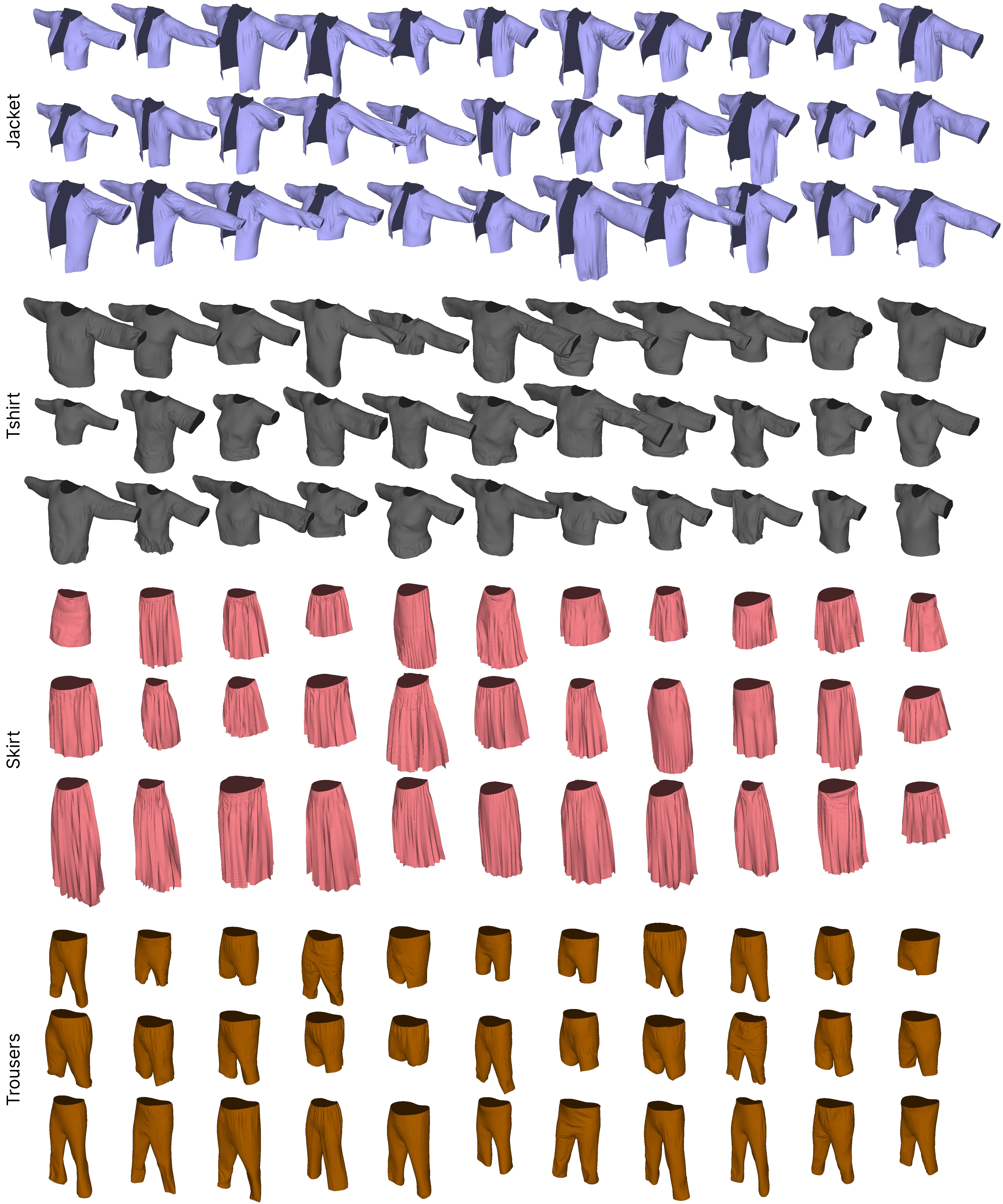}
    \caption{\re{{CLOTH3D dataset on the 33 garments in each category.}}}
    \label{fig:supp_dataset}
\end{figure*}
\begin{figure*}[!t]
    \centering
    \includegraphics[width=\linewidth]{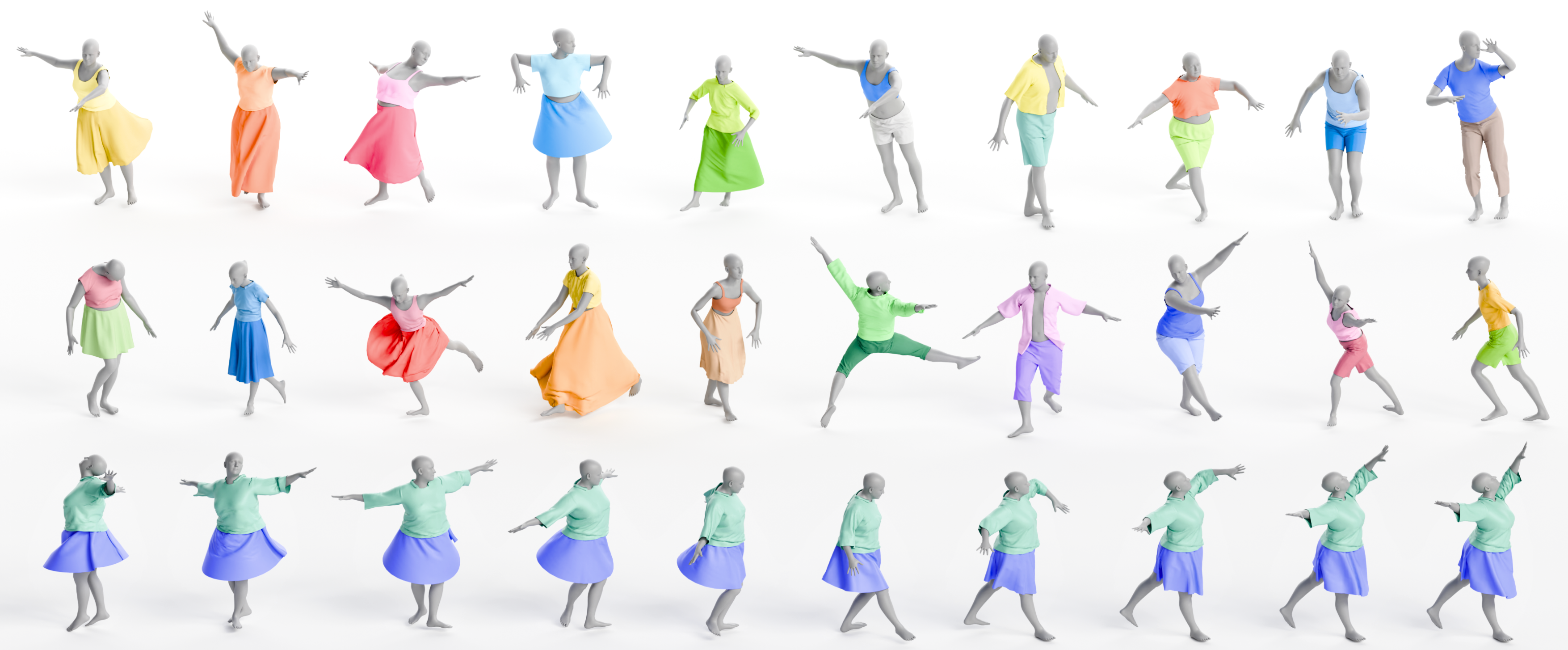}
    \caption{\re{{Re-simulated CLOTH3D garments using AMASS motion sequences.}}}
    \label{fig:supp_amass}
\end{figure*}

\subsection{Dataset Visualization} 
Fig.~\ref{fig:supp_dataset} shows the four clothing categories used from CLOTH3D, including jackets, shirts, skirts, and trousers, with 33 samples in each category. These samples cover diverse garment shapes, ranging from tight-fitting to loose-fitting garments.
Beyond static garment shape, looseness is also influenced by dynamic deformation. Fig.~\ref{fig:supp_amass} shows examples where each garment-body pair is re-simulated using motion data from the dance category of AMASS~\cite{Mahmood19}, which contains diverse and complex motions such as dancing, spinning, and hopping.

\end{document}